%% file: main.tex
\documentclass{article}

\usepackage[margin=1in]{geometry}
\usepackage[numbers, compress]{natbib} 

\usepackage[utf8]{inputenc}    
\usepackage[T1]{fontenc}       
\usepackage{url}               
\usepackage{float}             
\usepackage{booktabs}          
\usepackage{amsfonts}          
\usepackage{nicefrac}          
\usepackage{microtype}         
\usepackage{xcolor}            
\usepackage{placeins}

\usepackage{graphicx}

\RequirePackage{xspace}
\makeatletter
\DeclareRobustCommand\onedot{\futurelet\@let@token\@onedot}
\def\@onedot{\ifx\@let@token.\else.\null\fi\xspace}
\def\eg{\emph{e.g}\onedot}

\def\ie{\emph{i.e}\onedot}

\def\iid{i.i.d\onedot}

\makeatother

\usepackage{amsthm}
\newtheorem{theorem}{Theorem}[section]
\newtheorem{proposition}[theorem]{Proposition}
\newtheorem{lemma}[theorem]{Lemma}

\newtheorem{definition}{Definition}[section]

\input{math_commands}

\input{preamble}

\usepackage{orcidlink}
\usepackage{hyperref}

\usepackage[capitalise]{cleveref}

\crefname{thrm}{theorem}{theorems}
\Crefname{thrm}{Theorem}{Theorems}
\crefname{corollary}{corollary}{corollaries}
\Crefname{corollary}{Corollary}{Corollaries}
\usepackage[capitalise]{cleveref}
\crefname{section}{Sec.}{Secs.}
\Crefname{section}{Sec.}{Secs.} 
\crefname{algorithm}{Alg.}{Algs.}
\Crefname{algorithm}{Alg.}{Algs.} 
\crefname{appendix}{Appx.}{Appxs.}
\Crefname{appendix}{Appx.}{Appxs.} 

\hypersetup{
    colorlinks=true,
    linkcolor=blue,
    filecolor=magenta,      
    urlcolor=cyan,
    citecolor=blue,
}

\definecolor{ad_color}{HTML}{FF7F00}
\definecolor{min_color}{HTML}{E41A1C}
\definecolor{max_color}{HTML}{4DAF4A}
\definecolor{avg_color}{HTML}{A65628}
\definecolor{zero_color}{HTML}{0000FF}
\definecolor{masking_color}{HTML}{006400}
\definecolor{lm_color}{HTML}{984EA3}      
\definecolor{lma_color}{HTML}{17BECF}   

\begin{document}

\title{Activation-Deactivation: A General Framework for Robust Post-hoc Explainable AI}

\author{Akchunya Chanchal, David A. Kelly, and Hana Chockler\\
Department of Informatics\\
King's College London\\
Strand, WC2R 2LS, London, UK \\
\texttt{\{akchunya.chanchal,david.a.kelly,hana.chockler\}@kcl.ac.uk} \\
}

\date{}

\maketitle

\begin{abstract}
Perturbation-based explainability methods face criticism due to their reliance on out-of-distribution
mutants. This raises doubts about the quality of the explanations. In this paper, we introduce a novel
forward pass paradigm, Activation-Deactivation (AD), which obviates the need for perturbation of the input. AD replaces
perturbation of input features with switching off parts of the model corresponding to to the intended perturbations. 
We implement \convad, an \ad approximation algorithm for CNNs. 
\convad is a drop-in mechanism that can be easily added to any trained CNN and,
without any additional training, generates more robust and more transferable explanations. 
We provide evaluation results across multiple architectures, datasets, methods and perturbation strategies, demonstrating
the superior quality of \convad compared to the SOTA.
\end{abstract}

\section{Introduction}

Deep learning models are widely used in a variety of computer systems, including in mission- and safety-critical
applications such as healthcare. Due to the opacity of these models' decision-making
processes, there is an acute need for explanations of their decisions. A good explanation allows for increased
confidence in the system, and for uncovering unexpected failure modes. 

\begin{figure}[t]
\centering
    \begin{subfigure}[b]{0.15\textwidth}
    \label{fig:goat}
        \centering
        \includegraphics[width=\linewidth,height=\linewidth]{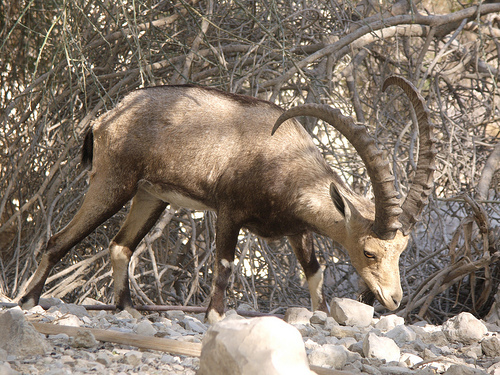}
        \caption{Ibex}
    \end{subfigure}
    \hfill
    \begin{subfigure}[b]{0.15\textwidth}
        \centering
        \includegraphics[width=\linewidth]{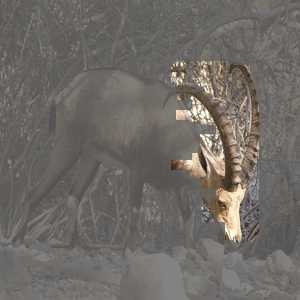}
        \caption{\AD}
    \end{subfigure}
    \hfill
    \begin{subfigure}[b]{0.15\textwidth}
        \centering
        \includegraphics[width=\linewidth]{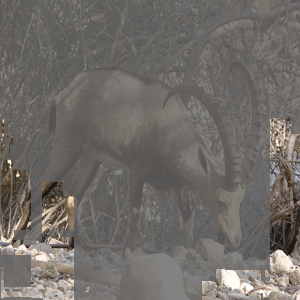}
        \caption{\Min}
    \end{subfigure}
    \hfill
    \begin{subfigure}[b]{0.15\textwidth}
        \centering
        \includegraphics[width=\linewidth]{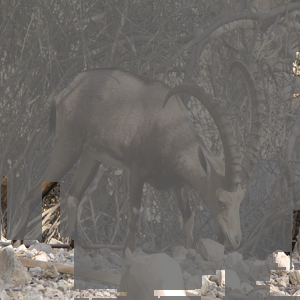}
        \caption{\Zero}
    \end{subfigure}
    \hfill
    \begin{subfigure}[b]{0.15\textwidth}
        \centering
        \includegraphics[width=\linewidth]{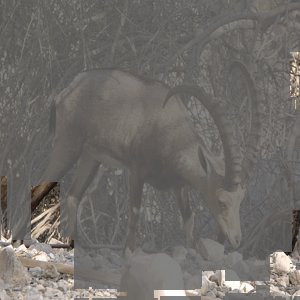}
        \caption{\Avg}
    \end{subfigure}
    \hfill
    \begin{subfigure}[b]{0.15\textwidth}
        \centering
        \includegraphics[width=\linewidth]{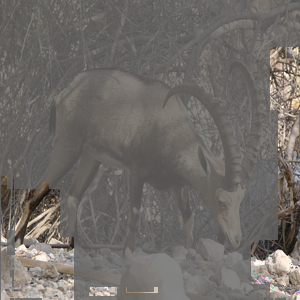}
        \caption{\Max}
    \end{subfigure}
\caption{The choice of masking value can often have a profound effect on the explanation. 
An image of an ibex (a) and its explanations using \ad (b)---our method---compared to occlusion-based methods with
different occlusion values (c)-(f) on the EfficientNet-v2 model. All the images (b)-(f) are classified as ibex, though only
(b)---our method---overlaps with a part of the ibex.}
\label{fig:ibex}
\end{figure}

Unfortunately, definitions of ``explanation'' are almost as abundant as explanations themselves: various
domains, such as computer science \citep{CH97,Gar88,Pea88}, philosophy \citep{Hem65}, and statistics \citep{Sal89}, all
offer different suggestions. \emph{Causal explanations}---also called 
\emph{minimal sufficient pixel sets} (MSPS)~\cite{kelly2025big}---
provide just enough information to get the desired classification from a model. 
An MSPS need not be understandable to a human~\cite{bhusalface}. They reveal what the model does, and does not, accept as belonging to a class. \cref{fig:ibex} shows that an MSPS can frequently be unintuitive to a human, though still accepted by the model.We use MSPS and explanation interchangeably
throughout the paper. 

Existing explainability approaches for neural networks can be roughly divided into \emph{white-box} (model-specific) and
\emph{black-box} (model-agnostic) techniques. White-box, or model-specific techniques, use architecture-specific
manipulations and access to model internals to calculate explanation. Black-box, or model-agnostic
techniques, use only model input and output. Typically, such methods \emph{perturb (mutate)} 
or \emph{occlude (mask)} the
original input and observe the new output. Among the most common e\underline{X}plainable \underline{AI} (\xai) tools using the mutation method are
\lime~\cite{lime}, \shap~\cite{lundberg2017unified} and \rex~\cite{CKKS26}. These create mutants by
masking some of the original features with \emph{masking values}. Masking values can take many forms, such as constants, gaussian blurs or 
functions mapping pixels to values.

Recent work, however, observes that perturbation-based \xai tools often result in out-of-distribution (\ood) inputs,
casting doubt on their ability to produce ``faithful'' explanations~\citep{hooker2019benchmark,zheng2024f}. Choosing the
right mutation strategy is important: if the masking strategy or masking values carry semantic meaning for the 
model, the results might not reflect model behavior on the original input (\cref{fig:ibex}). It is often assumed that
high quality explanations require 
domain-specific adaptations
~\citep{blake2025specrex,yang2026enhancing,zhao2021baylime,botari2020melime,meng2024segal,sivill2022limesegment,haunschmid2020audiolime,mishra2017local}.
In particular, the choice of masking value(s) is assumed to be important. 

In this paper, we introduce \emph{Activation-Deactivation (\ad)}. \ad solves the problem of choosing the correct masking values for
perturbation \xai by removing the need for masking values completely. \ad 
forces the model to disregard subsets of the input by restricting the propagation of the corresponding activations. 
An implementation of \ad is, of course, architecture-specific, but it is model-agnostic within the same architecture, that is, 
the algorithm is the same for all models of the same architecture. In this paper we present an implementation, \convad, for
Convolutional Neural Networks (CNN). 

The idea of input masking without a specific masking value was recently proposed by~\citet{balasubramanian2023towards},
who also implemented it in the tool LayerMask. 
Specifically, LayerMask uses a max-pooling strategy along with neighborhood padding for iteratively restricting the effects of the masked regions on the input.
We show in this paper that \ad addresses the challenge of deactivation in a significantly
more principled fashion than LayerMask, and obtains significantly better results.

We evaluate \convad on three CNN models of different sizes, and provide experimental results on three different
standard datasets. We also compare with LayerMask and three state-of-the-art \xai tools, both black- and white-box. The quality of explanations is evaluated on a number of parameters, including redundancy, transferability, robustness and size. 
\convad is consistently the best, outperforming other methods by $\mathbf{30}$-$\mathbf{40\%}$ on average robustness and up to $\mathbf{22\%}$ better transferability  across all thresholds, while in line with the best methods in terms of redundancy. 
Moreover, the proposed framework does not require any knowledge of masking values and is consistent between the different 
inputs and different datasets.




The rest of the paper is organized as follows. The most important background material is presented in~\cref{sec:Background}. We describe
our theoretical activation-deactivation framework in~\cref{sec:ad_framework} and its implementation, the \convad algorithm, in~\cref{sec:convad}.
The evaluation results and their analysis are in~\cref{sec:exp_results}, and related work is overviewed in~\cref{sec:related}. We conclude
with a brief summary and future work in~\cref{sec:conclusions}.

Due to the lack of space, all proofs, background on actual causality, and detailed experimental results are placed in the appendix. 
The full set of results, code, and datasets are submitted as a part of the supplementary material.


\section{Background}\label{sec:Background}
In this section, we make precise the notion of explanation (MSPS) used in this paper. We also introduce the different \xai tools that we use in our evaluation.

\paragraph*{Causal Explanations} Let the set $\vec{V}$ of boolean variables correspond to the set of pixels $P(x)$ of an image $x$ and the single output variable $O$. Essentially, $\vec{V}$ is a \emph{mask}, indicating which pixels of $x$ are visible and which are occluded, and
the output variable $O$ indicates whether the classification of a partially masked image, $x' = x \odot\vec{V}$, stays the same as the original image ($O=1$). Assigning $1$ to a variable $v_i \in \vec{V}$ means the pixel $p_i$, corresponding
to $v_i$, has its original value (taken from $x$). Assigning $0$ to this variable means that $p_i$ is masked -- replaced with some predefined masking value. 

\begin{definition}[Causal Explanation (MSPS)~\cite{CKKS26}]\label{defn:simple-exp}
A subset $\vec{V}_{exp}$ of $\vec{V}$ is an \emph{explanation} of a classification $\mathcal{N}(x)$ of 
an input image $x$ by a classifier 
$\mathcal{N}$ if the following conditions hold:
\begin{description}
\item[{\rm EXIM1.}] $(M,\vec{u}_0) \models [\vec{V}_{exp} = 1](O=1)$.
\item[{\rm EXIM2.}] $\vec{V}_{exp}$ is minimal; there is no strict subset $\vec{V}'_{exp}$ of $\vec{V}_{exp}$ that satisfies EXIM1.
\end{description}
\end{definition}
In other words, an explanation is a minimal subset $x' \subseteq x$ of pixels of $x$ that is sufficient for the 
model $\mathcal{N}$ to classify both images with the same class, \ie
$\mathcal{N}(x) = \mathcal{N}(x')$.

\rex~\cite{CKKS26} is based on the theory of actual causality~\citep{Hal19,CH24}. Of all the tools used in this
paper, it is the only one which directly produces causal explanations, or MSPSs. While there are versions of \rex for different
modalities, we use the implementation for black-box
image classifiers. Roughly speaking, \rex approximates \emph{sufficient responsibility}~\cite{CKKS26} and uses
the resulting \emph{responsibility map} as a saliency landscape from which it extracts single (or multiple~\cite{chockler2023multiple}) MSPSs.
 \lime~\cite{lime} is a black-box explainability tool which 
approximates a model's behavior in the local vicinity of the instance to be explained. \lime calculates a set
of perturbed instances, derived from the original sample via different strategies. A set of labels is then calculated
for the perturbed instances by querying the model on the samples. A surrogate model is constructed using the perturbed samples with their labels. The surrogate model's parameters are interpreted as \emph{saliency} scores. 
\gradcam~\cite{selvaraju2017grad} is a white-box tool for CNNs which leverages the gradients of logit for a target class
with respect to the activations of a specific convolutional layer in the model. Typically, gradients are calculated with
the activations of the final convolutional layer. 

It is a relatively straightforward task to use \lime and \gradcam saliency output as input for the \rex MSPS discovery algorithm~\cite{CKKS26}. This allows up to calculate MSPSs for these tools as well. We use MSPSs---causal explanations---as they allow for a consistent evaluation of a wide variety of tools which otherwise have orthogonal notions of explanation.

\section{The Activation-Deactivation (\ad) framework}
\label{sec:ad_framework}

\input{block_diagram}

The \ad algorithm \emph{deactivates} the activations that correspond to occluded input features during inference (\cref{fig:architecture}).
This halts the propagation of the effects to the next layer of the model. In what follows, we assume the same model $\mc{N}$, which
we omit from the notation.
We introduce two new families of functions: position ($\mathit{pos_i}$) and position-attribution ($\Phi_i$), where
$i$ is a layer of $\mc{N}$. For ease of presentation, we describe the framework for two-dimensional inputs, but the procedure generalizes to $n$ dimensions.

\dfn[Position function]\label{def:pos}
$\mathit{pos_i}(\mbf{z}^{(i)}_{ab})$ is an inverse
mapping from a position $(a,b)$
in the set of intermediate representations $\mbf{z^{(i)}}$ to the set of previous layer features pertaining to $\mbf{z}^{(i)}_{ab}$.  It returns the set of positions of elements from the previous layer that map to $\mbf{z}^{(i)}_{ab}$.
\[pos_i(\mbf{z}^{(i)}_{ab}) = \left[(k,l), (c,d), \dots, (m,n) \right]^\mbf{T}\]
\edfn

\dfn[Position-Attribution function]\label{def:posat}
$\Phi_i(\mbf{z}^{(i)}_{ab},M_i)$, given mask $M_i$,
returns the ratio of masked to unmasked input features pertaining to $\mbf{z}^{(i)}_{ab}$.
    \[\Phi_{i}(\mbf{z}^{(i)}_{ab},M_{i}) = \frac{1}{|pos_{i}(\mbf{z}^{(i)}_{ab})|}\sum_{pos_{i}(\mbf{z}^{(i)}_{ab})} \mbf{1}(M_{i}[k,l] > 0),\]
    where $\mbf{1}$ is the indicator function, and $M_{i} [k,l]$ is a cell in the mask $M_{i}$. A total of $0$ indicates that all the features in the current operation are masked, and $1$ that they are all unmasked. Intermediate values represent varying ratios of masked and unmasked features. Again, similar definitions apply for $n$-dimensional inputs.
 \edfn

We set a hyperparameter, $\tau$, as a threshold for $\Phi_i(\mbf{z}^{(i)}_{ab},M_{i})$, below which sets of activations are discarded. Setting $\tau$ to $0$ only deactivates fully masked regions; values above $0$  deactivate some unmasked parts that border the
masked regions.

The output of  $\Phi_i(\mbf{z}^{(i)}_{ab},M_{i})$ is used to compute an updated mask $M'_i$ for the layer $i$ as follows:
    \[M'_i [a,b] = \mbf{1}(\Phi_i(\mbf{z}^{(i)}_{ab},M_{i}) > \tau).\]

Deactivation need not be performed at every step in the model. It is only performed prior to a layer resizing operation
or an operation which averages or converts data which includes deactivated values.
The locations where we perform deactivation are called \emph{checkpoints}.

We can now formally define Activation-Deactivation. Note that we use $\Phi_i(\cdot)$ to illustrate that position-attribution and thresholding have been performed over all $\mbf{z}$.
\dfn[Activation-Deactivation (\ad)]\label{def:AD}
Given a pair $(\mbf{x}, M)$ of the input $\mbf{x}$ and the accompanying binary mask $M \in \mb{R}^{m \times n}$ and a Deep Neural Network (DNN) $\mc{N}$, the output, $O^{(i)}$, at a checkpoint after an intermediate layer $\mc{N}^{(i)}$ in a \ad forward pass, is defined as:
\[O^{(i)} = \mc{N}_{(i)}(\mbf{z^{(i)}}) \odot M'_{i},\]
where $M'_{i}$ is the updated mask at the checkpoint $i$. In the first checkpoint we set $M'_{i} = M$.
The \ad output of a given layer is defined as the 2-tuple:
\[\ad^{(i)} = (O^{(i)}, M'_{i}).\]
\edfn

\section{The \convad Algorithm}\label{sec:convad}
\label{subsec:intuition+algo}

We implement \ad for a CNN in \convad (\cref{alg:ad_forward}). \convad sets the values of intermediate representations pertaining to masked regions to zero at strategic locations (\convad checkpoints) throughout the model. Checkpoints are locations where the current intermediate representations pertaining to masked regions will have a downstream influence unless their effects are removed in situ. \convad handles two different types of checkpoints, namely: i) $shape(\mbf{z})$ is constant between subsequent convolution layers $C^{i}$ and $C^{i+1}$; and ii) $shape(\mbf{z})$ is altered between $C^{i}$ and $C^{i+1}$.

Case (i) is the scenario in which only shape-preserving operations have been performed since the last convolution operation. These are layers such as activation layers, dropout layers and normalization layers. In this case, the checkpoint is immediately after the last operation prior to the convolutional layer $C^{i+1}$. In case (ii), \emph{dimensionality-altering (DA)} operations have been performed prior to the convolutional layer $C^{i+1}$. Examples are parametric operations which alter dimensionality (such as bottleneck layers) or an external additive or subtractive procedure (\eg interpolation). In this scenario, in addition to the checkpoint for case (i), two additional checkpoints are added, one immediately prior to the DA operation and one immediately after. The checkpoint immediately prior ensures that we deactivate the effects of prior shape-preserving operations over the masked regions that may have occurred between $C^{i}$ and the DA operation.

The position-attribution function for the binary mask in case (i) is an identical convolution to $C^{i}$ with a mean-value kernel ($C_{ij} = \frac{1}{m \times n}$). This results in an identically-shaped representation as $\mbf{z}^{(i)}$, where each entry is a scalar between 0 and 1, denoting the fraction of unmasked-masked features in the corresponding output of $C^{i}$. In case (ii), the position-attribution function for a parametric dimension-altering operation is dependent on the operation being performed. In the case of external additive procedures, depending on whether the influence of the addition is to be considered, the mask is padded with $1$s or $0$s at positions corresponding to the added representations. Similarly, for subtractive procedures, we remove the positions from the mask corresponding to the removed representations. Once the required representations and their masking values have been added or removed due to external procedures, we convolve the updated mask with the mean-value kernel in order to get the position-attribution.

The \convad operation can be expressed as:
\begin{equation}
    O^{(i)} = \begin{cases}
        \mbf{z}^{(i)} \odot \mbf{1}^{\oplus}(\sum(\frac{J_{m \times n}}{m \times n} * M_{i}) > \tau), &  \text{if}\, (i) \\
        \begin{cases}
        \mbf{z}^{(i)} \odot \mbf{1}^{\oplus}(\sum(\frac{J_{m \times n}}{m \times n} * M_{i}) > \tau) & \text{prior to DA}\\
        \mbf{z}^{(i)} \odot \mbf{1}^{\oplus}(\Phi_{i}(M'_{i}) > \tau)) & \text{post DA}
        \end{cases} &\text{if}\, (ii)
    \end{cases}
\end{equation}
The symbol $\oplus$ denotes that the updated mask is broadcast to $\mb{R}^{shape(\mbf{z})}$ in order to apply the Hadamard product. $J_{m \times n}$ denotes a $m \times n$ matrix of ones. $M'_{i}$ is the updated mask, post checkpoint.

\begin{algorithm}[t]
    \caption{\convad Forward Pass}
    \label{alg:ad_forward}
    \begin{algorithmic}
        \STATE \textbf{Input:} Input $x$, Binary Mask $M$, AD model $\mc{N}$, an activation threshold $\tau$
        \STATE \textbf{Output:} Final layer output $\mbf{z}$
    \end{algorithmic}
    \begin{algorithmic}[1]
         \STATE $\mbf{z} \leftarrow x$ 
        \FOR{each layer $\mc{N}_{(i)} \in \mc{N}$}
            \STATE $\mbf{z} \leftarrow \mc{N}_{(i)}(\mbf{z})$
            \IF{$\mc{N}_{(i)}$ is conv}
                \STATE $\frac{J_{m \times n}}{m \times n} \leftarrow$ initialize kernel with $shape(\mc{N}^{k}_{(i)})$ and all values set to $\frac{1}{m \times n}$
                \STATE $M \leftarrow \frac{J_{m \times n}}{m \times n} * M$
                \STATE $M \leftarrow M > \tau$
            \ENDIF

            \IF{$\mc{N}_{(i)}$ is DA operation}
                \STATE $M \leftarrow \Phi_{i}(M)$
                \STATE $M \leftarrow M > \tau$
            \ENDIF

            \IF{checkpoint}
                \STATE $\mbf{z} \leftarrow \mbf{z} \odot M$
            \ENDIF

        \ENDFOR
        \STATE \textbf{return} $\mbf{z}$
        \end{algorithmic}
\end{algorithm}

Given a CNN $\mc{N}$, we denote by $\mc{N'}$ the result of
instrumenting $\mc{N}$ with \convad checkpoints. Let $b_1$ be a binary mask with all values set $1$ (no masking), 
and $x$ is a valid input for $\mc{N}$. Then we have:

\begin{theorem}
\label{thm:equivalence}
    $\forall x, \mc{N}(x) = \mc{N'}(x, b_1)$.
\end{theorem}
\noindent
Intuitively, if the mask is ``empty'', then both models behave exactly the same in terms of their input/output characteristics. See \cref{sup:sec:equiv_proof} for proof. This is an important difference between \convad and LayerMask~\cite{balasubramanian2023towards}. LayerMask does not provide such guarantees, leading to scenarios where the original model and the LayerMask-modified variant do not agree even in the fully unmasked case.

\begin{figure}[t]
    \centering
    \begin{subfigure}[t]{0.3\textwidth}
        \centering
        \includegraphics[scale=0.15]{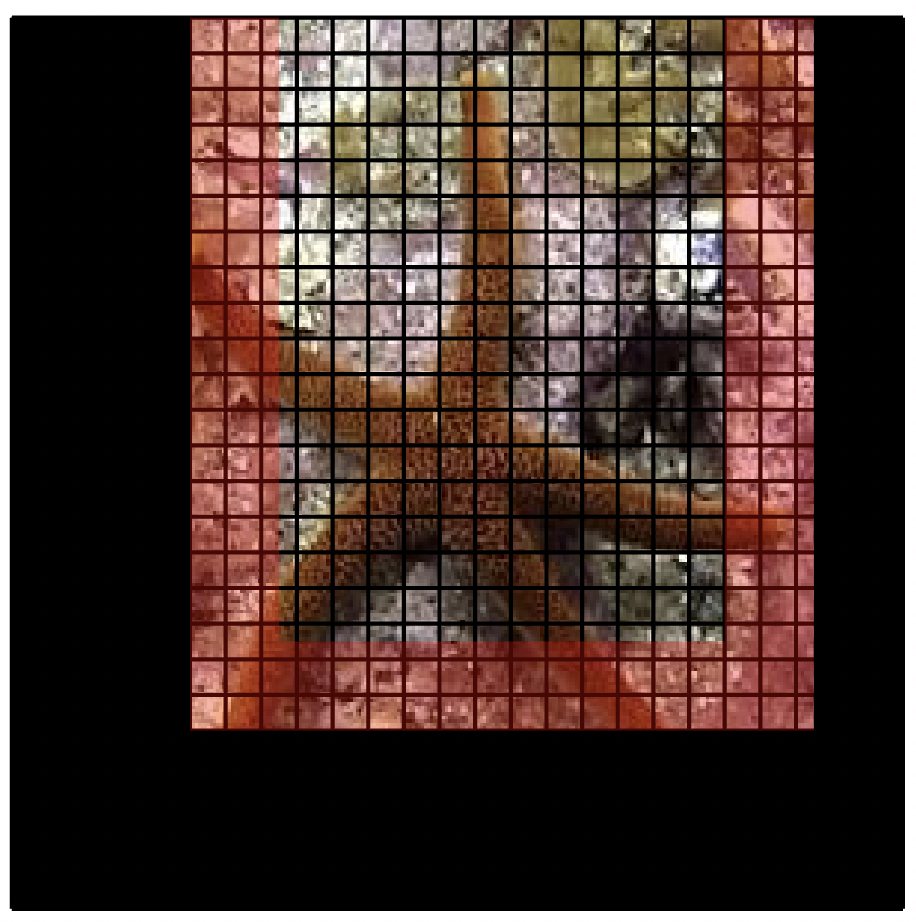}
        \caption{Mixture of masked and\\unmasked regions }\label{fig:leak1}
    \end{subfigure}
    \begin{subfigure}[t]{0.3\textwidth}
        \centering
        \includegraphics[scale=0.15]{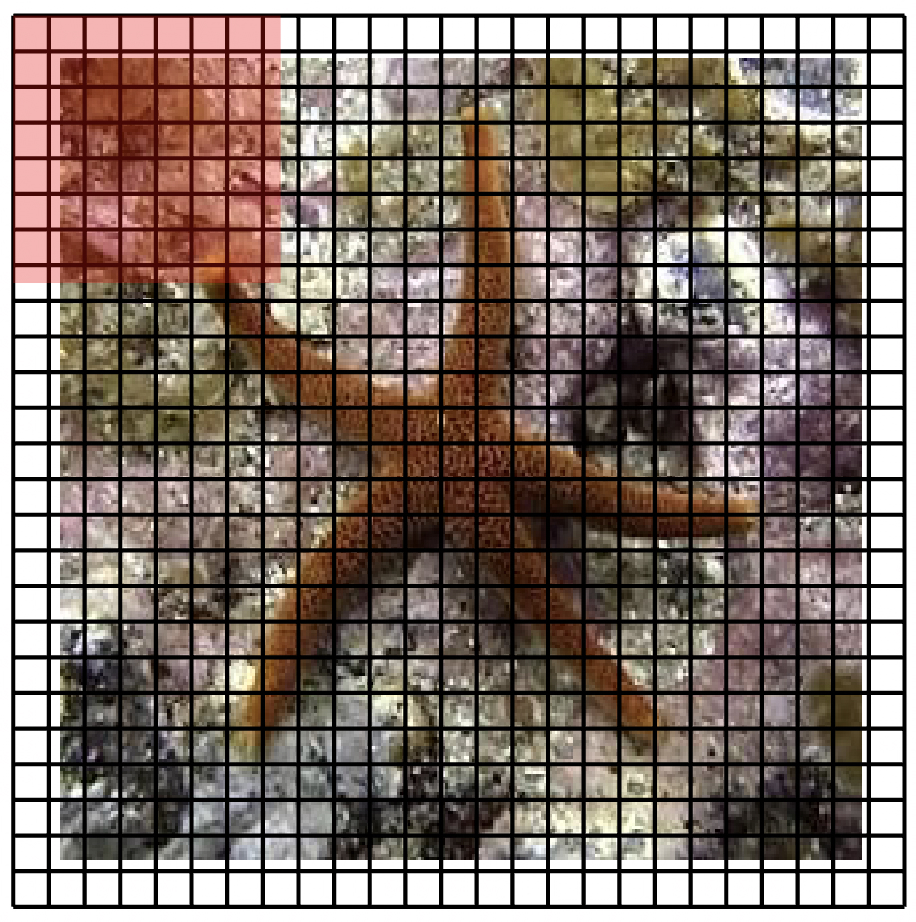}
        \caption{Mixture of input features\\ and padding}\label{fig:leak3}
    \end{subfigure}
    \begin{subfigure}[t]{0.3\textwidth}
        \centering
        \includegraphics[scale=0.15]{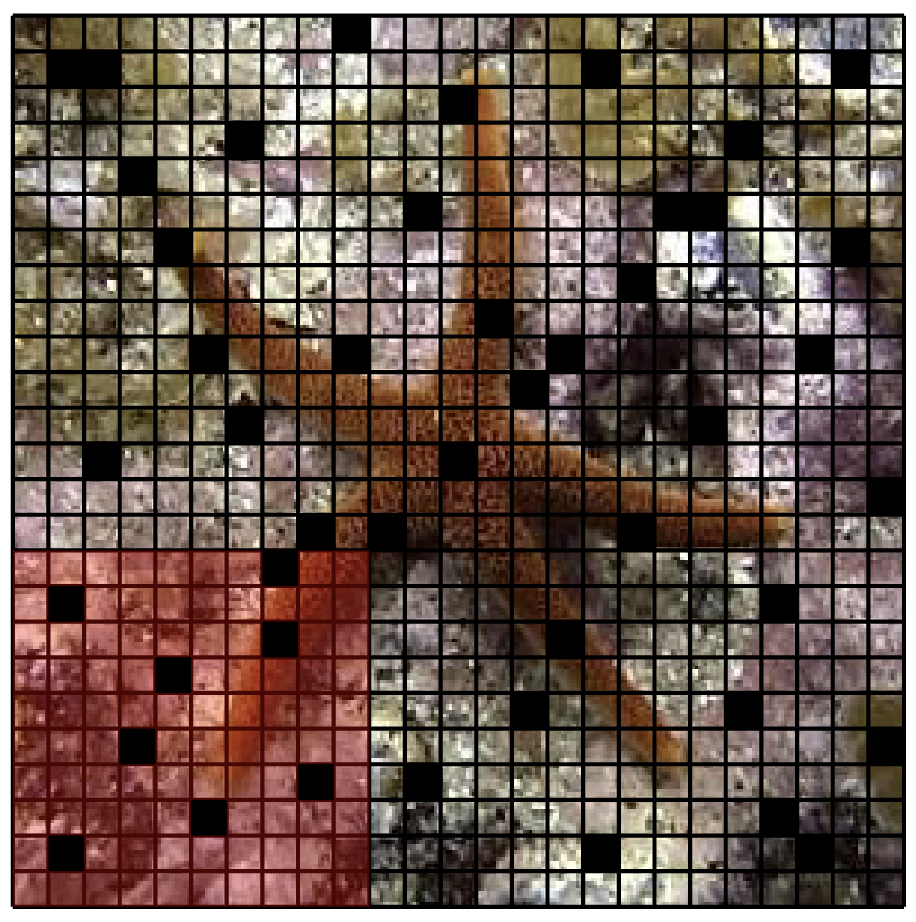}
        \caption{Random perturbations}\label{fig:leak2}
    \end{subfigure}
    \caption{Scenarios where activation leaks can occur when applying \convad.}\label{fig:leaks}  
\end{figure}

\paragraph*{Conceptual limitations}%
\label{subsec:leaks}
One conceptual limitation of \convad is \emph{leakage}. 
As discussed, \convad relies on a threshold value, $\tau$, used to determine if an activation is turned on or off. This is because it is impossible to separate the effects of masked and unmasked examples for certain edge cases and masking strategies.
\cref{fig:leaks} depicts three possible scenarios where the convolution operation straddles masked and unmasked regions of the input. In such scenarios, we are faced with two issues.
\textbf{1}) for scenarios such as \cref{fig:leak1,fig:leak3}, the convolution window includes, in addition to an unmasked
region, a portion of either the masked region (\ref{fig:leak1}) or the padding (\ref{fig:leak3}).
\textbf{2}) in \cref{fig:leak2}, due to non-contiguous perturbation strategies, the convolution windows overlaps with each perturbations, making it
impossible to determine which regions to deactivate.

Another limitation is the need to access and modify the internals of the model. While the \convad algorithm does not change between the CNNs,
its implementation requires a white-box access to the model. Moreover, the current version of the algorithm works only for CNNs.

\section{Experimental Results}\label{sec:exp_results}

\input{masking_examples}


In this section we describe our experimental setup and evaluation of \convad. We compared \convad against a selection of
masking values. We also evaluated against LayerMask~\cite{balasubramanian2023towards}. 
We evaluated \convad using $3$ convolutional models:
RegNetY-12GF~\citep{radosavovic2020designing},
ResNet-50~\citep{He2016DeepRL}, and EfficientNet-V2~\citep{tan2021efficientnetv2}. We used $3$ standard datasets:
ImageNet-1k~\citep{deng2009imagenet}, ImageNet-v2~\citep{recht2019imagenet} and PASCAL-VOC~\citep{pascal-voc}. 
We selected 150 images uniformly at random from the validation sets from
each dataset. For all models and datasets, MSPSs were calculated using \rex~\cite{CKKS26}, \lime\cite{lime} and
\gradcam\cite{selvaraju2017grad}. We used the \rex discovery algorithm to obtain MSPSs from
\lime and \gradcam. We evaluated explanation quality using five different properties: \textit{robustness}, \textit{transferability}, 
\textit{size}, \textit{redundancy} and \textit{overlap}. All experiments where evaluated on Nvidia A100s, A30s and H200 GPUs. \cref{fig:masking_examples} provides a number of examples in which all masking strategies output unintuitive results, whereas
\convad's output is always consistent with our intuition.

The $\tau$ hyperparameter (\cref{sec:ad_framework}) is set to $0$ for all experiments. As LayerMask and the other \xai
tools have no equivalent to $\tau$, we maximally disadvantage \convad by using it in its strictest setting. Our results
show that, even with the over-approximation of deactivating all relevant neurons, \convad still outperforms the other tools.
For each image, we calculated a set of explanations by varying the \emph{confidence threshold}
$\gamma$ of the explanation as a multiplication factor of the model's confidence on the original example. We calculated
explanations at $\gamma = 0, 0.1, 0.3, 0.5, 0.7$ and $0.9$ (that is, for \eg $\gamma=0.9$ we only consider explanations 
with confidence  $\geq 90\%$ of the original classification). \

\paragraph*{Robustness}
We adapt the definition of goodness of partial explanations in~\citet{Hal19} to quantify the 
robustness of a causal explanation, $\vec{X} = \vec{x}$. 
$\K$ is a set of images in which we insert an MSPS.

\dfn[Robustness of explanation]\label{def:robust}
The robustness, $\beta$, of an explanation $\vec{X} = \vec{x}$ for $O=o$ wrt a set of images $\K$
is the proportion of $\K$ in which setting $\vec{X}$ to $\vec{x}$ results in $O=o$.
\edfn

$\beta = 1$ indicates the output $O=o$ is retained in all images in $\K$. Likewise, an
explanation with $\beta = 0$ is unable to retain the classification in any images of $\K$.
We considered two sets of images for robustness. The set
$\K_1$ consists of solid-colored images, while 
the set $\K_2$ consists of images of different classes drawn from the same dataset. 
In both cases, we use $100$ backgrounds against which to measure each explanation.

\begin{figure*}[t]
    \centering
    \input{Plots/updated_robustness_rex}
    \caption{$\beta$-robustness (rows, from $0$ to $1$ in steps of $0.2$) of explanation against low information
backgrounds on ImageNet-1k, ImageNet-v2 and PASCAL-VOC for our models, with different confidence
thresholds $\gamma$. \AD explanations are consistently more robust than ones computed using masking values, across all
masking values and confidence thresholds. Error bars represents Standard Error of the Mean (SEM).}%
\label{fig:robustness_bg_color}
\end{figure*}





\textbf{Robustness against solid-color backgrounds}:
\cref{fig:robustness_bg_color} demonstrates the results for \AD, \Min, \Max, \Avg and \Zero, \LM and \LMA for solid backgrounds.
The results are reported for all thresholds $\gamma$.
In most cases, the \AD explanations are significantly more robust than all masking strategies; this is especially prominent
for RegNetY-12GF and EfficientNet-V2 models.
See~\cref{appendix:robust_iid} for results on \lime and \gradcam.
An important observation is that there is \emph{no consistent best masking value} for a given dataset or model. In RegNetY, we observe similar results for \Min and \Max for ImageNet-1k and ImageNet-v2, however, in PASCAL-VOC, using the \Zero value seems to be the most appropriate strategy. In contrast, \ad consistently is the best---or near-best---performing method across all datasets and models, and does not require any decision about occlusion technique or value.

\textbf{Robustness against IID backgrounds}:
A similar robustness pattern is found when inserting explanations over \iid backgrounds (\cref{appendix:robust_iid}). 
The overall level of robustness is lower, with \convad having approximately $20\%$
robustness at $\gamma = 0.9$. This behavior is easy to explain: every model tested \emph{must} provide a
classification. Even a solid color background has a classification, albeit of low confidence. Examination of the output
tensor for solid backgrounds reveals a close-to-flat probability distribution, making it relatively easy to change the
top class by injecting an explanation for a different class. This distribution flatness does not hold for \iid images
It therefore requires more ``work'' from the injected
explanation in order to overturn the existing classification.

\paragraph*{Inter-model transferability}
We calculated the \textit{transferability} of MSPSs across models~\cite{kelly2026if}, that is, we checked if the MSPS for one model is also accepted by another. A high transferability score signifies that the MSPS calculated for the model rely on truly discriminative features and are not simply a product of shortcut learning. Transferability is evaluated for all combinations of model pairs and \textit{across} all of the evaluated strategies, \eg, an explanation calculated via strategy A in the first model is evaluated in the second model by filling the masked region with values of strategy B (in the case of \convad, by deactivating the masked region). Note, transferability is also evaluated in the scenario when the strategy is kept fixed.
\convad consistently has higher rates of transferability compared to LayerMask and the different masking strategies. A red herring in the results is the high transferability of LayerMask (ablated). This is because LayerMask (ablated) identifies significantly larger explanations (on average $241.7\%$ larger than the next largest explanation) with drastically poorer identification of truly salient features. This, however, leads to a skewing of the transferability results due to a majority of the image still being present in the explanation 
(full results are in~\cref{appendix:transferability}).

\paragraph*{Relationship between robustness, size of explanations and redundancy}
We identify the relationship between the robustness of the explanation, the size and the amount of ``noise'' within the calculated explanation. To calculate noise within an MSPS, we try to find the expected percentage of the MSPS that can be deleted, while still maintaining the correct classification and threshold barrier. We do this by normalizing the saliency values pertaining to the MSPS, subtracting the normalized value from 1 to attain the sampling distribution of removal. We then take a sample from the distribution (pixels to remove), remove the pixels and query the model to check if the updated MSPS is still above the confidence threshold and returns the original classification. The sampling and removal continue until the updated MSPS fails to meet the confidence threshold or original classification criteria, giving us the deletion percentage. To approximate the expected deleted percentage, for each image, we perform 100 iterations of the sample and delete procedure and then take the mean of all the deletion percentages (full results are in~\cref{sup:sec:avg_size}). 
\cref{fig:3d_plots} shows the relationship for ResNet and RegNet in ImageNet-1k. It 
demonstrates that \convad explanations lie primarily in the 'goldilocks' zone: they have high robustness, low noise and small explanation size. MSPSs generated from other strategies fall short on one or more of these factors.

\definecolor{antiquefuchsia}{rgb}{0.57, 0.36, 0.51}
\definecolor{arylideyellow}{rgb}{0.91, 0.84, 0.42}
\definecolor{babyblue}{rgb}{0.54, 0.81, 0.94}
\definecolor{darkspringgreen}{rgb}{0.09, 0.45, 0.27}
\definecolor{darkpastelblue}{rgb}{0.47, 0.62, 0.8}
\begin{figure}[t]
    \centering
    \begin{subfigure}[b]{0.3\textwidth}
        \centering
        \includegraphics[scale=0.22]{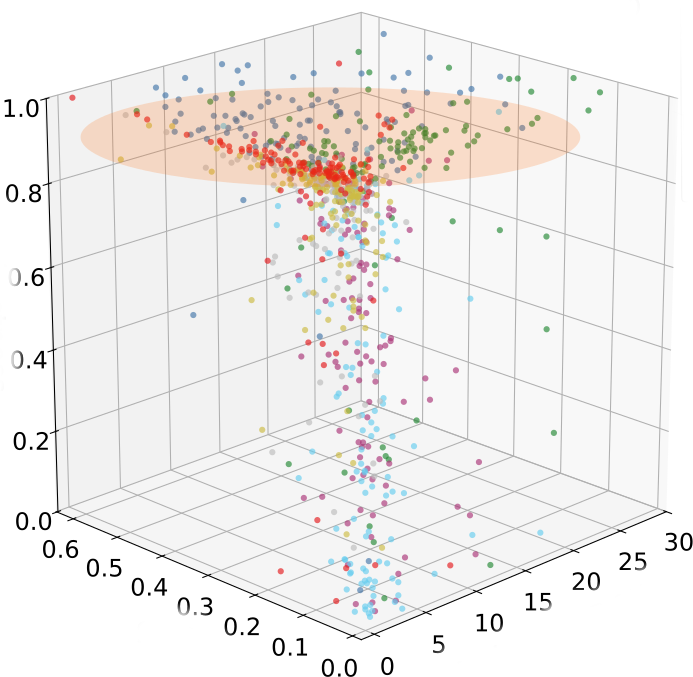}
        \caption{Regnet}\label{subfig:regnet}
    \end{subfigure}
    \hfill
    \begin{subfigure}[b]{0.3\textwidth}
        \centering
        \includegraphics[scale=0.22]{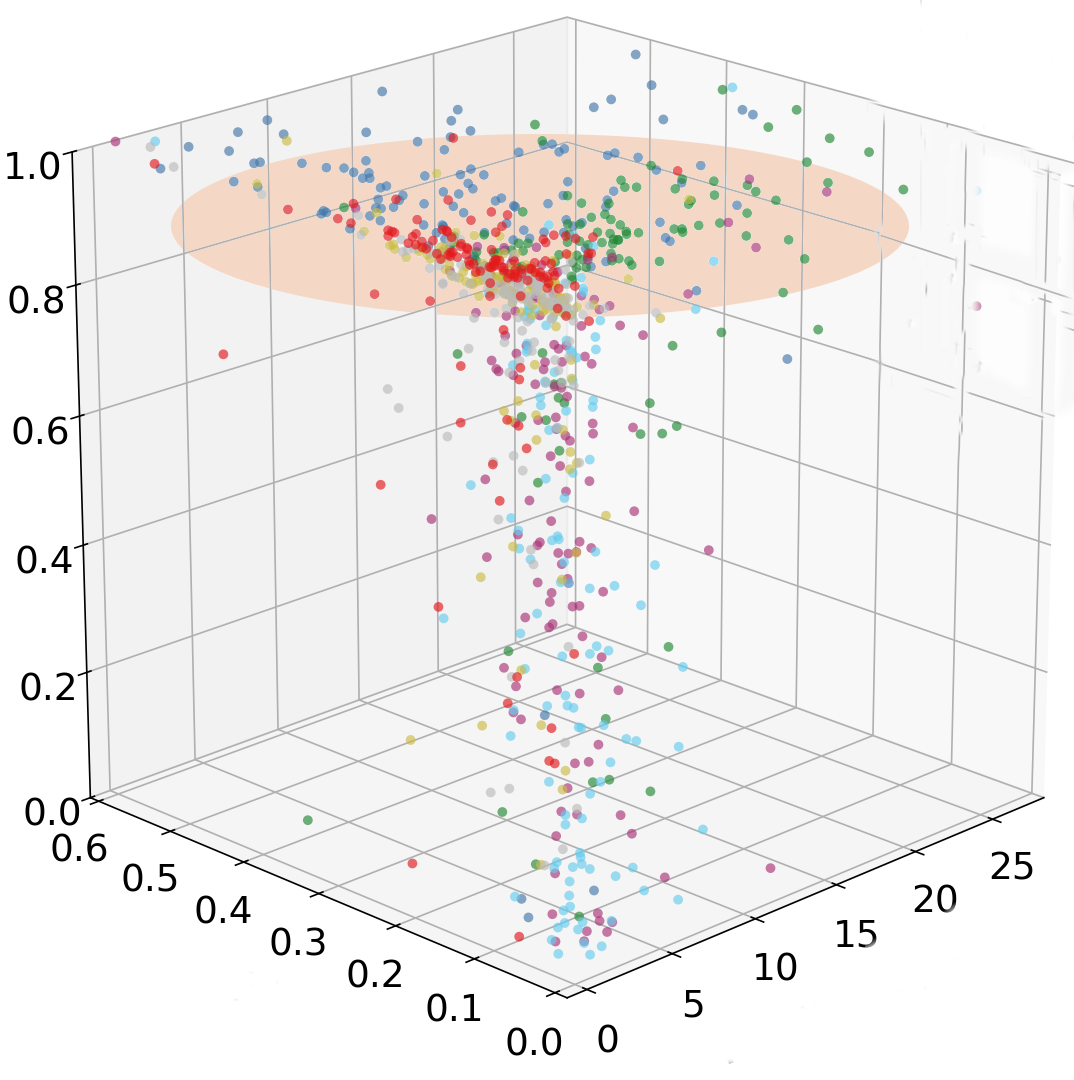}
        \caption{Resnet}\label{subfig:regnet_correct}
    \end{subfigure}
    \hfill
    \captionsetup[subfigure]{labelformat=empty}
    \begin{subfigure}[b]{0.3\textwidth}
        \begin{tikzpicture}
             \node at (0,0) [
             minimum height=3cm,
             align=left,
             ] (legend) {
             \textcolor{red}{ConvAD}\\
             \textcolor{darkpastelblue}{LayerMask (ablated)}\\
             \textcolor{darkspringgreen!60}{LayerMask}\\
             \textcolor{babyblue}{Zero}\\
             \textcolor{arylideyellow}{Max}\\
             \textcolor{antiquefuchsia}{Mean}\\
             \textcolor{black!20}{Min}\\ \\};
        \end{tikzpicture} 
        \caption{}
    \end{subfigure}
    \caption{3D plot demonstrating the results of Noise $\times$ Robustness $\times$ Size of explanations for RegNet and ResNet on ImageNet-1k. The ovals mark the
    `goldilocks' zone.}
    \label{fig:3d_plots}
\end{figure}

\paragraph*{Overlap of explanations}
We calculated the \textit{overlap} of the MSPSs for each model and dataset to investigate the similarity between each strategy, given the differences in the robustness. We report the DICE\cite{Dic45} and IoU\cite{finley1884tornado, jaccard_index_iou}.
MSPSs generated by masking strategies tend to have high similarity to each other, having a DICE score between $0.56$ and $0.78$. The overlap of masking MSPSs against \convad decreases noticeably, with maximum DICE scores between $0.2$ and $0.47$. Interestingly, LayerMask explanation have the lowest overlap against masking with DICE scores between $0.17$ and $0.27$ (\cref{appendix:overlap}).
The results suggest that \convad identifies different MSPSs to masking strategies, not merely pixel supersets. Additionally, 
LayerMask explanations are distinct from either masking or \convad explanations, which may explain its poor robustness scores.

\paragraph{Limitations of evaluation}
There is no consensus in the literature regarding the evaluation, or even the correct definition of
quality of explanations. Multiple proxies has been used as an indication of
quality. In this work, we follow this approach, but aim to minimize this limitation by computing a number of well-justified measures.
\input{related}

\section{Conclusions and Future Work}\label{sec:conclusions}
We have introduced \ad, a forward-pass paradigm which obviates the need for choosing occlusion values in post-hoc explainability. We have proved theoretical results regarding the explanations calculated in \ad. We have presented
\convad, a drop-in mechanism that can be added to any trained CNN and which, without any further training or fine-tuning, provides more robust explanations that also adhere more strictly to the confidence threshold of the model and do not require expert knowledge in choosing a neutral masking value. 
We will expand the \ad framework to transformers and other attention-based models in future work.


\bibliographystyle{unsrtnat}
\bibliography{all}

\input{appendix}
\end{document}

%% file: math_commands.tex
\usepackage{amsmath,amsfonts,bm}

\def\eqref#1{equation~\ref{#1}}

\def\1{\bm{1}}

\DeclareMathAlphabet{\mathsfit}{\encodingdefault}{\sfdefault}{m}{sl}
\SetMathAlphabet{\mathsfit}{bold}{\encodingdefault}{\sfdefault}{bx}{n}



%% file: preamble.tex
\usepackage[utf8]{inputenc}
\usepackage{pgfplots}
\usepgfplotslibrary{groupplots,dateplot}
\usetikzlibrary{patterns,shapes.arrows}
\pgfplotsset{compat=newest}

\usepackage{xspace}
\usepackage{amsmath}
\usepackage{mathtools}
\usepackage{amsthm}
\usepackage{nicefrac}
\usepackage{amsfonts}
\usepackage{pifont}
\usepackage{floatflt}
\usepackage{url}
\usepackage{dsfont}

\usepackage{tikz}
\usetikzlibrary{backgrounds,calc,shadings,shapes.arrows,arrows,shapes.symbols,shadows,positioning,decorations.markings,backgrounds,arrows.meta,shadows,matrix}

\makeatletter
\newif\if@showgrid@grid
\newif\if@showgrid@left
\newif\if@showgrid@right
\newif\if@showgrid@below
\newif\if@showgrid@above
\tikzset{%
    every show grid/.style={},
    show grid/.style={execute at end picture={\@showgrid{grid=true,#1}}},%
    show grid/.default={true},
    show grid/.cd,
    labels/.style={font={\sffamily\small},help lines},
    xlabels/.style={},
    ylabels/.style={},
    keep bb/.code={\useasboundingbox (current bounding box.south west) rectangle (current bounding box.north west);},
    true/.style={left,below},
    false/.style={left=false,right=false,above=false,below=false,grid=false},
    none/.style={left=false,right=false,above=false,below=false},
    all/.style={left=true,right=true,above=true,below=true},
    grid/.is if=@showgrid@grid,
    left/.is if=@showgrid@left,
    right/.is if=@showgrid@right,
    below/.is if=@showgrid@below,
    above/.is if=@showgrid@above,
    false,
}

\def\@showgrid#1{%
    \begin{scope}[every show grid,show grid/.cd,#1]
    \if@showgrid@grid
    \begin{pgfonlayer}{background}
    \draw [help lines]
        (current bounding box.south west) grid
        (current bounding box.north east);
    \pgfpointxy{1}{1}%
    \edef\xs{\the\pgf@x}%
    \edef\ys{\the\pgf@y}%
    \pgfpointanchor{current bounding box}{south west}
    \edef\xa{\the\pgf@x}%
    \edef\ya{\the\pgf@y}%
    \pgfpointanchor{current bounding box}{north east}
    \edef\xb{\the\pgf@x}%
    \edef\yb{\the\pgf@y}%
    \pgfmathtruncatemacro\xbeg{ceil(\xa/\xs)}
    \pgfmathtruncatemacro\xend{floor(\xb/\xs)}
    \if@showgrid@below
    \foreach \X in {\xbeg,...,\xend} {
        \node [below,show grid/labels,show grid/xlabels] at (\X,\ya) {\X};
    }
    \fi
    \if@showgrid@above
    \foreach \X in {\xbeg,...,\xend} {
        \node [above,show grid/labels,show grid/xlabels] at (\X,\yb) {\X};
    }
    \fi
    \pgfmathtruncatemacro\ybeg{ceil(\ya/\ys)}
    \pgfmathtruncatemacro\yend{floor(\yb/\ys)}
    \if@showgrid@left
    \foreach \Y in {\ybeg,...,\yend} {
        \node [left,show grid/labels,show grid/ylabels] at (\xa,\Y) {\Y};
    }
    \fi
    \if@showgrid@right
    \foreach \Y in {\ybeg,...,\yend} {
        \node [right,show grid/labels,show grid/ylabels] at (\xb,\Y) {\Y};
    }
    \fi
    \end{pgfonlayer}
    \fi
    \end{scope}
}
\makeatother
\tikzset{every show grid/.style={show grid/keep bb}}%

\usepackage{multirow}
\usepackage{booktabs}
\usepackage{tabularx}
\newcolumntype{L}{>{\raggedright\arraybackslash}X}

\usepackage{algorithm}
\usepackage{algorithmic}

\newtheorem{assumption}{Assumption}
\newcommand{\mbf}[1]{\mathbf{#1}}
\newcommand{\mc}[1]{\mathcal{#1}}
\newcommand{\mb}[1]{\mathbb{#1}}

\makeatletter
\DeclareRobustCommand*\cal{\@fontswitch\relax\mathcal}
\makeatother

\newcommand{\xai}{\textsc{XAI}\xspace}
\newcommand{\gradcam}{\textsc{g}rad-\textsc{cam}\xspace}
\newcommand{\lime}{\textsc{lime}\xspace}
\newcommand{\plainshap}{\textsc{shap}\xspace}
\newcommand{\shap}{\textsc{g}radient\plainshap\xspace}

\newcommand{\rex}{\textsc{r}e\textsc{x}\xspace}

\newcommand{\AD}{\textcolor{ad_color}{ConvAD}\xspace}
\newcommand{\Min}{\textcolor{min_color}{Min}\xspace}
\newcommand{\Max}{\textcolor{max_color}{Max}\xspace}
\newcommand{\Avg}{\textcolor{avg_color}{Avg}\xspace}
\newcommand{\Zero}{\textcolor{zero_color}{Zero}\xspace}

\newcommand{\LM}{\textcolor{lm_color}{LayerMask}\xspace}
\newcommand{\LMA}{\textcolor{lma_color}{LayerMask (Ablated)}\xspace}

\newcommand{\ood}{\emph{o.o.d.}\xspace}
\newcommand{\ad}{\textsc{AD}\xspace}
\newcommand{\convad}{\textsc{c}onv\textsc{ad}\xspace}

\newcommand{\commentout}[1]{}

\newcommand{\cs}{{\cal CS}}

\newcommand{\U}{{\cal U}}

\newcommand{\cF}{{\cal F}}
\newcommand{\V}{{\cal V}}

\newcommand{\K}{{\cal K}}

\newcommand{\sat}{\models}

\newcommand{\lem}{\begin{lemma}}
\newcommand{\elem}{\end{lemma}}
\newcommand{\pro}{\begin{proposition}}
\newcommand{\epro}{\end{proposition}}

\newcommand{\dfn}{\begin{definition}}
\newcommand{\edfn}{\end{definition}}

\newcommand{\xam}{\begin{example}}
\newcommand{\exam}{\end{example}}

\usepackage{algorithm}
\usepackage{algorithmic}

\usepackage{longtable}
\usepackage{subcaption}
\usepackage{pgfplots}
\usepackage{pgfplotstable}
\usepackage{paralist}
\usepgfplotslibrary{fillbetween}
\pgfplotsset{compat=1.16}

%% file: block_diagram.tex
\definecolor{darkblue}{HTML}{1f4e79}
\definecolor{lightblue}{HTML}{00b0f0}
\definecolor{salmon}{HTML}{ff9c6b}
\definecolor{applegreen}{rgb}{0.55, 0.71, 0.0}

\makeatletter
\pgfkeys{/pgf/.cd,
  parallelepiped offset x/.initial=2mm,
  parallelepiped offset y/.initial=2mm
}
\pgfdeclareshape{parallelepiped}
{
  \inheritsavedanchors[from=rectangle] 
  \inheritanchorborder[from=rectangle]
  \inheritanchor[from=rectangle]{north}
  \inheritanchor[from=rectangle]{north west}
  \inheritanchor[from=rectangle]{north east}
  \inheritanchor[from=rectangle]{center}
  \inheritanchor[from=rectangle]{west}
  \inheritanchor[from=rectangle]{east}
  \inheritanchor[from=rectangle]{mid}
  \inheritanchor[from=rectangle]{mid west}
  \inheritanchor[from=rectangle]{mid east}
  \inheritanchor[from=rectangle]{base}
  \inheritanchor[from=rectangle]{base west}
  \inheritanchor[from=rectangle]{base east}
  \inheritanchor[from=rectangle]{south}
  \inheritanchor[from=rectangle]{south west}
  \inheritanchor[from=rectangle]{south east}
  \backgroundpath{
    \southwest \pgf@xa=\pgf@x \pgf@ya=\pgf@y
    \northeast \pgf@xb=\pgf@x \pgf@yb=\pgf@y
    \pgfmathsetlength\pgfutil@tempdima{\pgfkeysvalueof{/pgf/parallelepiped
      offset x}}
    \pgfmathsetlength\pgfutil@tempdimb{\pgfkeysvalueof{/pgf/parallelepiped
      offset y}}
    \def\ppd@offset{\pgfpoint{\pgfutil@tempdima}{\pgfutil@tempdimb}}
    \pgfpathmoveto{\pgfqpoint{\pgf@xa}{\pgf@ya}}
    \pgfpathlineto{\pgfqpoint{\pgf@xb}{\pgf@ya}}
    \pgfpathlineto{\pgfqpoint{\pgf@xb}{\pgf@yb}}
    \pgfpathlineto{\pgfqpoint{\pgf@xa}{\pgf@yb}}
    \pgfpathclose
    \pgfpathmoveto{\pgfqpoint{\pgf@xb}{\pgf@ya}}
    \pgfpathlineto{\pgfpointadd{\pgfpoint{\pgf@xb}{\pgf@ya}}{\ppd@offset}}
    \pgfpathlineto{\pgfpointadd{\pgfpoint{\pgf@xb}{\pgf@yb}}{\ppd@offset}}
    \pgfpathlineto{\pgfpointadd{\pgfpoint{\pgf@xa}{\pgf@yb}}{\ppd@offset}}
    \pgfpathlineto{\pgfqpoint{\pgf@xa}{\pgf@yb}}
    \pgfpathmoveto{\pgfqpoint{\pgf@xb}{\pgf@yb}}
    \pgfpathlineto{\pgfpointadd{\pgfpoint{\pgf@xb}{\pgf@yb}}{\ppd@offset}}
  }
}
\makeatother

\tikzset{
  block/.style={
    parallelepiped,fill=white, draw,
    minimum width=0.8cm,
    minimum height=0.05cm,
    parallelepiped offset x=0.3cm,
    parallelepiped offset y=0.3cm,
    path picture={
      \draw[top color=applegreen,bottom color=applegreen]
        (path picture bounding box.south west) rectangle 
        (path picture bounding box.north east);
    },
    text=white,
  },
  conv/.style={
    parallelepiped,fill=white, draw,
    minimum width=0.01cm,
    minimum height=2cm,
    parallelepiped offset x=0.3cm,
    parallelepiped offset y=0.3cm,
    path picture={
      \draw[top color=salmon,bottom color=salmon]
        (path picture bounding box.south west) rectangle 
        (path picture bounding box.north east);
    },
    text=white,
  },
  plate/.style={
    parallelepiped,fill=white, draw,
    minimum width=0.1cm,
    minimum height=1cm,
    parallelepiped offset x=0.3cm,
    parallelepiped offset y=0.3cm,
    path picture={
      \draw[top color=lightblue,bottom color=lightblue]
        (path picture bounding box.south west) rectangle 
        (path picture bounding box.north east);
    },
    text=white,
  },
   occ/.style={
    parallelepiped,fill=white, draw,
    minimum width=0.1cm,
    minimum height=0.2cm,
    parallelepiped offset x=0.17cm,
    parallelepiped offset y=0.17cm,
    path picture={
      \draw[top color=black!60,bottom color=black!60]
        (path picture bounding box.south west) rectangle 
        (path picture bounding box.north east);
    },
    text=white,
  },
  link/.style={
    color=lightblue,
    line width=1.5mm,
  },
}

\begin{figure}[t]
\centering
\resizebox{0.8\linewidth}{!}{
\begin{tikzpicture}
    \node[inner sep=0pt,label=above:Starfish] (Starfish) at (-10, 0) {
    \includegraphics[width=2cm, height=2cm]{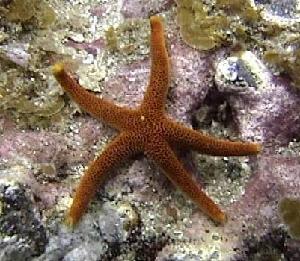}
    };

    \begin{scope}[scale=.2, xshift=-43cm, yshift=-5cm]
        \draw (0, 0) grid (10, 10);
        \fill[black] (0,9) rectangle ++(7,1);
        \fill[black] (0,8) rectangle ++(7,1);
        \fill[black] (0,7) rectangle ++(5,1);
        \fill[black] (0,6) rectangle ++(5,1);
        \node at (5, 11) {Mask};
    \end{scope}

    \node[fill=zero_color!15,draw=zero_color,thick,rectangle,rounded corners,
        minimum width=8cm, minimum height=2.8cm,
        label=above:\ding{172} Image Perturbation (SOTA $\rightarrow$ OOD)] (sota) at (-2.3,1.7) {};

    \draw[-Triangle,very thick] (-10, 1.5) |- (-6, 2);
    \draw[-Triangle,very thick] (-7.5, 1.5) |- (-6, 1.7);
    \draw[-Triangle,very thick] (-4.5, 1.7) |- (-4, 1.7);

    \node (edit) at (-5.25, 1.7) {
    \includegraphics[width=1.4cm, height=1.4cm]{Figures/starfish.jpg}
    };
    \begin{scope}[scale=.14, xshift=-42.5cm,yshift=7.1cm]
        \draw (0, 0) grid (10, 10);
        \fill[black] (0,9) rectangle ++(7,1);
        \fill[black] (0,8) rectangle ++(7,1);
        \fill[black] (0,7) rectangle ++(5,1);
        \fill[black] (0,6) rectangle ++(5,1);
    \end{scope}

    \node[conv,right=0.5cm of edit,yshift=-0.4em](conv1){};
    \node[conv,right=0.1cm of conv1](conv2){};
    \node[plate,right=0.4cm of conv2](plate1){};
    \node[conv, minimum height=1cm, right=0.1cm of plate1](conv3){};
    \node[conv, minimum height=1cm, right=0.1cm of conv3](conv4){};

    \node[plate, minimum height=0.5cm,right=0.4cm of conv4](plate2){};
    \node[conv, minimum height=0.5cm, right=0.1cm of plate2](conv5){};
    \node[conv, minimum height=0.5cm, right=0.1cm of conv5](conv6){};

    \node[block,right=0.4cm of conv6](connected){};

    \draw[-,link] ([xshift=0.2cm,yshift=0.05cm]conv2.east) -- ([yshift=0.05cm]plate1.west);
    \draw[-,link] ([xshift=0.2cm,yshift=0.05cm]conv4.east) -- ([yshift=0.05cm]plate2.west);
    \draw[-,link] ([xshift=0.1cm,yshift=0.01cm]conv6.east) -- ([yshift=0.01cm]connected.west);
    \draw[-{Latex[length=5mm,width=3mm]},link] ([xshift=0.1cm,yshift=0.05cm]connected.east) -- ([xshift=0.9cm,yshift=0.05cm]connected.east);

    \node[fill=ad_color!5,draw=ad_color,thick,rectangle,rounded corners,
        minimum width=8cm, minimum height=2.8cm,
        label=above:\ding{173} Deactivate Nodes with Mask (\ad)] (AD) at (-2.3,-2) {};

    \draw[-Triangle,very thick,ad_color] (-10, -1.2) |- (-6, -2.3);
    \draw[-Triangle,very thick,ad_color] (-7.5, -1.2) |- (-4, -1.4);
    \draw[-Triangle,very thick,ad_color] (-4.5, -2.3) |- (-4, -2.3);

    \node (edit2) at (-5.25, -2.3) {
        \includegraphics[width=1.4cm, height=1.4cm]{Figures/starfish.jpg}
    };

    \node[conv,right=0.5cm of edit2,yshift=0.3em](conv1){};
    \node[occ,right=0.5cm of edit2,yshift=2.8em]{};
    \node[conv,right=0.1cm of conv1](conv2){};
    \node[occ,right=0.1cm of conv1,yshift=2.5em]{};
    \node[plate,right=0.4cm of conv2](plate1){};
    \node[occ,right=0.4cm of conv2,yshift=1.1em]{};

    \node[conv, minimum height=1cm, right=0.1cm of plate1](conv3){};
    \node[occ, right=0.1cm of plate1, yshift=1.1em]{};
    \node[conv, minimum height=1cm, right=0.1cm of conv3](conv4){};
    \node[occ, right=0.1cm of conv3, yshift=1.1em]{};

    \node[plate, minimum height=0.5cm,right=0.4cm of conv4](plate2){};
    \node[occ, right=0.4cm of conv4]{};
    
    \node[conv, minimum height=0.5cm, right=0.1cm of plate2](conv5){};
    \node[occ, right=0.1cm of plate2]{};
    
    \node[conv, minimum height=0.5cm, right=0.1cm of conv5](conv6){};
    \node[occ, right=0.1cm of conv5]{};

    \node[block,right=0.4cm of conv6](connected){};
    
    \node[occ,minimum width=0.8cm,right=0.4cm of conv6]{};

    \draw[-,link] ([xshift=0.2cm,yshift=0.05cm]conv2.east) -- ([yshift=0.05cm]plate1.west);
    \draw[-,link] ([xshift=0.2cm,yshift=0.05cm]conv4.east) -- ([yshift=0.05cm]plate2.west);
    \draw[-,link] ([xshift=0.1cm,yshift=0.01cm]conv6.east) -- ([yshift=0.01cm]connected.west);
    \draw[-{Latex[length=5mm,width=3mm]},link] ([xshift=0.1cm,yshift=0.05cm]connected.east) -- ([xshift=0.9cm,yshift=0.05cm]connected.east);
    
\end{tikzpicture}}
\caption{\ding{172} The state-of-the-art approach to generating perturbations for black-box \emph{post hoc} explainability: masking parts of the input image. \ding{173}
    Activation-Deactivation (\ad): occluding parts of the \emph{model}.
    \ad preserves the spatial locality of the unmasked features while removing any consideration of the masked features, remaining \emph{in distribution} for the input features to the model.}%
    \label{fig:architecture}
\end{figure}

%% file: masking_examples.tex
%
%
%
\begin{figure}[t]
\centering
\captionsetup[subfigure]{labelformat=empty}
    \begin{subfigure}[t]{0.15\textwidth}
        \centering
        \caption{}
        \includegraphics[width=\linewidth]{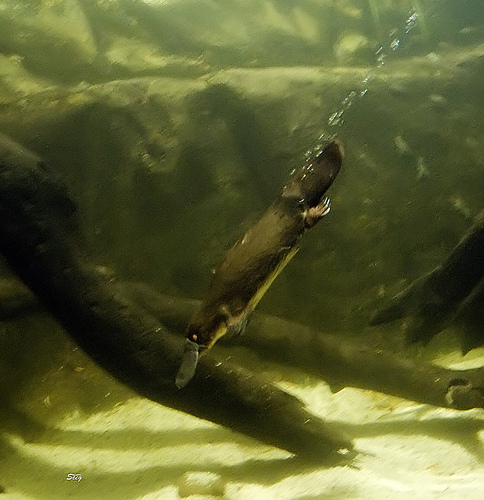}
        \caption{Platypus}
    \end{subfigure}
    \hfill
    \begin{subfigure}[t]{0.15\textwidth}
        \centering
        \caption{\AD}
        \includegraphics[width=\linewidth]{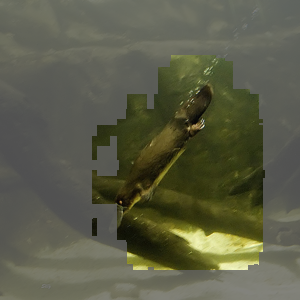}
    \end{subfigure}
    \hfill
    \begin{subfigure}[t]{0.15\textwidth}
        \centering
        \caption{\Min}
        \includegraphics[width=\linewidth]{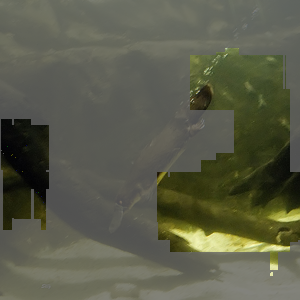}
    \end{subfigure}
    \hfill
    \begin{subfigure}[t]{0.15\textwidth}
        \centering
        \caption{\Zero}
        \includegraphics[width=\linewidth]{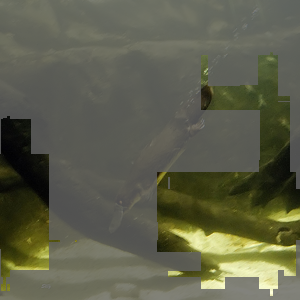}
    \end{subfigure}
    \hfill
    \begin{subfigure}[t]{0.15\textwidth}
        \centering
        \caption{\Avg}
        \includegraphics[width=\linewidth]{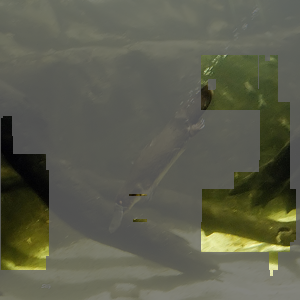}
    \end{subfigure}
    \hfill
    \begin{subfigure}[t]{0.15\textwidth}
        \centering
        \caption{\Max}
        \includegraphics[width=\linewidth]{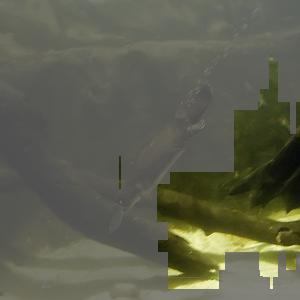}
    \end{subfigure}

    \begin{subfigure}[t]{0.15\textwidth}
        \centering
        \includegraphics[width=\linewidth, height=\linewidth]{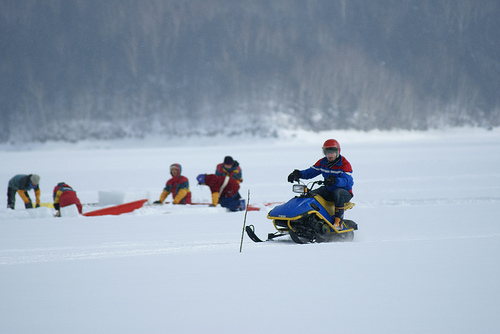}
\caption{Snow}
    \end{subfigure}
    \hfill
    \begin{subfigure}[t]{0.15\textwidth}
        \centering
        \includegraphics[width=\linewidth]{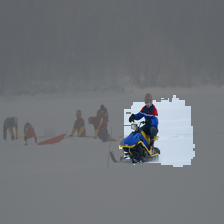}
    \end{subfigure}
    \hfill
    \begin{subfigure}[t]{0.15\textwidth}
        \centering
        \includegraphics[width=\linewidth]{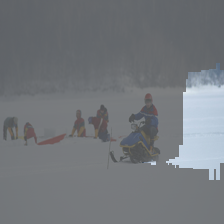}
    \end{subfigure}
    \hfill
    \begin{subfigure}[t]{0.15\textwidth}
        \centering
        \includegraphics[width=\linewidth]{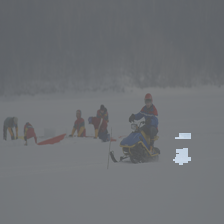}
    \end{subfigure}
    \hfill
    \begin{subfigure}[t]{0.15\textwidth}
        \centering
        \includegraphics[width=\linewidth]{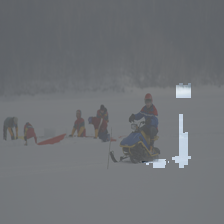}
    \end{subfigure}
    \hfill
    \begin{subfigure}[t]{0.15\textwidth}
        \centering
        \includegraphics[width=\linewidth]{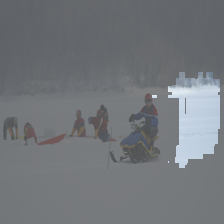}
    \end{subfigure}


    \begin{subfigure}[t]{0.15\textwidth}
        \centering
        \includegraphics[width=\linewidth, height=\linewidth]{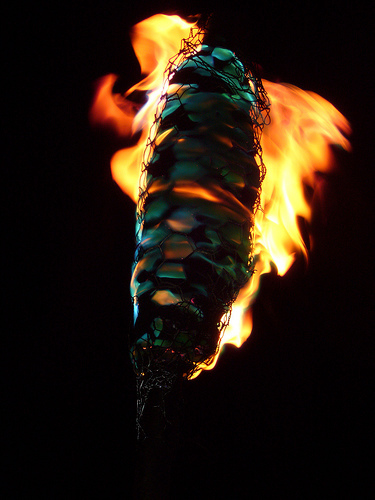}
        \caption{Torch}
    \end{subfigure}
    \hfill
    \begin{subfigure}[t]{0.15\textwidth}
        \centering
        \includegraphics[width=\linewidth]{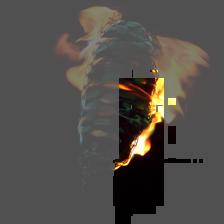}
    \end{subfigure}
    \hfill
    \begin{subfigure}[t]{0.15\textwidth}
        \centering
        \includegraphics[width=\linewidth]{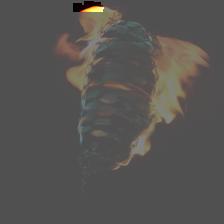}
    \end{subfigure}
    \hfill
    \begin{subfigure}[t]{0.15\textwidth}
        \centering
        \includegraphics[width=\linewidth]{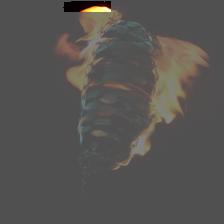}
    \end{subfigure}
    \hfill
    \begin{subfigure}[t]{0.15\textwidth}
        \centering
        \includegraphics[width=\linewidth]{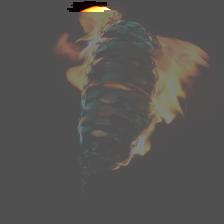}
    \end{subfigure}
    \hfill
    \begin{subfigure}[t]{0.15\textwidth}
        \centering
        \includegraphics[width=\linewidth]{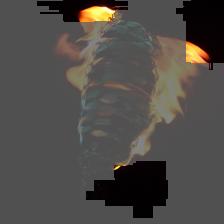}
    \end{subfigure}

    \begin{subfigure}[t]{0.15\textwidth}
        \centering
        \includegraphics[width=\linewidth, height=\linewidth]{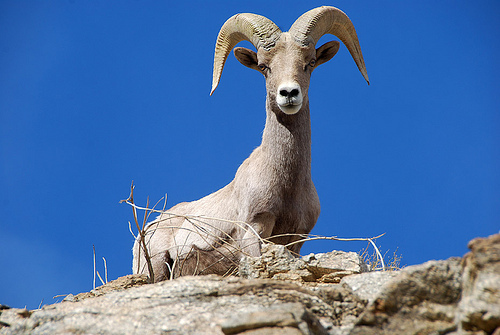}
        \caption{Sheep}
    \end{subfigure}
    \hfill
    \begin{subfigure}[t]{0.15\textwidth}
        \centering
        \includegraphics[width=\linewidth]{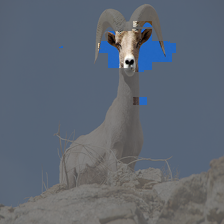}
    \end{subfigure}
    \hfill
    \hfill
    \begin{subfigure}[t]{0.15\textwidth}
        \centering
        \includegraphics[width=\linewidth]{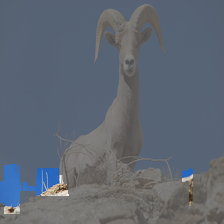}
    \end{subfigure}
    \hfill
    \begin{subfigure}[t]{0.15\textwidth}
        \centering
        \includegraphics[width=\linewidth]{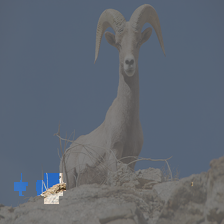}
    \end{subfigure}
    \hfill
    \begin{subfigure}[t]{0.15\textwidth}
        \centering
        \includegraphics[width=\linewidth]{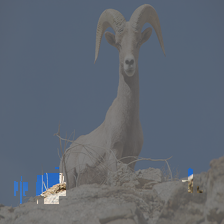}
    \end{subfigure}
    \hfill
    \begin{subfigure}[t]{0.15\textwidth}
        \centering
        \includegraphics[width=\linewidth]{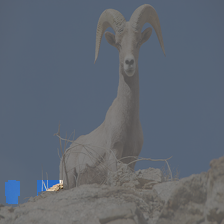}
    \end{subfigure}

\caption{
The choice of masking value can often have a profound effect on the sufficient pixels. An image of e.g. a platypus (top
row) and its sufficient pixel sets using \textcolor{ad_color}{\convad} (our approach) compared to occlusion-based methods with different
occlusion values (\Min-\Max) on the EfficientNet-v2 model. All the images in the top row are classified as platypus, though only
\textcolor{ad_color}{\convad} contains any platypus. All images and their explanations have been taken from the ImageNet-1k dataset.}\label{fig:grid_comparison}
\label{fig:masking_examples}
\end{figure}
%

%% file: Plots/updated_robustness_rex.tex
    \begin{subfigure}{0.32\linewidth}
        \centering
        \begin{tikzpicture}
            \begin{axis}[grid, ylabel=ResNet50, xticklabel=\empty, width=4.5cm, height=3cm, ymin=0, ymax=1, xmin=0, xmax=1,xtick={0, 0.25, 0.5, 0.75, 1},title=ImageNet-1k]
                \addplot[color=min_color, error bars/.cd, y dir=both, y explicit] 
                    coordinates {(0, 0.0968) +- (0, 0.0147) (0.2, 0.2210) +- (0, 0.0227) (0.4, 0.4314) +- (0, 0.0322) (0.6, 0.5561) +- (0, 0.0321) (0.8, 0.6678) +- (0, 0.0303) (1, 0.7843) +- (0, 0.0275)};
                \addplot[color=avg_color, error bars/.cd, y dir=both, y explicit]
                    coordinates {(0, 0.0207) +- (0, 0.0034) (0.2, 0.0748) +- (0, 0.0110) (0.4, 0.1976) +- (0, 0.0239) (0.6, 0.3058) +- (0, 0.0294) (0.8, 0.4218) +- (0, 0.0331) (1, 0.5490) +- (0, 0.0334)};
                \addplot[color=zero_color, error bars/.cd, y dir=both, y explicit]
                    coordinates {(0, 0.0228) +- (0, 0.0071) (0.2, 0.0628) +- (0, 0.0136) (0.4, 0.1578) +- (0, 0.0220) (0.6, 0.2579) +- (0, 0.0280) (0.8, 0.3648) +- (0, 0.0330) (1, 0.4912) +- (0, 0.0347)};
                \addplot[color=max_color, error bars/.cd, y dir=both, y explicit]
                    coordinates {(0, 0.2123) +- (0, 0.0254) (0.2, 0.3949) +- (0, 0.0284) (0.4, 0.5956) +- (0, 0.0302) (0.6, 0.6945) +- (0, 0.0292) (0.8, 0.7683) +- (0, 0.0290) (1, 0.8577) +- (0, 0.0216)};
                \addplot[color=ad_color, error bars/.cd, y dir=both, y explicit]
                    coordinates {(0, 0.2027) +- (0, 0.0245) (0.2, 0.3981) +- (0, 0.0289) (0.4, 0.6382) +- (0, 0.0318) (0.6, 0.7481) +- (0, 0.0296) (0.8, 0.8014) +- (0, 0.0278) (1, 0.8668) +- (0, 0.0218)};
                \addplot[color=lma_color, error bars/.cd, y dir=both, y explicit]
                    coordinates {(0, 0.1000) +- (0, 0.0001) (0.2, 0.2500) +- (0, 0.0200) (0.4, 0.3500) +- (0, 0.0405) (0.6, 0.4658) +- (0, 0.0441) (0.8, 0.5362) +- (0, 0.0441) (1, 0.8019) +- (0, 0.0332)};
                \addplot[color=lm_color, error bars/.cd, y dir=both, y explicit]
                    coordinates {(0, 0.0500) +- (0, 0.0013) (0.2, 0.1000) +- (0, 0.0158) (0.4, 0.1200) +- (0, 0.0245) (0.6, 0.1481) +- (0, 0.0257) (0.8, 0.2153) +- (0, 0.0300) (1, 0.3296) +- (0, 0.0344)};
            \end{axis}
        \end{tikzpicture}
    \end{subfigure}
    \quad
    \begin{subfigure}{0.27\linewidth}
        \centering
        \begin{tikzpicture}
            \begin{axis}[grid, yticklabel=\empty, xticklabel=\empty, width=4.5cm, height=3cm, ymin=0, ymax=1, xmin=0, xmax=1,xtick={0, 0.25, 0.5, 0.75, 1},title=ImageNet-v2]
                \addplot[color=min_color, error bars/.cd, y dir=both, y explicit] 
                    coordinates {(0, 0.0863) +- (0, 0.0135) (0.2, 0.2294) +- (0, 0.0236) (0.4, 0.4666) +- (0, 0.0310) (0.6, 0.6195) +- (0, 0.0324) (0.8, 0.7003) +- (0, 0.0306) (1, 0.7885) +- (0, 0.0275)};
                \addplot[color=avg_color, error bars/.cd, y dir=both, y explicit]
                    coordinates {(0, 0.0361) +- (0, 0.0029) (0.2, 0.0853) +- (0, 0.0108) (0.4, 0.2090) +- (0, 0.0225) (0.6, 0.3405) +- (0, 0.0301) (0.8, 0.4566) +- (0, 0.0325) (1, 0.5761) +- (0, 0.0328)};
                \addplot[color=zero_color, error bars/.cd, y dir=both, y explicit]
                    coordinates {(0, 0.0337) +- (0, 0.0062) (0.2, 0.0562) +- (0, 0.0091) (0.4, 0.1415) +- (0, 0.0193) (0.6, 0.2577) +- (0, 0.0279) (0.8, 0.3647) +- (0, 0.0329) (1, 0.4924) +- (0, 0.0349)};
                \addplot[color=max_color, error bars/.cd, y dir=both, y explicit]
                    coordinates {(0, 0.1787) +- (0, 0.0173) (0.2, 0.3606) +- (0, 0.0252) (0.4, 0.6260) +- (0, 0.0289) (0.6, 0.7353) +- (0, 0.0288) (0.8, 0.8047) +- (0, 0.0286) (1, 0.8569) +- (0, 0.0235)};
                \addplot[color=ad_color, error bars/.cd, y dir=both, y explicit]
                    coordinates {(0, 0.2930) +- (0, 0.0230) (0.2, 0.4242) +- (0, 0.0295) (0.4, 0.6752) +- (0, 0.0293) (0.6, 0.7754) +- (0, 0.0277) (0.8, 0.7938) +- (0, 0.0285) (1, 0.8750) +- (0, 0.0237)};
                \addplot[color=lma_color, error bars/.cd, y dir=both, y explicit]
                    coordinates {(0, 0.2283) +- (0, 0.0002) (0.2, 0.3060) +- (0, 0.0243) (0.4, 0.3478) +- (0, 0.0415) (0.6, 0.4455) +- (0, 0.0430) (0.8, 0.5164) +- (0, 0.0429) (1, 0.7683) +- (0, 0.0349)};
                \addplot[color=lm_color, error bars/.cd, y dir=both, y explicit]
                    coordinates {(0, 0.0533) +- (0, 0.0004) (0.2, 0.0702) +- (0, 0.0195) (0.4, 0.1173) +- (0, 0.0229) (0.6, 0.1742) +- (0, 0.0275) (0.8, 0.2248) +- (0, 0.0309) (1, 0.3739) +- (0, 0.0354)};
            \end{axis}
        \end{tikzpicture}
    \end{subfigure}
    \quad
    \begin{subfigure}{0.25\linewidth}
        \centering
        \begin{tikzpicture}
            \begin{axis}[grid, yticklabel=\empty,xticklabel=\empty, width=4.5cm, height=3cm, ymin=0, ymax=1, xmin=0, xmax=1,
            xtick={0, 0.25, 0.5, 0.75, 1}, title=PASCAL-VOC]
                \addplot[color=min_color, error bars/.cd, y dir=both, y explicit] 
                    coordinates {(0, 0.3915) +- (0, 0.0000) (0.2, 0.3915) +- (0, 0.0121) (0.4, 0.3993) +- (0, 0.0309) (0.6, 0.4661) +- (0, 0.0372) (0.8, 0.5758) +- (0, 0.0359) (1, 0.6559) +- (0, 0.0331)};
                \addplot[color=avg_color, error bars/.cd, y dir=both, y explicit]
                    coordinates {(0, 0.1866) +- (0, 0.0000) (0.2, 0.1866) +- (0, 0.0069) (0.4, 0.1918) +- (0, 0.0161) (0.6, 0.2190) +- (0, 0.0170) (0.8, 0.2906) +- (0, 0.0246) (1, 0.3798) +- (0, 0.0287)};
                \addplot[color=zero_color, error bars/.cd, y dir=both, y explicit]
                    coordinates {(0, 0.1383) +- (0, 0.0000) (0.2, 0.1383) +- (0, 0.0124) (0.4, 0.1497) +- (0, 0.0118) (0.6, 0.1866) +- (0, 0.0143) (0.8, 0.2363) +- (0, 0.0238) (1, 0.3312) +- (0, 0.0297)};
                \addplot[color=max_color, error bars/.cd, y dir=both, y explicit]
                    coordinates {(0, 0.2801) +- (0, 0.0000) (0.2, 0.2801) +- (0, 0.0149) (0.4, 0.3064) +- (0, 0.0212) (0.6, 0.3861) +- (0, 0.0256) (0.8, 0.4775) +- (0, 0.0274) (1, 0.5780) +- (0, 0.0265)};
                \addplot[color=ad_color, error bars/.cd, y dir=both, y explicit]
                    coordinates {(0, 0.3202) +- (0, 0.0000) (0.2, 0.3202) +- (0, 0.0250) (0.4, 0.3742) +- (0, 0.0307) (0.6, 0.4607) +- (0, 0.0315) (0.8, 0.5523) +- (0, 0.0315) (1, 0.5913) +- (0, 0.0314)};
                \addplot[color=lma_color, error bars/.cd, y dir=both, y explicit]
                    coordinates {(0, 0.0000) +- (0, 0.0668) (0.2, 0.0000) +- (0, 0.0000) (0.4, 0.0000) +- (0, 0.0000) (0.6, 0.0165) +- (0, 0.0115) (0.8, 0.0162) +- (0, 0.0114) (1, 0.0700) +- (0, 0.0204)};
                \addplot[color=lm_color, error bars/.cd, y dir=both, y explicit]
                    coordinates {(0, 0.0000) +- (0, 0.0682) (0.2, 0.0033) +- (0, 0.0076) (0.4, 0.0323) +- (0, 0.0131) (0.6, 0.0336) +- (0, 0.0126) (0.8, 0.0545) +- (0, 0.0175) (1, 0.1116) +- (0, 0.0226)};
            \end{axis}
        \end{tikzpicture}
    \end{subfigure}

    \begin{subfigure}{0.32\textwidth}
        \centering
        \begin{tikzpicture}
            \begin{axis}[grid, ylabel=RegNetY,xticklabel=\empty,width=4.5cm,height=3cm, ymin=0, ymax=1, xmin=0, xmax=1,xtick={0, 0.25, 0.5, 0.75, 1}]
                \addplot[color=min_color, error bars/.cd, y dir=both, y explicit] 
                    coordinates {(0, 0.2888) +- (0, 0.0236) (0.2, 0.3945) +- (0, 0.0299) (0.4, 0.5762) +- (0, 0.0312) (0.6, 0.6998) +- (0, 0.0314) (0.8, 0.7979) +- (0, 0.0277) (1, 0.8945) +- (0, 0.0175)};
                \addplot[color=avg_color, error bars/.cd, y dir=both, y explicit]
                    coordinates {(0, 0.1370) +- (0, 0.0102) (0.2, 0.1855) +- (0, 0.0180) (0.4, 0.2847) +- (0, 0.0240) (0.6, 0.4050) +- (0, 0.0287) (0.8, 0.5199) +- (0, 0.0291) (1, 0.6962) +- (0, 0.0263)};
                \addplot[color=zero_color, error bars/.cd, y dir=both, y explicit]
                    coordinates {(0, 0.0995) +- (0, 0.0113) (0.2, 0.1269) +- (0, 0.0161) (0.4, 0.2177) +- (0, 0.0220) (0.6, 0.3188) +- (0, 0.0267) (0.8, 0.3945) +- (0, 0.0309) (1, 0.5503) +- (0, 0.0314)};
                \addplot[color=max_color, error bars/.cd, y dir=both, y explicit]
                    coordinates {(0, 0.2914) +- (0, 0.0209) (0.2, 0.4200) +- (0, 0.0289) (0.4, 0.5998) +- (0, 0.0308) (0.6, 0.7299) +- (0, 0.0280) (0.8, 0.8265) +- (0, 0.0271) (1, 0.8911) +- (0, 0.0164)};
                \addplot[color=ad_color, error bars/.cd, y dir=both, y explicit]
                    coordinates {(0, 0.6544) +- (0, 0.0351) (0.2, 0.7519) +- (0, 0.0368) (0.4, 0.8028) +- (0, 0.0350) (0.6, 0.8568) +- (0, 0.0288) (0.8, 0.8998) +- (0, 0.0249) (1, 0.9319) +- (0, 0.0201)};
                \addplot[color=lma_color, error bars/.cd, y dir=both, y explicit]
                    coordinates {(0, 0.1500) +- (0, 0.0412) (0.2, 0.3000) +- (0, 0.0383) (0.4, 0.4500) +- (0, 0.0379) (0.6, 0.6000) +- (0, 0.0367) (0.8, 0.7000) +- (0, 0.0386) (1, 0.9350) +- (0, 0.0169)};
                \addplot[color=lm_color, error bars/.cd, y dir=both, y explicit]
                    coordinates {(0, 0.0800) +- (0, 0.0358) (0.2, 0.1500) +- (0, 0.0351) (0.4, 0.2000) +- (0, 0.0328) (0.6, 0.3000) +- (0, 0.0314) (0.8, 0.4000) +- (0, 0.0270) (1, 0.8887) +- (0, 0.0227)};
            \end{axis}
        \end{tikzpicture}
    \end{subfigure}
    \quad
    \begin{subfigure}{0.27\textwidth}
        \centering
        \begin{tikzpicture}
            \begin{axis}[grid, yticklabel=\empty,xticklabel=\empty,width=4.5cm,height=3cm, ymin=0, ymax=1, xmin=0, xmax=1,xtick={0, 0.25, 0.5, 0.75, 1}]
                \addplot[color=min_color, error bars/.cd, y dir=both, y explicit] 
                    coordinates {(0, 0.2681) +- (0, 0.0224) (0.2, 0.3887) +- (0, 0.0303) (0.4, 0.5963) +- (0, 0.0328) (0.6, 0.6913) +- (0, 0.0305) (0.8, 0.8243) +- (0, 0.0289) (1, 0.8783) +- (0, 0.0196)};
                \addplot[color=avg_color, error bars/.cd, y dir=both, y explicit]
                    coordinates {(0, 0.1332) +- (0, 0.0092) (0.2, 0.1864) +- (0, 0.0173) (0.4, 0.3100) +- (0, 0.0247) (0.6, 0.4205) +- (0, 0.0270) (0.8, 0.5383) +- (0, 0.0289) (1, 0.6749) +- (0, 0.0264)};
                \addplot[color=zero_color, error bars/.cd, y dir=both, y explicit]
                    coordinates {(0, 0.0758) +- (0, 0.0038) (0.2, 0.1087) +- (0, 0.0128) (0.4, 0.2154) +- (0, 0.0204) (0.6, 0.3104) +- (0, 0.0253) (0.8, 0.4081) +- (0, 0.0278) (1, 0.5330) +- (0, 0.0297)};
                \addplot[color=max_color, error bars/.cd, y dir=both, y explicit]
                    coordinates {(0, 0.2786) +- (0, 0.0215) (0.2, 0.3905) +- (0, 0.0283) (0.4, 0.6379) +- (0, 0.0304) (0.6, 0.7546) +- (0, 0.0278) (0.8, 0.8589) +- (0, 0.0234) (1, 0.8905) +- (0, 0.0179)};
                \addplot[color=ad_color, error bars/.cd, y dir=both, y explicit]
                    coordinates {(0, 0.6065) +- (0, 0.0371) (0.2, 0.7039) +- (0, 0.0357) (0.4, 0.7939) +- (0, 0.0304) (0.6, 0.8466) +- (0, 0.0314) (0.8, 0.8747) +- (0, 0.0273) (1, 0.9174) +- (0, 0.0192)};
                \addplot[color=lma_color, error bars/.cd, y dir=both, y explicit]
                    coordinates {(0, 0.1000) +- (0, 0.0421) (0.2, 0.2000) +- (0, 0.0380) (0.4, 0.3500) +- (0, 0.0370) (0.6, 0.5000) +- (0, 0.0357) (0.8, 0.6000) +- (0, 0.0344) (1, 0.8000) +- (0, 0.0158)};
                \addplot[color=lm_color, error bars/.cd, y dir=both, y explicit]
                    coordinates {(0, 0.0500) +- (0, 0.0364) (0.2, 0.1000) +- (0, 0.0351) (0.4, 0.1500) +- (0, 0.0330) (0.6, 0.2000) +- (0, 0.0285) (0.8, 0.3000) +- (0, 0.0300) (1, 0.6000) +- (0, 0.0221)};
            \end{axis}
        \end{tikzpicture}
    \end{subfigure} 
    \quad
    \begin{subfigure}{0.25\textwidth}
        \centering
        \begin{tikzpicture}
            \begin{axis}[grid, yticklabel=\empty,xticklabel=\empty,width=4.5cm,height=3cm, ymin=0, ymax=1, xmin=0, xmax=1,xtick={0, 0.25, 0.5, 0.75, 1}]
                \addplot[color=min_color, error bars/.cd, y dir=both, y explicit] 
                    coordinates {(0, 0.1504) +- (0, 0.0000) (0.2, 0.1504) +- (0, 0.0141) (0.4, 0.1934) +- (0, 0.0185) (0.6, 0.2705) +- (0, 0.0226) (0.8, 0.4079) +- (0, 0.0254) (1, 0.5813) +- (0, 0.0259)};
                \addplot[color=avg_color, error bars/.cd, y dir=both, y explicit]
                    coordinates {(0, 0.1409) +- (0, 0.0000) (0.2, 0.1409) +- (0, 0.0120) (0.4, 0.1456) +- (0, 0.0151) (0.6, 0.1960) +- (0, 0.0184) (0.8, 0.2891) +- (0, 0.0214) (1, 0.4360) +- (0, 0.0261)};
                \addplot[color=zero_color, error bars/.cd, y dir=both, y explicit]
                    coordinates {(0, 0.1332) +- (0, 0.0000) (0.2, 0.1332) +- (0, 0.0154) (0.4, 0.1559) +- (0, 0.0141) (0.6, 0.2050) +- (0, 0.0175) (0.8, 0.3045) +- (0, 0.0213) (1, 0.4813) +- (0, 0.0251)};
                \addplot[color=max_color, error bars/.cd, y dir=both, y explicit]
                    coordinates {(0, 0.1965) +- (0, 0.0144) (0.2, 0.1935) +- (0, 0.0161) (0.4, 0.2579) +- (0, 0.0204) (0.6, 0.3627) +- (0, 0.0234) (0.8, 0.4809) +- (0, 0.0247) (1, 0.6255) +- (0, 0.0232)};
                \addplot[color=ad_color, error bars/.cd, y dir=both, y explicit]
                    coordinates {(0, 0.8191) +- (0, 0.0000) (0.2, 0.8191) +- (0, 0.0347) (0.4, 0.8417) +- (0, 0.0416) (0.6, 0.8515) +- (0, 0.0346) (0.8, 0.8637) +- (0, 0.0291) (1, 0.8748) +- (0, 0.0278)};
                \addplot[color=lma_color, error bars/.cd, y dir=both, y explicit]
                    coordinates {(0, 0.0000) +- (0, 0.0000) (0.2, 0.0000) +- (0, 0.0275) (0.4, 0.0000) +- (0, 0.0338) (0.6, 0.0200) +- (0, 0.0365) (0.8, 0.0500) +- (0, 0.0386) (1, 0.1000) +- (0, 0.0413)};
                \addplot[color=lm_color, error bars/.cd, y dir=both, y explicit]
                    coordinates {(0, 0.0000) +- (0, 0.0000) (0.2, 0.0000) +- (0, 0.0384) (0.4, 0.0200) +- (0, 0.0421) (0.6, 0.0400) +- (0, 0.0338) (0.8, 0.0800) +- (0, 0.0305) (1, 0.1500) +- (0, 0.0280)};
            \end{axis}
        \end{tikzpicture}
    \end{subfigure}

    \begin{subfigure}{0.32\textwidth}
        \centering
        \begin{tikzpicture}
            \begin{axis}[grid, xlabel=Conf. threshold, ylabel=EfficientNet,width=4.5cm, height=3cm, ymin=0, ymax=1, xmin=0, xmax=1,xtick={0, 0.25, 0.5, 0.75, 1}]
                \addplot[color=min_color, error bars/.cd, y dir=both, y explicit] 
                    coordinates {(0, 0.2176) +- (0, 0.0126) (0.2, 0.2237) +- (0, 0.0205) (0.4, 0.2836) +- (0, 0.0271) (0.6, 0.3605) +- (0, 0.0286) (0.8, 0.4576) +- (0, 0.0318) (1, 0.5973) +- (0, 0.0328)};
                \addplot[color=avg_color, error bars/.cd, y dir=both, y explicit]
                    coordinates {(0, 0.2367) +- (0, 0.0103) (0.2, 0.2401) +- (0, 0.0184) (0.4, 0.3198) +- (0, 0.0235) (0.6, 0.4124) +- (0, 0.0260) (0.8, 0.5253) +- (0, 0.0274) (1, 0.6844) +- (0, 0.0278)};
                \addplot[color=zero_color, error bars/.cd, y dir=both, y explicit]
                    coordinates {(0, 0.2615) +- (0, 0.0129) (0.2, 0.2666) +- (0, 0.0210) (0.4, 0.3542) +- (0, 0.0256) (0.6, 0.4513) +- (0, 0.0271) (0.8, 0.5764) +- (0, 0.0281) (1, 0.6888) +- (0, 0.0260)};
                \addplot[color=max_color, error bars/.cd, y dir=both, y explicit]
                    coordinates {(0, 0.1736) +- (0, 0.0078) (0.2, 0.1742) +- (0, 0.0134) (0.4, 0.2253) +- (0, 0.0203) (0.6, 0.3173) +- (0, 0.0268) (0.8, 0.4551) +- (0, 0.0295) (1, 0.5927) +- (0, 0.0302)};
                \addplot[color=ad_color, error bars/.cd, y dir=both, y explicit]
                    coordinates {(0, 0.5534) +- (0, 0.0180) (0.2, 0.5553) +- (0, 0.0367) (0.4, 0.6075) +- (0, 0.0347) (0.6, 0.6673) +- (0, 0.0320) (0.8, 0.7572) +- (0, 0.0298) (1, 0.8256) +- (0, 0.0282)};
                \addplot[color=lma_color, error bars/.cd, y dir=both, y explicit]
                    coordinates {(0, 0.1200) +- (0, 0.0620) (0.2, 0.2800) +- (0, 0.0374) (0.4, 0.4200) +- (0, 0.0358) (0.6, 0.5500) +- (0, 0.0395) (0.8, 0.6500) +- (0, 0.0414) (1, 0.7800) +- (0, 0.0390)};
                \addplot[color=lm_color, error bars/.cd, y dir=both, y explicit]
                    coordinates {(0, 0.0600) +- (0, 0.0056) (0.2, 0.1200) +- (0, 0.0208) (0.4, 0.1800) +- (0, 0.0295) (0.6, 0.2500) +- (0, 0.0310) (0.8, 0.3500) +- (0, 0.0337) (1, 0.4500) +- (0, 0.0357)};
            \end{axis}
        \end{tikzpicture}
    \end{subfigure}
    \quad
    \begin{subfigure}{0.27\textwidth}
        \centering
        \begin{tikzpicture}
            \begin{axis}[grid, xlabel=Conf. Threshold, yticklabel=\empty,width=4.5cm,height=3cm, ymin=0, ymax=1, xmin=0, xmax=1,xtick={0, 0.25, 0.5, 0.75, 1}]
                \addplot[color=min_color, error bars/.cd, y dir=both, y explicit] 
                    coordinates {(0, 0.2474) +- (0, 0.0199) (0.2, 0.2491) +- (0, 0.0270) (0.4, 0.3339) +- (0, 0.0305) (0.6, 0.4576) +- (0, 0.0321) (0.8, 0.5408) +- (0, 0.0334) (1, 0.6457) +- (0, 0.0326)};
                \addplot[color=avg_color, error bars/.cd, y dir=both, y explicit]
                    coordinates {(0, 0.2482) +- (0, 0.0141) (0.2, 0.2586) +- (0, 0.0174) (0.4, 0.3536) +- (0, 0.0248) (0.6, 0.4836) +- (0, 0.0278) (0.8, 0.6024) +- (0, 0.0291) (1, 0.7429) +- (0, 0.0268)};
                \addplot[color=zero_color, error bars/.cd, y dir=both, y explicit]
                    coordinates {(0, 0.2374) +- (0, 0.0132) (0.2, 0.2466) +- (0, 0.0168) (0.4, 0.3556) +- (0, 0.0233) (0.6, 0.4933) +- (0, 0.0251) (0.8, 0.6493) +- (0, 0.0279) (1, 0.7678) +- (0, 0.0247)};
                \addplot[color=max_color, error bars/.cd, y dir=both, y explicit]
                    coordinates {(0, 0.1554) +- (0, 0.0104) (0.2, 0.1573) +- (0, 0.0168) (0.4, 0.2244) +- (0, 0.0218) (0.6, 0.2926) +- (0, 0.0240) (0.8, 0.4129) +- (0, 0.0293) (1, 0.5391) +- (0, 0.0306)};
                \addplot[color=ad_color, error bars/.cd, y dir=both, y explicit]
                    coordinates {(0, 0.5360) +- (0, 0.0296) (0.2, 0.5416) +- (0, 0.0374) (0.4, 0.6064) +- (0, 0.0347) (0.6, 0.7078) +- (0, 0.0316) (0.8, 0.7822) +- (0, 0.0317) (1, 0.8567) +- (0, 0.0261)};
                \addplot[color=lma_color, error bars/.cd, y dir=both, y explicit]
                    coordinates {(0, 0.1500) +- (0, 0.0559) (0.2, 0.2500) +- (0, 0.0305) (0.4, 0.3500) +- (0, 0.0328) (0.6, 0.4500) +- (0, 0.0359) (0.8, 0.5500) +- (0, 0.0390) (1, 0.6500) +- (0, 0.0390)};
                \addplot[color=lm_color, error bars/.cd, y dir=both, y explicit]
                    coordinates {(0, 0.0800) +- (0, 0.0240) (0.2, 0.1500) +- (0, 0.0242) (0.4, 0.2200) +- (0, 0.0297) (0.6, 0.3000) +- (0, 0.0333) (0.8, 0.4000) +- (0, 0.0340) (1, 0.5000) +- (0, 0.0358)};
            \end{axis}
        \end{tikzpicture}
    \end{subfigure}
    \quad
    \begin{subfigure}{0.25\textwidth}
        \centering
        \begin{tikzpicture}
            \begin{axis}[grid, xlabel=Conf. Threshold, yticklabel=\empty,width=4.5cm, height=3cm, ymin=0, ymax=1, xmin=0, xmax=1,xtick={0, 0.25, 0.5, 0.75, 1}]
                \addplot[color=min_color, error bars/.cd, y dir=both, y explicit] 
                    coordinates {(0, 0.3535) +- (0, 0.0000) (0.2, 0.3535) +- (0, 0.0233) (0.4, 0.3583) +- (0, 0.0233) (0.6, 0.4196) +- (0, 0.0280) (0.8, 0.4717) +- (0, 0.0306) (1, 0.5351) +- (0, 0.0290)};
                \addplot[color=avg_color, error bars/.cd, y dir=both, y explicit]
                    coordinates {(0, 0.2861) +- (0, 0.0000) (0.2, 0.2861) +- (0, 0.0205) (0.4, 0.3055) +- (0, 0.0218) (0.6, 0.3817) +- (0, 0.0267) (0.8, 0.4592) +- (0, 0.0288) (1, 0.5809) +- (0, 0.0278)};
                \addplot[color=zero_color, error bars/.cd, y dir=both, y explicit]
                    coordinates {(0, 0.2925) +- (0, 0.0000) (0.2, 0.2925) +- (0, 0.0186) (0.4, 0.3141) +- (0, 0.0234) (0.6, 0.3790) +- (0, 0.0255) (0.8, 0.4742) +- (0, 0.0269) (1, 0.6002) +- (0, 0.0275)};
                \addplot[color=max_color, error bars/.cd, y dir=both, y explicit]
                    coordinates {(0, 0.2898) +- (0, 0.0000) (0.2, 0.2898) +- (0, 0.0223) (0.4, 0.2959) +- (0, 0.0247) (0.6, 0.3259) +- (0, 0.0262) (0.8, 0.3506) +- (0, 0.0266) (1, 0.4473) +- (0, 0.0288)};
                \addplot[color=ad_color, error bars/.cd, y dir=both, y explicit]
                    coordinates {(0, 0.5910) +- (0, 0.0000) (0.2, 0.5910) +- (0, 0.0270) (0.4, 0.6179) +- (0, 0.0348) (0.6, 0.6901) +- (0, 0.0352) (0.8, 0.7440) +- (0, 0.0310) (1, 0.7981) +- (0, 0.0263)};
                \addplot[color=lma_color, error bars/.cd, y dir=both, y explicit]
                    coordinates {(0, 0.0000) +- (0, 0.0214) (0.2, 0.0000) +- (0, 0.0162) (0.4, 0.0000) +- (0, 0.0109) (0.6, 0.0100) +- (0, 0.0196) (0.8, 0.0300) +- (0, 0.0230) (1, 0.0800) +- (0, 0.0339)};
                \addplot[color=lm_color, error bars/.cd, y dir=both, y explicit]
                    coordinates {(0, 0.0000) +- (0, 0.0000) (0.2, 0.0000) +- (0, 0.0110) (0.4, 0.0100) +- (0, 0.0175) (0.6, 0.0200) +- (0, 0.0259) (0.8, 0.0500) +- (0, 0.0296) (1, 0.1200) +- (0, 0.0301)};
            \end{axis}
        \end{tikzpicture}
    \end{subfigure}

    \begin{tikzpicture}
        \centering
        \node[draw,color=black,rounded corners] (legend) { \textcolor{ad_color}{\textbf{ConvAD}} \quad \textcolor{min_color}{\textbf{Min}} \quad \textcolor{max_color}{\textbf{Max}} \quad \textcolor{avg_color}{\textbf{Avg}} \quad \textcolor{zero_color}{\textbf{Zero}} \quad \textcolor{lma_color}{\textbf{Layer Mask (Ablated)}} \quad \textcolor{lm_color}{\textbf{Layer Mask}} \quad};
    \end{tikzpicture}

%% file: related.tex
\section{Related Work}\label{sec:related}

\textbf{Explainability for AI models} There is a plethora of post-hoc explainability methods for DNNs, both white-box (model-specific) ones \citep{selvaraju2017grad,chattopadhay2018grad,LRP,deeplift,sundararajan2017axiomatic,springenberg2015striving,simonyan2013deep}, and black-box (model-agnostic)~\citep{CKKS26,lime,lundberg2017unified,petsiuk2018rise,ribeiro_anchors_2018,zeiler2014visualizing}. These methods calculate saliency landscapes, summarizing the contribution of different input features towards the model's decision. 
White-box methods calculate this attribution by analyzing the intermediate layers and
the weights of activation maps across different layers.
Black-box methods 
compute attribution using repeated calls to the model over perturbed variants of the inputs.

\textbf{Alternative perturbation strategies} Various methods have been propose to create mutants that more closely
reflect in-distribution samples. Generative in-painting is a popular method, specially in high-dimensional settings such
as images and complex signals. Previous methods utilized GANs and VAEs~\cite{shih2021ganmex, xiang2023stable},
however, newer methods rely on Diffusion models~\cite{ademi2025pomelo, solis2025diflime, jeanneret2022diffusion}.
Using generative methods, however, has the obvious problem of introducing yet another black-box model to the framework.
Optimization-based perturbations are another set
of methods, where the goal is to learn the optimal mask for a given instance. \cite{fong2017interpretable} 
is a seminal work in this domain, defining meaningful perturbations, formulating the problem of finding the smallest
mask $m$ that causes the score $f_c(\Phi(x_0;m))$ to drop significantly, and introducing blur as a masking strategy
to keep the mutants in distribution. 

\textbf{Metrics for attribution-based XAI methods} There is a number of metrics for evaluating the quality of
explanations in the literature, faithfulness (adherence to true decision-making
process)~\cite{zheng2024f,hooker2019benchmark,yeh2019fidelity}, robustness (stability against irrelevant
noise)~\cite{vascotto2024can,agarwal2022rethinking} and
plausibility/interpretability~\cite{wang2020scoreCAM,bohle2021convolutional,jacovi-goldberg-2020-towards}, to name
the most prominent ones. The latter one, human interpretability, is irrelevant if the goal of an explanation
is, as in this paper, to uncover the decision-making process of a model~\citet{jacovi-goldberg-2020-towards,bhusalface}.


%

%% file: appendix.tex
\newpage
\appendix
\numberwithin{equation}{section}
\numberwithin{figure}{section}
\numberwithin{table}{section}
\renewcommand{\thefigure}{\thesection\arabic{figure}}
\renewcommand{\thetable}{\thesection\arabic{table}}
\Crefname{appendix}{Sec.}{Secs.}

\newcommand{\additem}[2]{%
\item[\textbf{(\ref{#1})}]
    \textbf{#2} \dotfill\makebox{\textbf{\pageref{#1}}}
}

\newcommand{\addsubitem}[2]{%
    \textbf{(\ref{#1})}\hspace{1em}
    #2\\[.1em]
}

\begin{center}
{\Large\bf Appendix}
\end{center}

In this appendix to our work on Activation Deactivation, we provide:
\\[1em]

\begin{enumerate}
    \additem{appendix:gradcam_lime_robust_ood}{GradCAM and LIME robustness against \ood examples}\\
    In this section, we report the robustness against solid-color backgrounds for \gradcam and \lime, similar to the results reported in \cref{sec:exp_results} for \rex.
    \additem{appendix:robust_iid}{Robustness against \iid examples}\\
    In this section, we report the robustness scores of \convad vs masking values when using \iid examples as background.
    \additem{sup:sec:act_cause}{Background on Actual Causality} \\
    In this section, we provide a primer on Actual Causality, based on~\citep{Hal19}.
    \additem{sup:sec:equiv_proof}{Equivalence of \convad network to regular network in the unmasked case} \\
    In this section, we provide a proof to show that \convad network behaves exactly the same under normal inference condition as a model that does not contain the \convad mechanism.
    \additem{sup:sec:model_accuracy}{Accuracy of models on test sample}\\
    We report the accuracy of the different models across the test sample used from the different datasets.
    \additem{sup:sec:avg_size}{Average size of explanations}\\
    In this section, we provide the complete detail of the average explanation sizes across the different datasets and different models.
    \additem{sup:sec:conf_dropoff}{Confidence scores across different backgrounds}\\
    In this section, we report the average confidence across different backgrounds and different thresholds and report the confidence difference against the original background.
    \additem{appendix:Discoverability}{Discoverability of Explanations} \\
    In this section, we define the concept of \textit{discoverability}, the quantification of the ability of a strategy and XAI tool to discover a $\gamma-$confident explanation wth respect to a particular model and provide the results. \convad consistently has higher discoverability compared to LayerMask and has competitive results compared to the masking strategies. Thus, \convad provides competitive discoverability while being more robust and transferable explanations.
    \additem{appendix:overlap}{Overlap of Explanations}\\
    In this section, we report the DICE and IoU Scores of the calculated MSPSs, comparing Masking vs AD, Masking vs LayerMask and AD vs LayerMask for all models and datasets evaluated.
    
\end{enumerate}

\newpage

\section{GradCAM and LIME robustness scores against \ood examples}%
\label{appendix:gradcam_lime_robust_ood}
\input{Plots/updated_robustness_gradcam}
\input{Plots/updated_robustness_lime}

\FloatBarrier

\section{Robustness against IID examples}
\label{appendix:robust_iid}

\input{Plots/updated_robustness_rex_iid}
\input{Plots/updated_robustness_lime_iid}
\input{Plots/updated_robustness_gradcam_iid}

\FloatBarrier

\clearpage

\section{Background on Actual Causality}
\label{sup:sec:act_cause}
In what follows, we briefly introduce the relevant definitions from the theory of actual causality. The reader is referred
to~\cite{Hal19} for further reading.
We assume that the world is described in terms of
variables and their values.
Some variables may have a causal influence on others. This
influence is modeled by a set of {\em structural equations}.
It is conceptually useful to split the variables into two
sets: the {\em exogenous\/} variables $\U$, whose values are
determined by factors outside the model, and the {\em endogenous\/} variables $\V$, whose values are ultimately determined by
the exogenous variables.
The structural equations $\cF$ describe how these values are
determined. A \emph{causal model} $M$ is described by its variables and the structural
equations. We restrict the discussion to acyclic (recursive) causal models.
A \emph{context} $\vec{u}$ is a setting for the exogenous variables $\U$, which then
determines the values of all other variables.
We call a pair $(M,\vec{u})$ consisting of a causal model $M$ and a
context $\vec{u}$ a \emph{(causal) setting}.
An intervention is defined as setting the value of some
variable $X$ to $x$, and essentially amounts to replacing the equation for $X$
in $\cF$ by $X = x$.
A \emph{probabilistic causal model} is a pair $(M,\Pr)$, where $\Pr$ is a probability distribution on a given set of
contexts $\K$.

A causal formula $\psi$ is true or false in a setting.
We write $(M,\vec{u}) \sat \psi$  if
the causal formula $\psi$ is true in
the setting $(M,\vec{u})$.
Finally,
$(M,\vec{u}) \sat [\vec{Y} \gets \vec{y}]\varphi$ if
$(M_{\vec{Y} = \vec{y}},\vec{u}) \sat \varphi$,
where $M_{\vec{Y}\gets \vec{y}}$ is the causal model that is identical
to $M$, except that the
variables in $\vec{Y}$ are set to $Y = y$
for each $Y \in \vec{Y}$ and its corresponding
value $y \in \vec{y}$.

A standard use of causal models is to define \emph{actual causation}: that is,
what it means for some particular event that occurred to cause
another particular event.
There have been a number of definitions of actual causation given
for acyclic models
(\emph{e.g.}, \cite{beckers21c,GW07,Hall07,HP01b,Hal19,hitchcock:99,Hitchcock07,Weslake11,Woodward03}).
In this paper, we focus on the definitions
in~\cite{Hal19}, which we briefly review below.
The events that can be causes are arbitrary conjunctions of primitive
events.

\dfn[Actual cause]\label{def:AC}
$\vec{X} = \vec{x}$ is
an \emph{actual cause} of $\varphi$ in $(M,\vec{u})$ if the
following three conditions hold:
\begin{description}
\item[{\rm AC1.}]\label{ac1} $(M,\vec{u}) \models (\vec{X} = \vec{x})$ and $(M,\vec{u}) \models \varphi$.
\item[{\rm AC2.}] There is a
  a setting $\vec{x}'$ of the variables in $\vec{X}$, a
(possibly empty)  set $\vec{W}$ of variables in $\V - \vec{X}'$,
and a setting $\vec{w}$ of the variables in $\vec{W}$
such that $(M,\vec{u}) \models \vec{W} = \vec{w}$ and
$(M,\vec{u}) \models [\vec{X} \gets \vec{x}', \vec{W} \gets
    \vec{w}]\neg{\varphi}$, and moreover
\item[{\rm AC3.}] \label{ac3}
  $\vec{X}$ is minimal; there is no strict subset $\vec{X}'$ of
  $\vec{X}$ such that $\vec{X}' = \vec{x}''$ can replace $\vec{X} =
  \vec{x}'$ in
  AC2, where $\vec{x}''$ is the restriction of
$\vec{x}'$ to the variables in $\vec{X}'$.
\end{description}
\edfn
In the special case that $\vec{W} = \emptyset$, we get the
but-for definition. A variable $X$ in an actual cause $\vec{X}$ is called a \emph{part of a cause}. In what follows, following Halpern
we state that \emph{part of a cause is a cause}.
The \emph{degree of responsibility} quantifies actual causality~\cite{CH04,Hal19}.
Specifically, the degree of responsibility of a part of a cause $X=x$ in $\varphi$ is defined as $1/(|\vec{X}|+|\vec{W}|)$, where $X \in \vec{X}$,
$X$ is assigned $x$ in $\vec{x}$, and $\vec{X}=\vec{x}$ and $\vec{W}$ satisfy \Cref{def:AC}.

In this paper, we follow the construction of depth $2$ causal models with independent input variables,
introduced in~\cite{CH24} for black-box explainability of image classifiers. Essentially, we view the
neural network as a probabilistic causal model with a single internal node, the set of endogenous variables being the set
of the features of the input, and a single output node $O$.
The (unknown) equation for $O$ determines the output of the
neural network as a function of the input.
Thus, the causal network has depth $2$, with the exogenous variables
determining the feature variables, and the feature variables determining the output variable.
We assume \emph{causal independence} between the feature
variables $\vec{V}$; this is true in image classification and is a reasonable approximation in other domains we consider in this
paper. We also note that feature independence
is a common assumption in explainability tools.

It is easy to see that in this setting, in the condition AC2 of \Cref{def:AC}, the set $\vec{W}$ is empty. That is, we are
looking for a setting $\vec{x}'$ of the variables in $\vec{X}$ such that
$(M,\vec{u})\models[\vec{X} \gets \vec{x}']\neg{\varphi}$. Note also that the degree of responsibility of $X=x$ in these models is $1/|\vec{X}|$.

\section{Proof of Theorem~\ref{thm:equivalence}}
\label{sup:sec:equiv_proof}

\begin{assumption}
\label{assump:dim-change}
The dimensionality reductions/expansions are due to parametric operations or external additive/subtractive procedures to the intermediate representations.
\end{assumption}

For Theorem~\ref{thm:equivalence}, there are two cases that we need to account for, one of them being the case where the intermediate representations, $\mbf{z}$, changes in shape between successive convolutions in the model. The above assumption is used to prove the results for the broadest set of convolutional architectures, as parameterized alteration of the shape of $\mbf{z}$, is still dependent on the input features, and thus, is subject to AD. This setting covers most/all of the possible settings in which the shape of the intermediate representations, such as downsampling methods (max pooling, average pooling, global pooling etc.) or other feature-dependent reductions, upsampling methods or other feature-dependent expansions (such as replication of representations, interpolation) and external additions not informed by the representations (i.e. padding).

\paragraph{Equivalence to regular network in the unmasked case.} Given a CNN $\mc{N}$, we denote by $\mc{N'}$ the result of
application of \convad on $\mc{N}$. Then, $\mc{N'}$ performs identically to $\mc{N}$ for inputs without occlusions.


\begin{proof}
    Passing only an input image, $\mathbf{x}$, is equivalent to passing the tuple $(\mathbf{x},J_{m\times n})$, where $J_{m \times n}$ is a matrix of all ones and $\mathbf{x}, J_{m \times n} \in \mb{R}^{m \times n}$.

    The checkpoints in the \convad model, $\mc{N'}$, occur at least once between each two successive convolutional layers. We can separate the checkpoints to two  cases:

    \paragraph{Case 1:} $shape(\mc{N}_{(i)}(x)) = shape(\mc{N}_{(j-1)}(x))$ between successive convolutional layers, $\mc{N}_i$ and $\mc{N}_j$.\leavevmode

    This case holds for the set of parametric layers $\{\mc{N}_{i+1}...\mc{N}_{j-1}\}$ that do not alter the shape of the representations. Example of such operations include activation functions (ReLU, Tanh, GeLU etc.), normalization layers (Batch or Layer Normalization), regularization layers (Dropout) etc. Note, this case also holds if $\mc{N}_j$ is an immediate successor to $\mc{N}_i$ (i.e. $\mc{N}_{i} = \mc{N}_{j-1}$), which occurs in cases where we perform successive convolutions.

    Let $\mc{N}_{i..j-1} \in \mb{R}^{C \times K_{H} \times K_{w}}$, according to the AD definition, we require a position-attribution function at the checkpoint. For \convad, $\Phi_{j-1}(\cdot) \in \mb{R}^{1 \times d}$, it is defined as an accompanying mean-value convolution filter which we define at the checkpoint after layer $j-1$.

    The value of each element in $\Phi_{\mc{N}_{j-1}}$ is set to $\frac{1}{K_H \times K_W}$

    We attain the layer output, $O^{(i)}$, by convolving it by with convolution filter $\mc{N}^{(i)}$
    \[O^{(i)}_{n} [k] = (\mbf{z} * \mc{N}_{n}^{(i)}) [k] = \sum_{i=0}^{K_H - 1 }\sum_{j=0}^{K_W - 1 } \mbf{z}(h + i, w + j) \cdot \mc{N}_{n}^{(i)}(h,w) \]

    Where $O^{(i)}_{n}[k]$ is the $k^{th}$ entry of the $n^{th}$ channel of the output of the convolutional layer.

    After the convolution, all intermediate operations prior to the next convolution is performed wherein the dimensionality of the output is retained. Let $f: \mb{R}^{d} \to \mb{R}^d$ be the composition of these intermediate operations and therefore, we obtain the output till layer $j-1$:
    \[ O^{(j-1)} = f(O^{(i)})\]

    In parallel, we can calculate the output of the position attribution function for $\mc{N}'$, which in our case is the convolution of $J_{m \times n}$ with the filter $\Phi_{{j-1}}$.
    \begin{align*}
    {\Phi_{j-1}}(k,M) = (J_{m \times n} * \Phi_{j-1})[k] &= \sum_{i=0}^{K_H - 1 }\sum_{j=0}^{K_W - 1 } J_{m \times n}(h + i, w + j) \cdot {\Phi_{j-1}(h,w)} \\
    & = K_{H} \times K_{W} \cdot \left(1 \cdot \frac{1}{K_{H} \times K_W}\right) \\
    & = 1
    \end{align*}

    Thus we obtain $\Phi_{j-1}(M)$ such that all entries are 1 and $dim(\Phi_{j-1}(M)) = dim(O^{(j-1)}_{n})$. We then get the updated output by performing a Hadamard product between $O^{(j-1)}$ and $\Phi_{j-1}(M)$, which can be done by vectorizing $\Phi_{j-1}(M)$.
    \[\Phi_{{j-1}}^{\oplus}(M) \in \mb{R}^{dim(O^{(j-1)})} \]

    \[O'^{(j-1)} = O^{(j-1)} \odot \Phi_{j-1}^{\oplus}(M) = O^{(j-1)} \odot J_{dim(O^{(j-1)})} = O^{(j-1)} \]

    Which is equivalent to the output from the network $\mc{N}$ prior to the convolutional layer $\mc{N}_{(j)}$. It is clear that this is repeated at each checkpoint in $\mc{N}'$.

    \paragraph{Case 2:} $shape(\mc{N}_{i}(x)) \ne shape(\mc{N}_{j-1}(x))$ between successive convolutional layers, $\mc{N}_i$ and $\mc{N}_j$.\leavevmode

    This case occurs either when the set of parametric intermediate operations $\{\mc{N}_{i+1}...\mc{N}_{j-1}\}$ alter the dimension of the intermediate representation $\mbf{z}$ by reducing or expanding it, or via an independent external additive or subtractive process. This covers most, if not all possible settings where the shape of the representation is modified. Checkpoints are defined at each of the dimensionality altering positions.

    We first examine the case of checkpoints at parametric operations (i.e. downsampling methods such as pooling and upsampling methods such as interpolation).

    For such checkpoints, the tuple $(O^{(j-1)}, J_D), O^{(j-1)} \in \mb{R}^{N \times D}, J_D \in \mb{R}^{1 \times D} $ is the input representation and its corresponding mask before the dimension altering functions, this leads to:
    \[\mc{N}_{(j)}(O^{(j-1)}) = O^{(j)} \in \mb{R}^{N \times D'}\]
    Similar to the position attribution calculation in case $1$,
    \[\Phi_{j}(J_{D}) = J_{D'} \in \mb{R}^{1 \times D'}\]

    Therefore, similar to case $1$, the output for the \convad model $\mc{N}'^{(j)}$ is:

    \[O'^{(j)} = O^{(j)} \odot \Phi^\oplus_{j-1}(M) = O^{(j)} \odot \Phi^{\oplus}_{j}(J_{D'}) = O^{(j)}\]

    Which is equivalent to the output of the network $\mc{N}$.

    For the case of external additive effects, these are considered to be independent features and thus can be represented as an input. It can similarly be represented as a 2-tuple:
    \[(\mbf{z}'^{(j)}, J_{D'}), \mbf{z}'^{(j)} \in \mb{R}^{N \times D'}, J_{D'} \in \mb{R}^{1 \times D'}\]

    Where $\mbf{z'}^{(j)}$ is the set of representations that are going to be added to $O^{(j-1)}$. Given that the representations to be added are independent of the model's intermediate features, masking or unmasking its effects is an arbitrary decision. For the unmasked scenario, $J^{D'}$ is the accompanying mask for the external representations. We then apply the same combination function over representations and the binary masks.
    \[
        \mbf{z}'' = \text{combine}(O^{(j-1)}, \mbf{z'}), \mbf{z}'' \in \mb{R}^{N \times D''}
    \]
    \[
        M = J_{D''} = \text{combine}(J_{D}, J_{D'}), J_{D''} \in \mb{R}^{1 \times D''}
    \]

    For external subtractive procedures, the scenario is nearly identical, instead of the combine function, we have a reduce function.
    \[
        \mbf{z}'' = \text{reduce}(O^{(j-1)}), \mbf{z}'' \in \mb{R}^{N \times D''}
    \]
    \[
        M = J_{D''} = \text{reduce}(J_{D}), J_{D''} \in \mb{R}^{1 \times D''}
    \]

    The output of the regular network, $\mc{N}$ on the updated representations would be:
    \[\mc{N}_{(j)}(\mbf{z''}) = O^{(j)}\]

    The output of the AD network, $\mc{N}'$, is:
    \[O'^{(j)} = O^{(j)} \odot \Phi^{\oplus}_{j}(M) = O^{(j)} \odot \Phi^{\oplus}_{j}(J_{D''}) = O^{(j)}\]

    Which is equivalent to the output of the network $\mc{N}$.
\end{proof}

\clearpage

\section{Model Accuracy}
\label{sup:sec:model_accuracy}
\begin{table}[htbp]
  \centering
  \small
  \setlength{\tabcolsep}{6pt}
  \caption{Model accuracy}
  \label{tab:model_accuracy}
  \begin{tabular}{@{}lrrr@{}}
    \toprule
    Model & \multicolumn{3}{c}{Accuracy on sample set \% $\uparrow$} \\
    \cmidrule(l{3pt}r{3pt}){2-4}
     & IN-1k & IN-v2 & PASCAL-VOC \\
    \midrule
    ResNet-50         & 80.0 & 84.0 & 83.68 \\
    RegNetY-12GF      & 86.7 & 86.0 & 84.51 \\
    EfficientNet-V2-S & 86.0 & 86.0 & 82.61 \\
    \bottomrule
  \end{tabular}
\end{table}

\label{sup:sec:avg_size}

\FloatBarrier
\section{Average size of explanations}

 \begin{table}[htbp]
  \centering
  \small
  \setlength{\tabcolsep}{6pt}
  \caption{Average explanation size for ResNet on ImageNet-1k (ReX)}
  \label{tab:mask_coverage_imagenet_onek_resnet_rex}
  \begin{tabular}{@{}lrrrrrr@{}}
    \toprule
    Masking Value & \multicolumn{6}{c}{Threshold} \\
    \cmidrule(l{3pt}r{3pt}){2-7}
     & 0.9 & 0.7 & 0.5 & 0.3 & 0.1 & 0.0 \\
    \midrule
    Min & 20.27\% & 15.00\% & 11.79\% & 9.04\% & 6.45\% & 5.76\% \\
    Max & 21.84\% & 16.12\% & 12.60\% & 10.56\% & 7.92\% & 6.97\% \\
    Zero & 10.97\% & 8.46\% & 6.86\% & 5.31\% & 3.95\% & 2.91\% \\
    Avg & 13.04\% & 10.14\% & 8.18\% & 6.11\% & 4.31\% & 3.54\% \\
    \addlinespace[2pt]
    ConvAD & 23.78\% & 18.34\% & 15.12\% & 11.26\% & 8.02\% & 7.11\% \\
    LayerMask (Ablated) & 69.25\% & 68.65\% & 67.55\% & 67.29\% & 67.04\% & 66.49\% \\
    LayerMask & 7.66\% & 6.73\% & 6.17\% & 5.66\% & 5.39\% & 5.29\% \\
    \bottomrule
  \end{tabular}
\end{table}

\begin{table}[htbp]
  \centering
  \small
  \setlength{\tabcolsep}{6pt}
  \caption{Average explanation size for EfficientNet-V2 on ImageNet-1k (ReX)}
  \label{tab:mask_coverage_imagenet_onek_efn_v2_rex}
  \begin{tabular}{@{}lrrrrrr@{}}
    \toprule
    Masking Value & \multicolumn{6}{c}{Threshold} \\
    \cmidrule(l{3pt}r{3pt}){2-7}
     & 0.9 & 0.7 & 0.5 & 0.3 & 0.1 & 0.0 \\
    \midrule
    Min & 18.29\% & 12.93\% & 11.67\% & 10.80\% & 9.75\% & 9.66\% \\
    Max & 18.54\% & 14.08\% & 12.46\% & 11.28\% & 10.25\% & 10.25\% \\
    Zero & 17.40\% & 13.82\% & 11.85\% & 10.49\% & 9.50\% & 9.31\% \\
    Avg & 17.35\% & 13.34\% & 11.35\% & 10.51\% & 9.38\% & 9.28\% \\
    \addlinespace[2pt]
    ConvAD & 26.70\% & 20.51\% & 18.56\% & 16.90\% & 16.28\% & 16.26\% \\
    LayerMask (Ablated) & 73.67\% & 73.10\% & 72.54\% & 72.20\% & 71.94\% & 71.90\% \\
    LayerMask & 19.21\% & 15.37\% & 13.82\% & 12.76\% & 12.01\% & 11.91\% \\
    \bottomrule
  \end{tabular}
\end{table}

\begin{table}[htbp]
  \centering
  \small
  \setlength{\tabcolsep}{6pt}
  \caption{Average explanation size for RegNet on ImageNet-1k (ReX)}
  \label{tab:mask_coverage_imagenet_onek_regnet_rex}
  \begin{tabular}{@{}lrrrrrr@{}}
    \toprule
    Masking Value & \multicolumn{6}{c}{Threshold} \\
    \cmidrule(l{3pt}r{3pt}){2-7}
     & 0.9 & 0.7 & 0.5 & 0.3 & 0.1 & 0.0 \\
    \midrule
    Min & 10.16\% & 7.16\% & 5.89\% & 4.97\% & 4.17\% & 3.66\% \\
    Max & 9.68\% & 6.85\% & 5.68\% & 4.70\% & 3.93\% & 3.33\% \\
    Zero & 6.29\% & 4.33\% & 3.47\% & 2.90\% & 2.23\% & 1.83\% \\
    Avg & 6.70\% & 4.78\% & 3.92\% & 3.13\% & 2.54\% & 2.17\% \\
    \addlinespace[2pt]
    ConvAD & 15.92\% & 12.95\% & 11.27\% & 9.89\% & 8.76\% & 7.69\% \\
    LayerMask (Ablated) & 35.52\% & 31.98\% & 29.19\% & 26.23\% & 23.32\% & 20.35\% \\
    LayerMask & 15.11\% & 12.45\% & 10.73\% & 9.24\% & 7.79\% & 6.77\% \\
    \bottomrule
  \end{tabular}
\end{table}

\begin{table}[htbp]
  \centering
  \small
  \setlength{\tabcolsep}{6pt}
  \caption{Average explanation size for ResNet on ImageNet-1k (GradCAM)}
  \label{tab:mask_coverage_imagenet_onek_resnet_gradcam}
  \begin{tabular}{@{}lrrrrrr@{}}
    \toprule
    Masking Value & \multicolumn{6}{c}{Threshold} \\
    \cmidrule(l{3pt}r{3pt}){2-7}
     & 0.9 & 0.7 & 0.5 & 0.3 & 0.1 & 0.0 \\
    \midrule
    Min & 46.35\% & 35.28\% & 27.81\% & 21.30\% & 14.66\% & 13.28\% \\
    Max & 51.41\% & 39.30\% & 30.54\% & 22.22\% & 15.17\% & 12.10\% \\
    Zero & 27.54\% & 20.20\% & 15.64\% & 11.63\% & 9.31\% & 8.54\% \\
    Avg & 34.21\% & 24.01\% & 17.86\% & 13.63\% & 10.36\% & 9.01\% \\
    \addlinespace[2pt]
    ConvAD & 51.14\% & 41.85\% & 34.61\% & 27.16\% & 21.86\% & 20.75\% \\
    LayerMask (Ablated) & 6.61\% & 6.57\% & 6.49\% & 6.45\% & 6.45\% & 6.45\% \\
    LayerMask & 7.03\% & 6.99\% & 6.99\% & 6.99\% & 6.99\% & 6.99\% \\
    \bottomrule
  \end{tabular}
\end{table}

\begin{table}[htbp]
  \centering
  \small
  \setlength{\tabcolsep}{6pt}
  \caption{Average explanation size for EfficientNet-V2 on ImageNet-1k (GradCAM)}
  \label{tab:mask_coverage_imagenet_onek_efn_v2_gradcam}
  \begin{tabular}{@{}lrrrrrr@{}}
    \toprule
    Masking Value & \multicolumn{6}{c}{Threshold} \\
    \cmidrule(l{3pt}r{3pt}){2-7}
     & 0.9 & 0.7 & 0.5 & 0.3 & 0.1 & 0.0 \\
    \midrule
    Min & 19.70\% & 15.90\% & 13.80\% & 12.49\% & 11.89\% & 11.82\% \\
    Max & 18.67\% & 15.96\% & 14.41\% & 13.64\% & 13.12\% & 13.01\% \\
    Zero & 17.27\% & 13.69\% & 12.27\% & 10.67\% & 10.01\% & 9.99\% \\
    Avg & 16.79\% & 14.11\% & 11.81\% & 10.88\% & 10.26\% & 10.23\% \\
    \addlinespace[2pt]
    ConvAD & 28.65\% & 22.11\% & 19.53\% & 18.19\% & 17.29\% & 17.27\% \\
    LayerMask (Ablated) & 66.70\% & 64.10\% & 63.00\% & 62.33\% & 61.55\% & 61.32\% \\
    LayerMask & 22.07\% & 18.63\% & 16.89\% & 15.48\% & 14.50\% & 14.49\% \\
    \bottomrule
  \end{tabular}
\end{table}

\begin{table}[htbp]
  \centering
  \small
  \setlength{\tabcolsep}{6pt}
  \caption{Average explanation size for RegNet on ImageNet-1k (GradCAM)}
  \label{tab:mask_coverage_imagenet_onek_regnet_gradcam}
  \begin{tabular}{@{}lrrrrrr@{}}
    \toprule
    Masking Value & \multicolumn{6}{c}{Threshold} \\
    \cmidrule(l{3pt}r{3pt}){2-7}
     & 0.9 & 0.7 & 0.5 & 0.3 & 0.1 & 0.0 \\
    \midrule
    Min & 13.20\% & 9.71\% & 8.06\% & 7.00\% & 6.04\% & 5.72\% \\
    Max & 13.02\% & 8.48\% & 7.05\% & 5.88\% & 4.77\% & 4.20\% \\
    Zero & 10.49\% & 6.93\% & 5.82\% & 4.85\% & 4.11\% & 3.75\% \\
    Avg & 9.91\% & 7.11\% & 5.85\% & 5.08\% & 4.39\% & 4.15\% \\
    \addlinespace[2pt]
    ConvAD & 27.63\% & 22.71\% & 20.54\% & 18.13\% & 15.42\% & 13.87\% \\
    LayerMask (Ablated) & 43.11\% & 36.58\% & 33.92\% & 30.23\% & 26.19\% & 24.25\% \\
    LayerMask & 27.90\% & 23.62\% & 21.14\% & 18.87\% & 16.51\% & 14.20\% \\
    \bottomrule
  \end{tabular}
\end{table}

\begin{table}[htbp]
  \centering
  \small
  \setlength{\tabcolsep}{6pt}
  \caption{Average explanation size for ResNet on ImageNet-1k (LIME)}
  \label{tab:mask_coverage_imagenet_onek_resnet_lime}
  \begin{tabular}{@{}lrrrrrr@{}}
    \toprule
    Masking Value & \multicolumn{6}{c}{Threshold} \\
    \cmidrule(l{3pt}r{3pt}){2-7}
     & 0.9 & 0.7 & 0.5 & 0.3 & 0.1 & 0.0 \\
    \midrule
    Min & 26.88\% & 22.34\% & 19.05\% & 15.77\% & 13.00\% & 12.46\% \\
    Max & 38.22\% & 32.93\% & 29.01\% & 23.51\% & 20.03\% & 19.24\% \\
    Zero & 13.67\% & 11.52\% & 9.63\% & 7.98\% & 6.51\% & 5.91\% \\
    Avg & 16.48\% & 13.71\% & 11.48\% & 9.41\% & 8.22\% & 7.48\% \\
    \addlinespace[2pt]
    ConvAD & 25.53\% & 21.43\% & 18.75\% & 15.49\% & 13.07\% & 12.47\% \\
    LayerMask (Ablated) & 3.01\% & 3.00\% & 3.00\% & 3.00\% & 3.00\% & 3.00\% \\
    LayerMask & 4.70\% & 4.70\% & 4.70\% & 4.70\% & 4.70\% & 4.70\% \\
    \bottomrule
  \end{tabular}
\end{table}

\begin{table}[htbp]
  \centering
  \small
  \setlength{\tabcolsep}{6pt}
  \caption{Average explanation size for EfficientNet-V2 on ImageNet-1k (LIME)}
  \label{tab:mask_coverage_imagenet_onek_efn_v2_lime}
  \begin{tabular}{@{}lrrrrrr@{}}
    \toprule
    Masking Value & \multicolumn{6}{c}{Threshold} \\
    \cmidrule(l{3pt}r{3pt}){2-7}
     & 0.9 & 0.7 & 0.5 & 0.3 & 0.1 & 0.0 \\
    \midrule
    Min & 31.30\% & 26.46\% & 23.86\% & 21.74\% & 20.45\% & 20.38\% \\
    Max & 25.48\% & 22.03\% & 19.69\% & 18.21\% & 17.01\% & 16.98\% \\
    Zero & 18.70\% & 15.76\% & 14.29\% & 13.46\% & 12.30\% & 11.57\% \\
    Avg & 18.44\% & 15.73\% & 13.79\% & 12.81\% & 11.85\% & 11.64\% \\
    \addlinespace[2pt]
    ConvAD & 32.57\% & 27.66\% & 25.57\% & 24.00\% & 23.31\% & 23.30\% \\
    LayerMask (Ablated) & 64.06\% & 61.58\% & 59.88\% & 58.17\% & 57.35\% & 57.16\% \\
    LayerMask & 21.24\% & 18.11\% & 16.32\% & 15.47\% & 14.91\% & 14.90\% \\
    \bottomrule
  \end{tabular}
\end{table}

\begin{table}[htbp]
  \centering
  \small
  \setlength{\tabcolsep}{6pt}
  \caption{Average explanation size for RegNet on ImageNet-1k (LIME)}
  \label{tab:mask_coverage_imagenet_onek_regnet_lime}
  \begin{tabular}{@{}lrrrrrr@{}}
    \toprule
    Masking Value & \multicolumn{6}{c}{Threshold} \\
    \cmidrule(l{3pt}r{3pt}){2-7}
     & 0.9 & 0.7 & 0.5 & 0.3 & 0.1 & 0.0 \\
    \midrule
    Min & 23.40\% & 19.75\% & 16.33\% & 14.23\% & 12.14\% & 11.87\% \\
    Max & 25.12\% & 19.31\% & 16.60\% & 14.57\% & 12.91\% & 12.53\% \\
    Zero & 7.21\% & 5.92\% & 4.88\% & 4.23\% & 3.54\% & 3.07\% \\
    Avg & 10.01\% & 8.12\% & 6.84\% & 5.89\% & 5.29\% & 4.89\% \\
    \addlinespace[2pt]
    ConvAD & 14.61\% & 12.63\% & 11.39\% & 10.19\% & 9.27\% & 8.39\% \\
    LayerMask (Ablated) & 41.25\% & 38.05\% & 35.68\% & 33.66\% & 30.88\% & 29.10\% \\
    LayerMask & 14.87\% & 12.87\% & 11.55\% & 10.41\% & 9.33\% & 8.64\% \\
    \bottomrule
  \end{tabular}
\end{table}

\begin{table}[htbp]
  \centering
  \small
  \setlength{\tabcolsep}{6pt}
  \caption{Average explanation size for ResNet on ImageNet-1k V2 (ReX)}
  \label{tab:mask_coverage_in_1k_v2_resnet_rex}
  \begin{tabular}{@{}lrrrrrr@{}}
    \toprule
    Masking Value & \multicolumn{6}{c}{Threshold} \\
    \cmidrule(l{3pt}r{3pt}){2-7}
     & 0.9 & 0.7 & 0.5 & 0.3 & 0.1 & 0.0 \\
    \midrule
    Min & 22.09\% & 16.66\% & 12.73\% & 9.08\% & 5.91\% & 4.95\% \\
    Max & 22.64\% & 17.24\% & 13.19\% & 9.81\% & 6.76\% & 5.73\% \\
    Zero & 11.23\% & 7.88\% & 6.17\% & 4.82\% & 3.13\% & 2.25\% \\
    Avg & 13.54\% & 10.06\% & 7.42\% & 5.67\% & 3.69\% & 2.87\% \\
    \addlinespace[2pt]
    ConvAD & 26.50\% & 19.15\% & 14.84\% & 11.12\% & 7.45\% & 6.20\% \\
    LayerMask (Ablated) & 70.58\% & 69.33\% & 68.50\% & 66.94\% & 66.87\% & 65.98\% \\
    LayerMask & 8.00\% & 7.00\% & 6.22\% & 5.82\% & 5.53\% & 5.49\% \\
    \bottomrule
  \end{tabular}
\end{table}

\begin{table}[htbp]
  \centering
  \small
  \setlength{\tabcolsep}{6pt}
  \caption{Average explanation size for EfficientNet-V2 on ImageNet-1k V2 (ReX)}
  \label{tab:mask_coverage_in_1k_v2_efn_v2_rex}
  \begin{tabular}{@{}lrrrrrr@{}}
    \toprule
    Masking Value & \multicolumn{6}{c}{Threshold} \\
    \cmidrule(l{3pt}r{3pt}){2-7}
     & 0.9 & 0.7 & 0.5 & 0.3 & 0.1 & 0.0 \\
    \midrule
    Min & 16.98\% & 13.66\% & 12.28\% & 10.72\% & 9.80\% & 9.73\% \\
    Max & 15.09\% & 12.35\% & 10.93\% & 10.11\% & 9.49\% & 9.36\% \\
    Zero & 16.35\% & 13.70\% & 11.67\% & 10.08\% & 9.12\% & 8.97\% \\
    Avg & 15.99\% & 13.28\% & 11.36\% & 10.13\% & 9.14\% & 9.05\% \\
    \addlinespace[2pt]
    ConvAD & 23.69\% & 18.62\% & 16.52\% & 14.97\% & 13.93\% & 13.89\% \\
    LayerMask (Ablated) & 72.80\% & 72.13\% & 71.84\% & 71.65\% & 71.36\% & 71.23\% \\
    LayerMask & 18.07\% & 14.51\% & 13.00\% & 11.51\% & 11.09\% & 11.08\% \\
    \bottomrule
  \end{tabular}
\end{table}

\begin{table}[htbp]
  \centering
  \small
  \setlength{\tabcolsep}{6pt}
  \caption{Average explanation size for RegNet on ImageNet-1k V2 (ReX)}
  \label{tab:mask_coverage_in_1k_v2_regnet_rex}
  \begin{tabular}{@{}lrrrrrr@{}}
    \toprule
    Masking Value & \multicolumn{6}{c}{Threshold} \\
    \cmidrule(l{3pt}r{3pt}){2-7}
     & 0.9 & 0.7 & 0.5 & 0.3 & 0.1 & 0.0 \\
    \midrule
    Min & 10.79\% & 7.18\% & 5.69\% & 4.79\% & 3.96\% & 3.54\% \\
    Max & 10.86\% & 7.59\% & 6.33\% & 5.29\% & 4.27\% & 3.73\% \\
    Zero & 7.03\% & 4.52\% & 3.74\% & 2.97\% & 2.33\% & 2.01\% \\
    Avg & 7.53\% & 5.09\% & 4.06\% & 3.33\% & 2.63\% & 2.28\% \\
    \addlinespace[2pt]
    ConvAD & 16.61\% & 12.72\% & 11.12\% & 9.61\% & 8.27\% & 7.49\% \\
    LayerMask (Ablated) & 33.72\% & 30.19\% & 27.30\% & 25.42\% & 23.27\% & 21.16\% \\
    LayerMask & 14.06\% & 11.04\% & 9.45\% & 8.29\% & 7.23\% & 6.40\% \\
    \bottomrule
  \end{tabular}
\end{table}

\begin{table}[htbp]
  \centering
  \small
  \setlength{\tabcolsep}{6pt}
  \caption{Average explanation size for ResNet on ImageNet-1k V2 (GradCAM)}
  \label{tab:mask_coverage_in_1k_v2_resnet_gradcam}
  \begin{tabular}{@{}lrrrrrr@{}}
    \toprule
    Masking Value & \multicolumn{6}{c}{Threshold} \\
    \cmidrule(l{3pt}r{3pt}){2-7}
     & 0.9 & 0.7 & 0.5 & 0.3 & 0.1 & 0.0 \\
    \midrule
    Min & 48.40\% & 39.08\% & 29.47\% & 23.09\% & 16.01\% & 14.24\% \\
    Max & 50.00\% & 39.45\% & 32.23\% & 24.77\% & 15.65\% & 12.77\% \\
    Zero & 27.79\% & 21.62\% & 16.25\% & 13.11\% & 10.16\% & 8.40\% \\
    Avg & 36.72\% & 28.17\% & 20.32\% & 15.67\% & 11.76\% & 10.36\% \\
    \addlinespace[2pt]
    ConvAD & 54.12\% & 43.96\% & 36.54\% & 30.27\% & 23.96\% & 22.88\% \\
    LayerMask (Ablated) & 6.75\% & 6.39\% & 6.38\% & 6.35\% & 6.35\% & 6.35\% \\
    LayerMask & 7.12\% & 7.12\% & 7.12\% & 7.12\% & 7.12\% & 7.12\% \\
    \bottomrule
  \end{tabular}
\end{table}

\begin{table}[htbp]
  \centering
  \small
  \setlength{\tabcolsep}{6pt}
  \caption{Average explanation size for EfficientNet-V2 on ImageNet-1k V2 (GradCAM)}
  \label{tab:mask_coverage_in_1k_v2_efn_v2_gradcam}
  \begin{tabular}{@{}lrrrrrr@{}}
    \toprule
    Masking Value & \multicolumn{6}{c}{Threshold} \\
    \cmidrule(l{3pt}r{3pt}){2-7}
     & 0.9 & 0.7 & 0.5 & 0.3 & 0.1 & 0.0 \\
    \midrule
    Min & 18.55\% & 15.84\% & 13.57\% & 12.87\% & 12.36\% & 12.35\% \\
    Max & 17.99\% & 15.26\% & 14.00\% & 12.56\% & 12.05\% & 12.01\% \\
    Zero & 17.09\% & 13.78\% & 11.95\% & 11.02\% & 9.71\% & 9.61\% \\
    Avg & 16.49\% & 13.34\% & 11.36\% & 10.38\% & 9.64\% & 9.60\% \\
    \addlinespace[2pt]
    ConvAD & 26.04\% & 21.16\% & 18.60\% & 17.14\% & 16.56\% & 16.56\% \\
    LayerMask (Ablated) & 65.17\% & 63.27\% & 62.40\% & 61.36\% & 60.54\% & 60.43\% \\
    LayerMask & 21.97\% & 16.53\% & 15.23\% & 13.82\% & 13.26\% & 13.19\% \\
    \bottomrule
  \end{tabular}
\end{table}

\begin{table}[htbp]
  \centering
  \small
  \setlength{\tabcolsep}{6pt}
  \caption{Average explanation size for RegNet on ImageNet-1k V2 (GradCAM)}
  \label{tab:mask_coverage_in_1k_v2_regnet_gradcam}
  \begin{tabular}{@{}lrrrrrr@{}}
    \toprule
    Masking Value & \multicolumn{6}{c}{Threshold} \\
    \cmidrule(l{3pt}r{3pt}){2-7}
     & 0.9 & 0.7 & 0.5 & 0.3 & 0.1 & 0.0 \\
    \midrule
    Min & 13.22\% & 9.15\% & 7.66\% & 6.57\% & 5.63\% & 5.24\% \\
    Max & 14.32\% & 8.61\% & 6.83\% & 6.04\% & 4.07\% & 3.46\% \\
    Zero & 9.71\% & 6.14\% & 4.56\% & 3.81\% & 3.26\% & 2.94\% \\
    Avg & 9.75\% & 6.68\% & 5.05\% & 4.09\% & 3.28\% & 2.92\% \\
    \addlinespace[2pt]
    ConvAD & 25.68\% & 20.04\% & 17.67\% & 15.38\% & 13.75\% & 12.48\% \\
    LayerMask (Ablated) & 38.61\% & 32.25\% & 28.05\% & 25.18\% & 22.30\% & 20.14\% \\
    LayerMask & 25.05\% & 20.90\% & 18.52\% & 16.82\% & 14.48\% & 12.53\% \\
    \bottomrule
  \end{tabular}
\end{table}

\begin{table}[htbp]
  \centering
  \small
  \setlength{\tabcolsep}{6pt}
  \caption{Average explanation size for ResNet on ImageNet-1k V2 (LIME)}
  \label{tab:mask_coverage_in_1k_v2_resnet_lime}
  \begin{tabular}{@{}lrrrrrr@{}}
    \toprule
    Masking Value & \multicolumn{6}{c}{Threshold} \\
    \cmidrule(l{3pt}r{3pt}){2-7}
     & 0.9 & 0.7 & 0.5 & 0.3 & 0.1 & 0.0 \\
    \midrule
    Min & 26.58\% & 22.20\% & 20.10\% & 17.08\% & 14.33\% & 13.77\% \\
    Max & 38.75\% & 33.02\% & 28.24\% & 23.58\% & 19.21\% & 18.09\% \\
    Zero & 13.04\% & 9.64\% & 8.20\% & 6.67\% & 5.53\% & 5.08\% \\
    Avg & 16.77\% & 14.13\% & 11.37\% & 9.35\% & 8.21\% & 7.90\% \\
    \addlinespace[2pt]
    ConvAD & 23.24\% & 19.40\% & 16.08\% & 13.89\% & 11.85\% & 11.28\% \\
    LayerMask (Ablated) & 3.46\% & 3.43\% & 3.43\% & 3.43\% & 3.43\% & 3.43\% \\
    LayerMask & 5.59\% & 5.59\% & 5.59\% & 5.59\% & 5.59\% & 5.59\% \\
    \bottomrule
  \end{tabular}
\end{table}

\begin{table}[htbp]
  \centering
  \small
  \setlength{\tabcolsep}{6pt}
  \caption{Average explanation size for EfficientNet-V2 on ImageNet-1k V2 (LIME)}
  \label{tab:mask_coverage_in_1k_v2_efn_v2_lime}
  \begin{tabular}{@{}lrrrrrr@{}}
    \toprule
    Masking Value & \multicolumn{6}{c}{Threshold} \\
    \cmidrule(l{3pt}r{3pt}){2-7}
     & 0.9 & 0.7 & 0.5 & 0.3 & 0.1 & 0.0 \\
    \midrule
    Min & 25.33\% & 21.65\% & 18.28\% & 16.59\% & 15.42\% & 15.36\% \\
    Max & 22.50\% & 18.93\% & 17.50\% & 16.40\% & 15.42\% & 15.28\% \\
    Zero & 16.19\% & 13.78\% & 12.49\% & 11.61\% & 11.17\% & 11.13\% \\
    Avg & 15.90\% & 13.14\% & 12.01\% & 11.34\% & 10.75\% & 10.70\% \\
    \addlinespace[2pt]
    ConvAD & 30.68\% & 25.44\% & 22.93\% & 21.50\% & 20.67\% & 20.64\% \\
    LayerMask (Ablated) & 63.86\% & 61.10\% & 59.56\% & 58.26\% & 57.00\% & 56.84\% \\
    LayerMask & 22.60\% & 18.47\% & 16.86\% & 15.66\% & 14.88\% & 14.79\% \\
    \bottomrule
  \end{tabular}
\end{table}

\begin{table}[htbp]
  \centering
  \small
  \setlength{\tabcolsep}{6pt}
  \caption{Average explanation size for RegNet on ImageNet-1k V2 (LIME)}
  \label{tab:mask_coverage_in_1k_v2_regnet_lime}
  \begin{tabular}{@{}lrrrrrr@{}}
    \toprule
    Masking Value & \multicolumn{6}{c}{Threshold} \\
    \cmidrule(l{3pt}r{3pt}){2-7}
     & 0.9 & 0.7 & 0.5 & 0.3 & 0.1 & 0.0 \\
    \midrule
    Min & 22.69\% & 17.61\% & 15.02\% & 12.52\% & 10.31\% & 9.82\% \\
    Max & 24.40\% & 18.86\% & 16.00\% & 13.98\% & 12.35\% & 12.22\% \\
    Zero & 7.10\% & 5.94\% & 5.38\% & 4.43\% & 3.63\% & 3.22\% \\
    Avg & 9.27\% & 7.95\% & 6.88\% & 6.11\% & 5.48\% & 5.17\% \\
    \addlinespace[2pt]
    ConvAD & 14.07\% & 12.63\% & 11.68\% & 10.44\% & 9.35\% & 8.80\% \\
    LayerMask (Ablated) & 41.80\% & 37.09\% & 34.42\% & 32.55\% & 30.66\% & 29.42\% \\
    LayerMask & 14.17\% & 12.37\% & 11.38\% & 10.25\% & 9.23\% & 8.47\% \\
    \bottomrule
  \end{tabular}
\end{table}

\begin{table}[htbp]
  \centering
  \small
  \setlength{\tabcolsep}{6pt}
  \caption{Average explanation size for ResNet on PASCAL-VOC (ReX)}
  \label{tab:mask_coverage_pascal_voc_resnet_rex}
  \begin{tabular}{@{}lrrrrrr@{}}
    \toprule
    Masking Value & \multicolumn{6}{c}{Threshold} \\
    \cmidrule(l{3pt}r{3pt}){2-7}
     & 0.9 & 0.7 & 0.5 & 0.3 & 0.1 & 0.0 \\
    \midrule
    Min & 13.26\% & 10.45\% & 9.01\% & 8.60\% & 8.54\% & 8.54\% \\
    Max & 5.92\% & 3.93\% & 3.31\% & 2.84\% & 2.66\% & 2.66\% \\
    Zero & 4.60\% & 2.68\% & 2.17\% & 1.75\% & 1.60\% & 1.60\% \\
    Avg & 4.53\% & 3.11\% & 2.56\% & 2.28\% & 2.17\% & 2.17\% \\
    \addlinespace[2pt]
    ConvAD & 8.35\% & 5.34\% & 4.02\% & 3.23\% & 2.75\% & 2.75\% \\
    LayerMask (Ablated) & 25.57\% & 25.49\% & 25.86\% & 25.86\% & 25.86\% & 25.66\% \\
    LayerMask & 3.33\% & 2.61\% & 2.02\% & 1.60\% & 1.33\% & 1.24\% \\
    \bottomrule
  \end{tabular}
\end{table}

\begin{table}[htbp]
  \centering
  \small
  \setlength{\tabcolsep}{6pt}
  \caption{Average explanation size for EfficientNet-V2 on PASCAL-VOC (ReX)}
  \label{tab:mask_coverage_pascal_voc_efn_v2_rex}
  \begin{tabular}{@{}lrrrrrr@{}}
    \toprule
    Masking Value & \multicolumn{6}{c}{Threshold} \\
    \cmidrule(l{3pt}r{3pt}){2-7}
     & 0.9 & 0.7 & 0.5 & 0.3 & 0.1 & 0.0 \\
    \midrule
    Min & 7.35\% & 5.91\% & 5.29\% & 4.57\% & 4.30\% & 4.30\% \\
    Max & 5.74\% & 4.25\% & 3.65\% & 3.24\% & 3.09\% & 3.09\% \\
    Zero & 6.71\% & 5.22\% & 3.94\% & 3.05\% & 2.81\% & 2.81\% \\
    Avg & 5.81\% & 4.32\% & 3.48\% & 2.70\% & 2.54\% & 2.54\% \\
    \addlinespace[2pt]
    ConvAD & 12.70\% & 10.98\% & 9.17\% & 8.36\% & 8.09\% & 8.09\% \\
    LayerMask (Ablated) & 17.21\% & 16.97\% & 16.93\% & 16.91\% & 16.90\% & 16.90\% \\
    LayerMask & 6.04\% & 4.96\% & 4.09\% & 3.54\% & 3.47\% & 3.47\% \\
    \bottomrule
  \end{tabular}
\end{table}

\begin{table}[htbp]
  \centering
  \small
  \setlength{\tabcolsep}{6pt}
  \caption{Average explanation size for RegNet on PASCAL-VOC (ReX)}
  \label{tab:mask_coverage_pascal_voc_regnet_rex}
  \begin{tabular}{@{}lrrrrrr@{}}
    \toprule
    Masking Value & \multicolumn{6}{c}{Threshold} \\
    \cmidrule(l{3pt}r{3pt}){2-7}
     & 0.9 & 0.7 & 0.5 & 0.3 & 0.1 & 0.0 \\
    \midrule
    Min & 3.07\% & 2.18\% & 1.59\% & 1.15\% & 1.00\% & 1.00\% \\
    Max & 3.35\% & 2.52\% & 1.84\% & 1.51\% & 1.30\% & 1.30\% \\
    Zero & 2.86\% & 1.81\% & 1.48\% & 1.22\% & 1.09\% & 1.09\% \\
    Avg & 2.52\% & 1.89\% & 1.54\% & 1.21\% & 1.16\% & 1.16\% \\
    \addlinespace[2pt]
    ConvAD & 17.93\% & 15.61\% & 13.92\% & 12.84\% & 12.70\% & 12.70\% \\
    LayerMask (Ablated) & 22.72\% & 18.67\% & 16.05\% & 14.21\% & 13.62\% & 13.62\% \\
    LayerMask & 21.84\% & 18.75\% & 16.54\% & 14.70\% & 14.38\% & 14.38\% \\
    \bottomrule
  \end{tabular}
\end{table}

\begin{table}[htbp]
  \centering
  \small
  \setlength{\tabcolsep}{6pt}
  \caption{Average explanation size for ResNet on PASCAL-VOC (GradCAM)}
  \label{tab:mask_coverage_pascal_voc_resnet_gradcam}
  \begin{tabular}{@{}lrrrrrr@{}}
    \toprule
    Masking Value & \multicolumn{6}{c}{Threshold} \\
    \cmidrule(l{3pt}r{3pt}){2-7}
     & 0.9 & 0.7 & 0.5 & 0.3 & 0.1 & 0.0 \\
    \midrule
    Min & 19.01\% & 16.81\% & 15.30\% & 14.13\% & 14.04\% & 14.04\% \\
    Max & 10.11\% & 7.11\% & 5.86\% & 4.79\% & 4.58\% & 4.58\% \\
    Zero & 9.01\% & 6.63\% & 4.82\% & 4.32\% & 4.15\% & 4.15\% \\
    Avg & 9.03\% & 6.84\% & 5.66\% & 4.62\% & 4.39\% & 4.39\% \\
    \addlinespace[2pt]
    ConvAD & 14.18\% & 11.59\% & 8.79\% & 7.32\% & 7.06\% & 7.06\% \\
    LayerMask (Ablated) & 5.11\% & 4.70\% & 3.84\% & 3.54\% & 3.16\% & 3.16\% \\
    LayerMask & 1.43\% & 1.29\% & 1.29\% & 1.29\% & 1.29\% & 1.29\% \\
    \bottomrule
  \end{tabular}
\end{table}

\begin{table}[htbp]
  \centering
  \small
  \setlength{\tabcolsep}{6pt}
  \caption{Average explanation size for EfficientNet-V2 on PASCAL-VOC (GradCAM)}
  \label{tab:mask_coverage_pascal_voc_efn_v2_gradcam}
  \begin{tabular}{@{}lrrrrrr@{}}
    \toprule
    Masking Value & \multicolumn{6}{c}{Threshold} \\
    \cmidrule(l{3pt}r{3pt}){2-7}
     & 0.9 & 0.7 & 0.5 & 0.3 & 0.1 & 0.0 \\
    \midrule
    Min & 9.23\% & 8.00\% & 7.05\% & 6.58\% & 6.16\% & 6.16\% \\
    Max & 8.80\% & 7.18\% & 6.28\% & 5.20\% & 4.98\% & 4.98\% \\
    Zero & 6.56\% & 5.20\% & 4.56\% & 3.85\% & 3.73\% & 3.73\% \\
    Avg & 6.84\% & 5.58\% & 4.19\% & 3.51\% & 3.35\% & 3.35\% \\
    \addlinespace[2pt]
    ConvAD & 14.47\% & 11.03\% & 9.44\% & 8.95\% & 8.93\% & 8.93\% \\
    LayerMask (Ablated) & 18.06\% & 18.31\% & 17.96\% & 18.15\% & 17.85\% & 18.16\% \\
    LayerMask & 9.97\% & 8.44\% & 7.32\% & 6.53\% & 6.41\% & 6.41\% \\
    \bottomrule
  \end{tabular}
\end{table}

\begin{table}[htbp]
  \centering
  \small
  \setlength{\tabcolsep}{6pt}
  \caption{Average explanation size for RegNet on PASCAL-VOC (GradCAM)}
  \label{tab:mask_coverage_pascal_voc_regnet_gradcam}
  \begin{tabular}{@{}lrrrrrr@{}}
    \toprule
    Masking Value & \multicolumn{6}{c}{Threshold} \\
    \cmidrule(l{3pt}r{3pt}){2-7}
     & 0.9 & 0.7 & 0.5 & 0.3 & 0.1 & 0.0 \\
    \midrule
    Min & 3.44\% & 2.80\% & 1.80\% & 1.52\% & 1.41\% & 1.41\% \\
    Max & 3.24\% & 2.43\% & 1.85\% & 1.49\% & 1.33\% & 1.33\% \\
    Zero & 2.23\% & 1.76\% & 1.43\% & 1.32\% & 1.23\% & 1.23\% \\
    Avg & 2.39\% & 1.87\% & 1.58\% & 1.14\% & 1.07\% & 1.07\% \\
    \addlinespace[2pt]
    ConvAD & 43.68\% & 40.92\% & 37.93\% & 34.53\% & 34.52\% & 34.52\% \\
    LayerMask (Ablated) & 37.89\% & 35.99\% & 33.82\% & 30.15\% & 29.83\% & 29.83\% \\
    LayerMask & 57.12\% & 52.53\% & 51.14\% & 50.48\% & 50.48\% & 50.48\% \\
    \bottomrule
  \end{tabular}
\end{table}

\begin{table}[htbp]
  \centering
  \small
  \setlength{\tabcolsep}{6pt}
  \caption{Average explanation size for ResNet on PASCAL-VOC (LIME)}
  \label{tab:mask_coverage_pascal_voc_resnet_lime}
  \begin{tabular}{@{}lrrrrrr@{}}
    \toprule
    Masking Value & \multicolumn{6}{c}{Threshold} \\
    \cmidrule(l{3pt}r{3pt}){2-7}
     & 0.9 & 0.7 & 0.5 & 0.3 & 0.1 & 0.0 \\
    \midrule
    Min & 17.22\% & 14.38\% & 13.19\% & 12.76\% & 12.67\% & 12.67\% \\
    Max & 14.20\% & 11.16\% & 9.53\% & 7.98\% & 7.46\% & 7.46\% \\
    Zero & 4.48\% & 3.47\% & 2.92\% & 2.47\% & 2.34\% & 2.34\% \\
    Avg & 4.64\% & 3.77\% & 3.32\% & 3.00\% & 2.83\% & 2.83\% \\
    \addlinespace[2pt]
    ConvAD & 8.21\% & 6.92\% & 5.88\% & 5.16\% & 4.48\% & 4.48\% \\
    LayerMask (Ablated) & 3.99\% & 3.20\% & 2.70\% & 2.31\% & 2.25\% & 2.24\% \\
    LayerMask & 0.97\% & 0.93\% & 0.82\% & 0.76\% & 0.76\% & 0.76\% \\
    \bottomrule
  \end{tabular}
\end{table}

\begin{table}[htbp]
  \centering
  \small
  \setlength{\tabcolsep}{6pt}
  \caption{Average explanation size for EfficientNet-V2 on PASCAL-VOC (LIME)}
  \label{tab:mask_coverage_pascal_voc_efn_v2_lime}
  \begin{tabular}{@{}lrrrrrr@{}}
    \toprule
    Masking Value & \multicolumn{6}{c}{Threshold} \\
    \cmidrule(l{3pt}r{3pt}){2-7}
     & 0.9 & 0.7 & 0.5 & 0.3 & 0.1 & 0.0 \\
    \midrule
    Min & 9.92\% & 8.11\% & 6.87\% & 5.57\% & 5.51\% & 5.51\% \\
    Max & 10.12\% & 7.21\% & 6.15\% & 5.21\% & 5.05\% & 5.05\% \\
    Zero & 7.71\% & 6.30\% & 5.07\% & 4.31\% & 4.11\% & 4.11\% \\
    Avg & 7.91\% & 6.04\% & 5.00\% & 4.35\% & 4.22\% & 4.22\% \\
    \addlinespace[2pt]
    ConvAD & 13.61\% & 11.44\% & 9.89\% & 8.24\% & 8.16\% & 8.16\% \\
    LayerMask (Ablated) & 16.99\% & 16.67\% & 16.55\% & 16.50\% & 16.50\% & 16.50\% \\
    LayerMask & 8.32\% & 7.03\% & 6.11\% & 5.53\% & 5.51\% & 5.51\% \\
    \bottomrule
  \end{tabular}
\end{table}

\begin{table}[htbp]
  \centering
  \small
  \setlength{\tabcolsep}{6pt}
  \caption{Average explanation size for RegNet on PASCAL-VOC (LIME)}
  \label{tab:mask_coverage_pascal_voc_regnet_lime}
  \begin{tabular}{@{}lrrrrrr@{}}
    \toprule
    Masking Value & \multicolumn{6}{c}{Threshold} \\
    \cmidrule(l{3pt}r{3pt}){2-7}
     & 0.9 & 0.7 & 0.5 & 0.3 & 0.1 & 0.0 \\
    \midrule
    Min & 5.46\% & 4.08\% & 3.47\% & 3.12\% & 2.58\% & 2.58\% \\
    Max & 5.78\% & 4.44\% & 3.72\% & 3.24\% & 2.91\% & 2.91\% \\
    Zero & 3.70\% & 2.91\% & 2.56\% & 2.00\% & 1.73\% & 1.73\% \\
    Avg & 3.53\% & 2.85\% & 2.46\% & 2.01\% & 1.86\% & 1.86\% \\
    \addlinespace[2pt]
    ConvAD & 13.89\% & 12.45\% & 11.06\% & 10.15\% & 9.83\% & 9.83\% \\
    LayerMask (Ablated) & 38.67\% & 30.95\% & 25.38\% & 22.14\% & 20.28\% & 20.28\% \\
    LayerMask & 21.61\% & 18.58\% & 17.35\% & 16.13\% & 15.70\% & 15.70\% \\
    \bottomrule
  \end{tabular}
\end{table}

\section{Confidence scores of explanations and confidence difference across different backgrounds}
\label{sup:sec:conf_dropoff}

\begin{figure}[H]
    \centering
    \includegraphics[width=\linewidth]{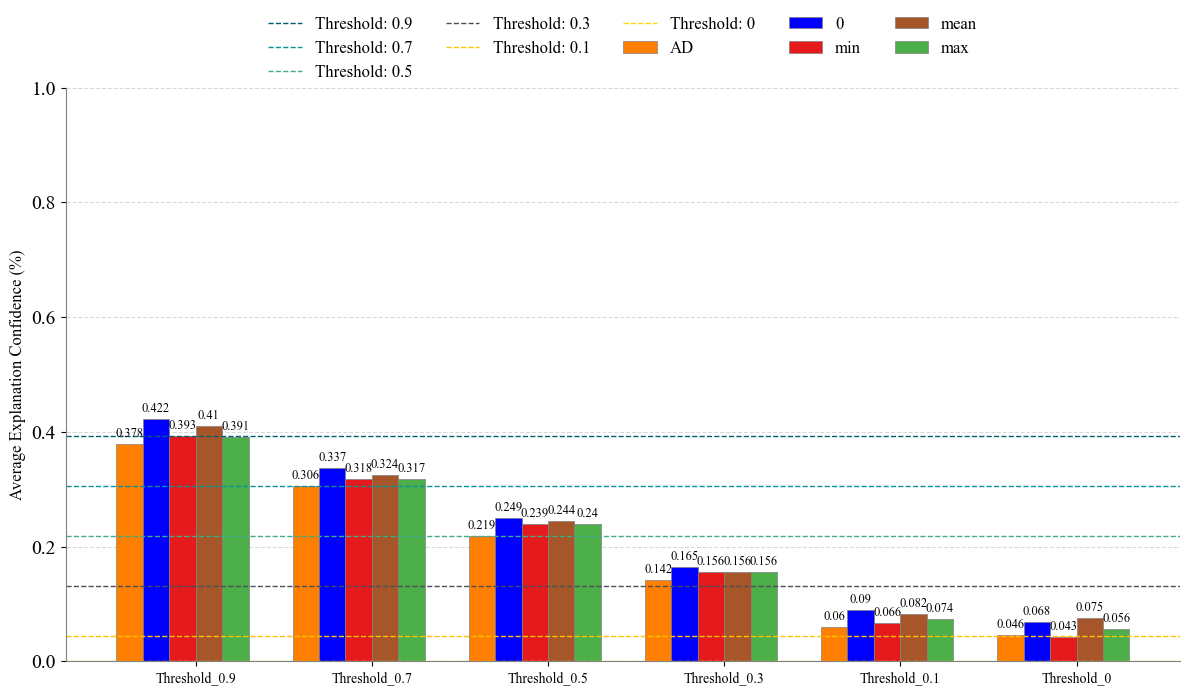}
    \caption{Average Explanation Confidence of different methods for ResNet on ImageNet-1k}
    \label{fig:resnet_avg_conf_in1k_app}
\end{figure}

\begin{figure}[H]
    \centering
    \includegraphics[width=\linewidth]{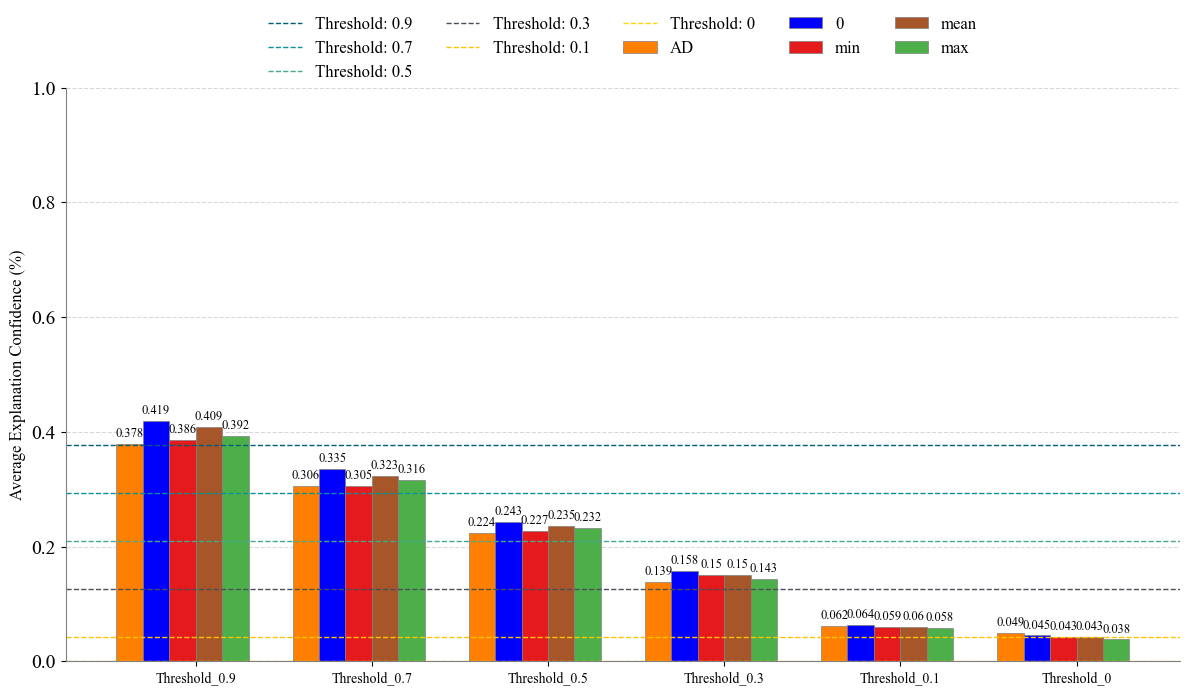}
    \caption{Average Explanation Confidence of different methods for ResNet on IN-1k\_V2}
    \label{fig:resnet_avg_conf_in1k_v2_app}
\end{figure}

\begin{figure}[H]
    \centering
    \includegraphics[width=\linewidth]{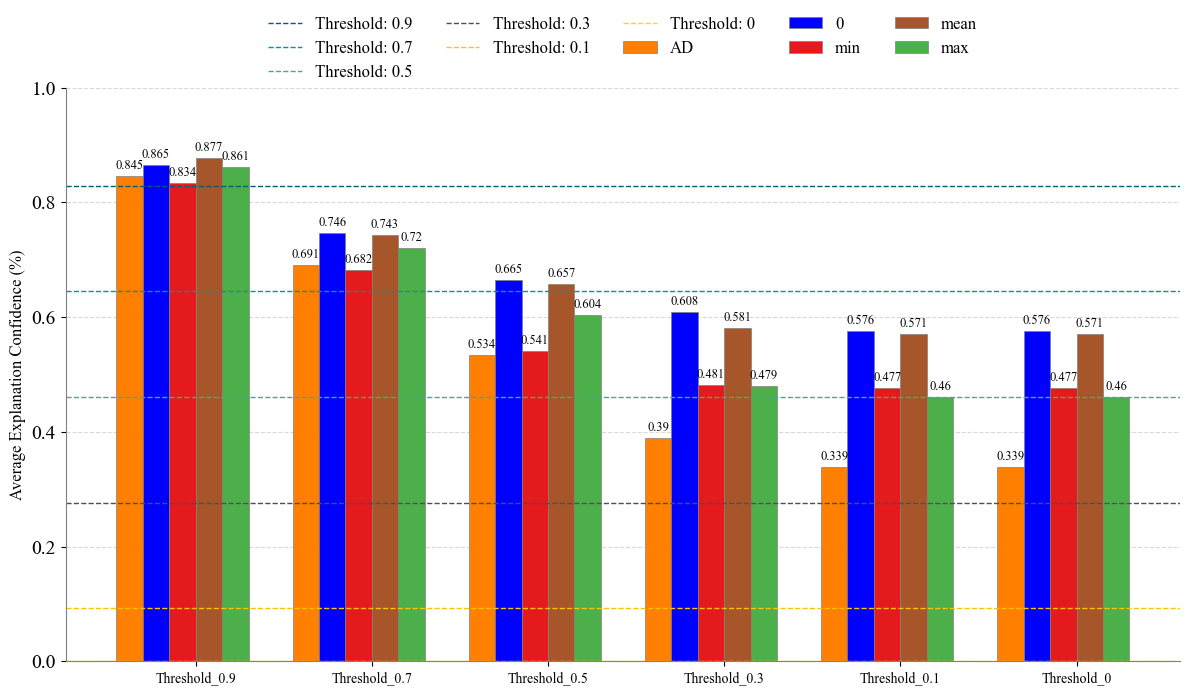}
    \caption{Average Explanation Confidence of different methods for ResNet on PASCAL-VOC}
    \label{fig:resnet_avg_conf_voc_app}
\end{figure}

\begin{figure}[H]
    \centering
    \includegraphics[width=\linewidth]{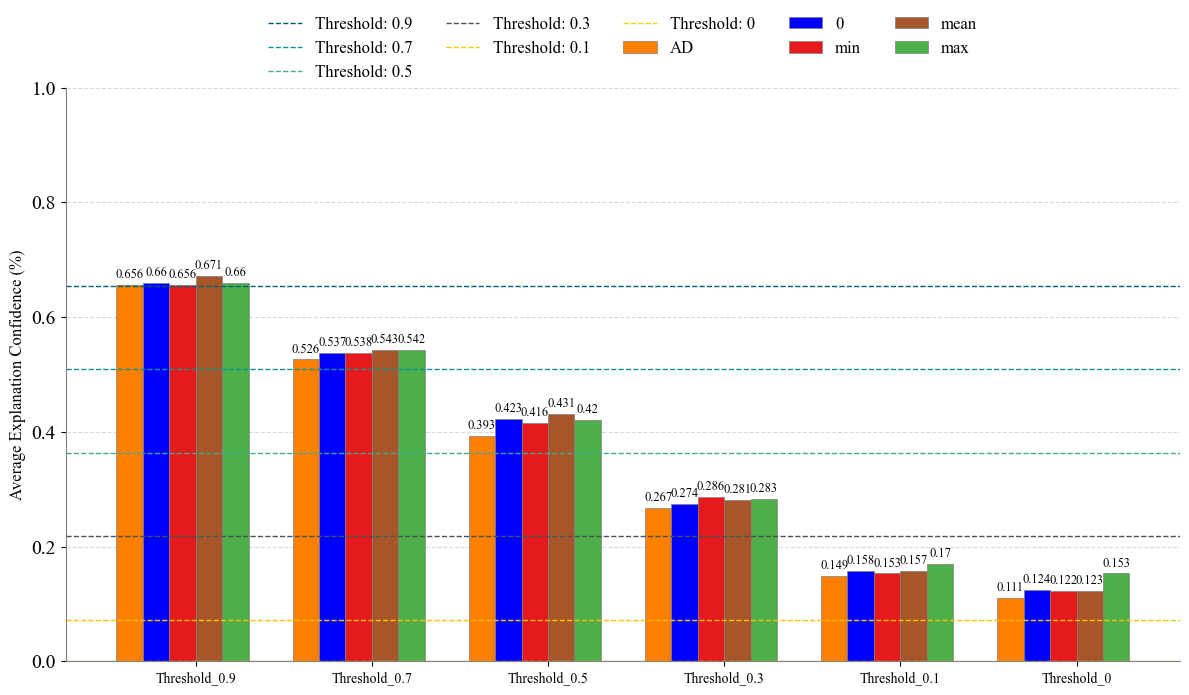}
    \caption{Average Explanation Confidence of different methods for RegNet-Y on ImageNet-1k}
    \label{fig:regnet_avg_conf_in1k_app}
\end{figure}

\begin{figure}[H]
    \centering
    \includegraphics[width=\linewidth]{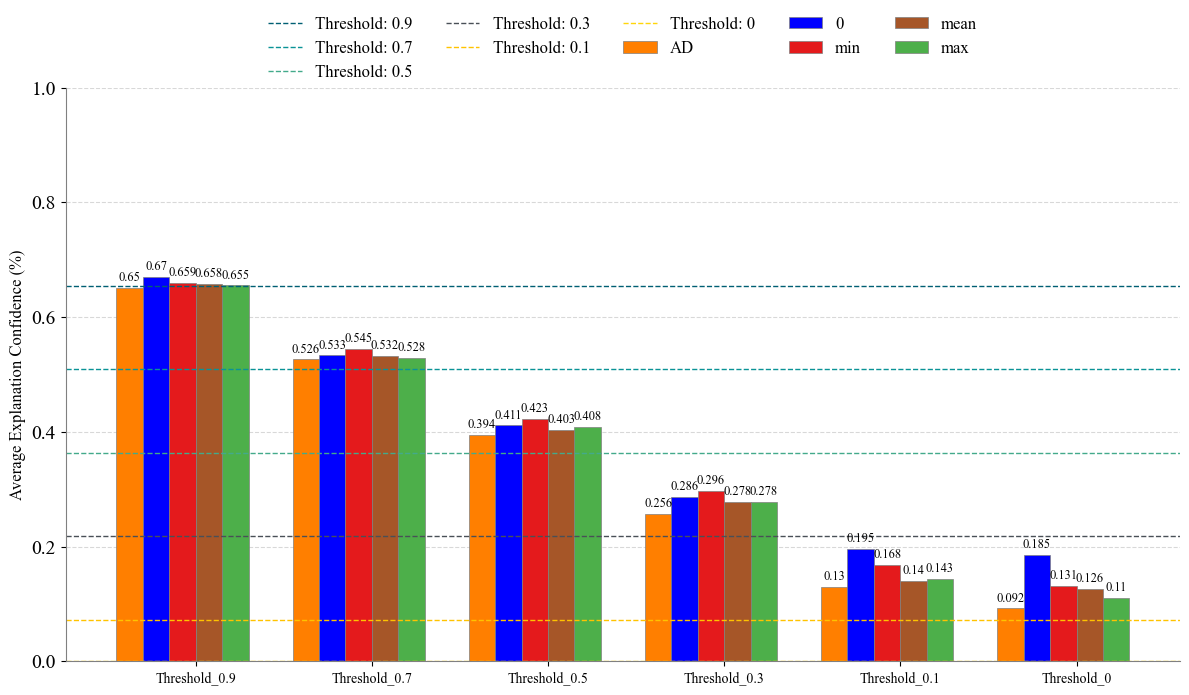}
    \caption{Average Explanation Confidence of different methods for RegNet-Y on IN-1k\_V2}
    \label{fig:regnet_avg_conf_in1k_v2_app}
\end{figure}

\begin{figure}[H]
    \centering
    \includegraphics[width=\linewidth]{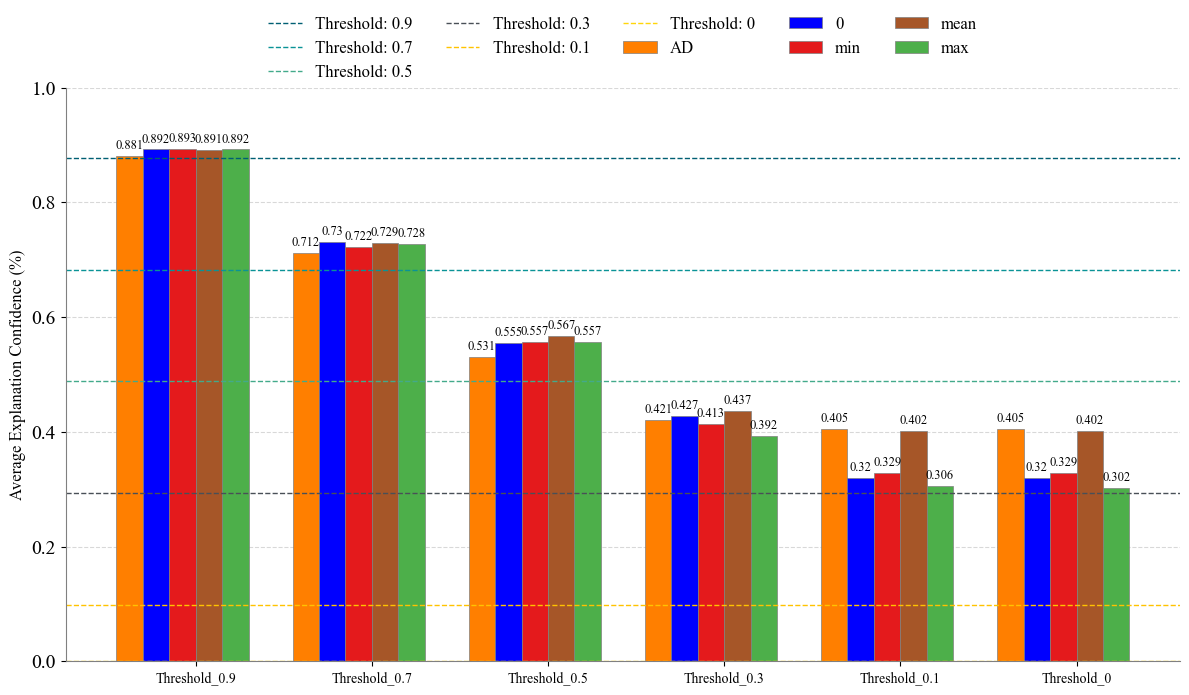}
    \caption{Average Explanation Confidence of different methods for RegNet-Y on PASCAL-VOC}
    \label{fig:regnet_avg_conf_voc_app}
\end{figure}

\begin{figure}[H]
    \centering
    \includegraphics[width=\linewidth]{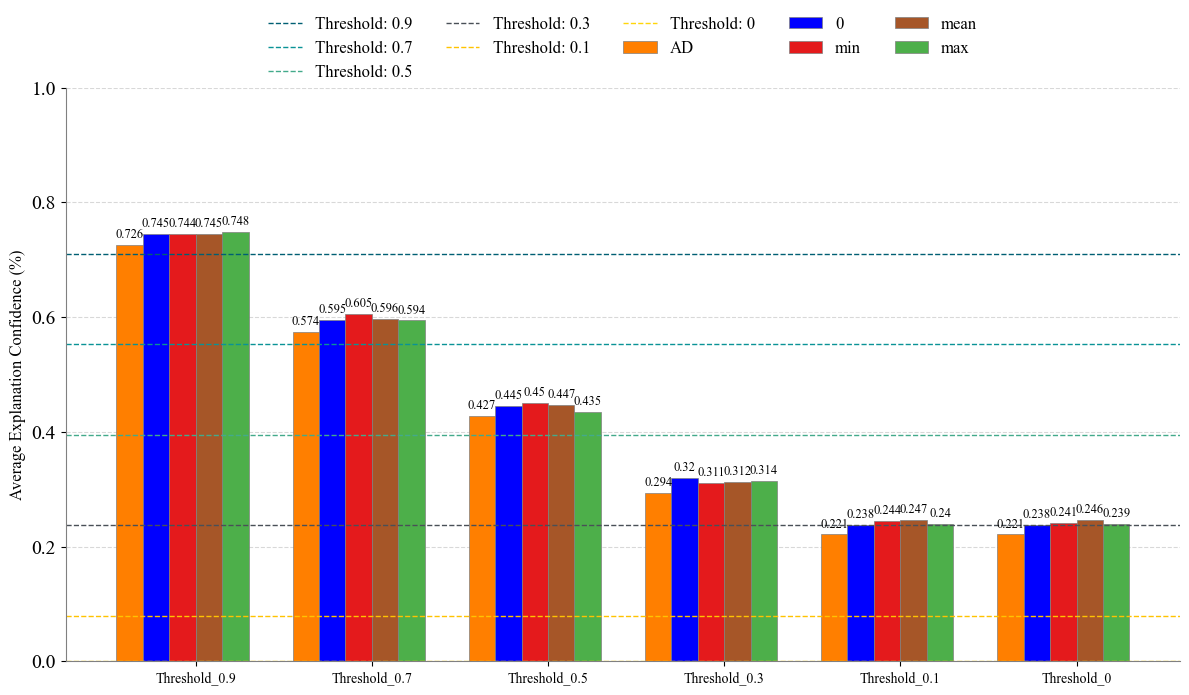}
    \caption{Average Explanation Confidence of different methods for EFN-V2 on ImageNet-1k}
    \label{fig:efn_v2_avg_conf_in1k_app}
\end{figure}

\begin{figure}[H]
    \centering
    \includegraphics[width=\linewidth]{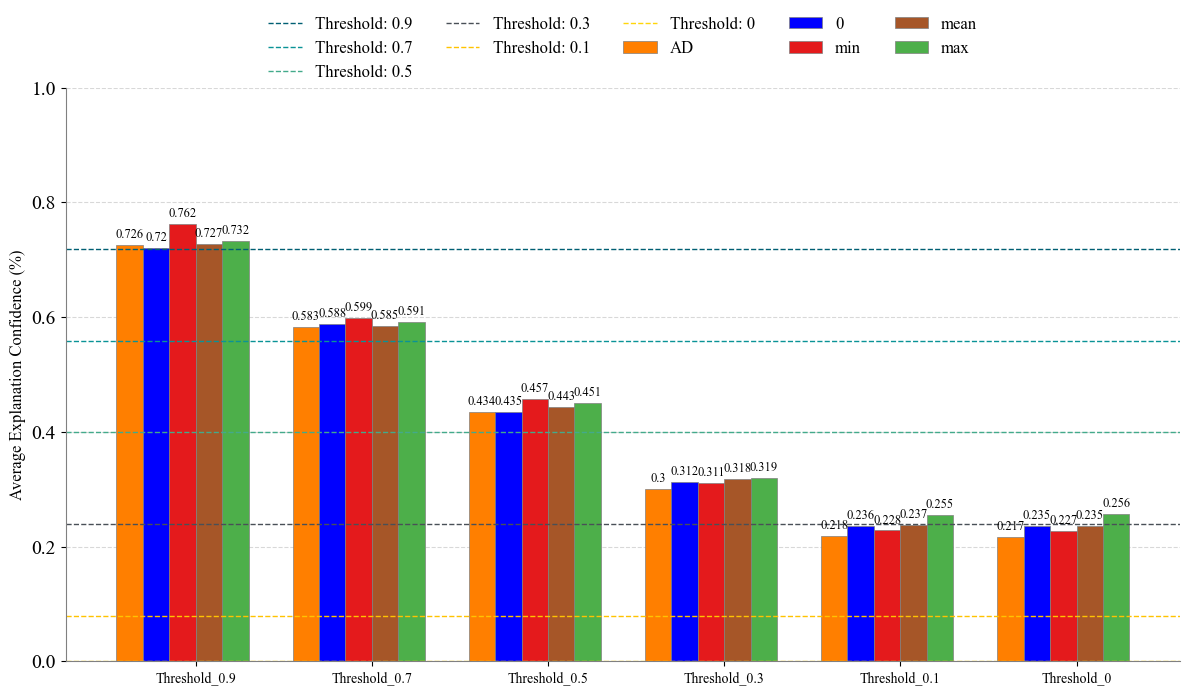}
    \caption{Average Explanation Confidence of different methods for EFN-V2 on IN-1k\_V2}
    \label{fig:efn_v2_avg_conf_in1k_v2_app}
\end{figure}

\begin{figure}[H]
    \centering
    \includegraphics[width=\linewidth]{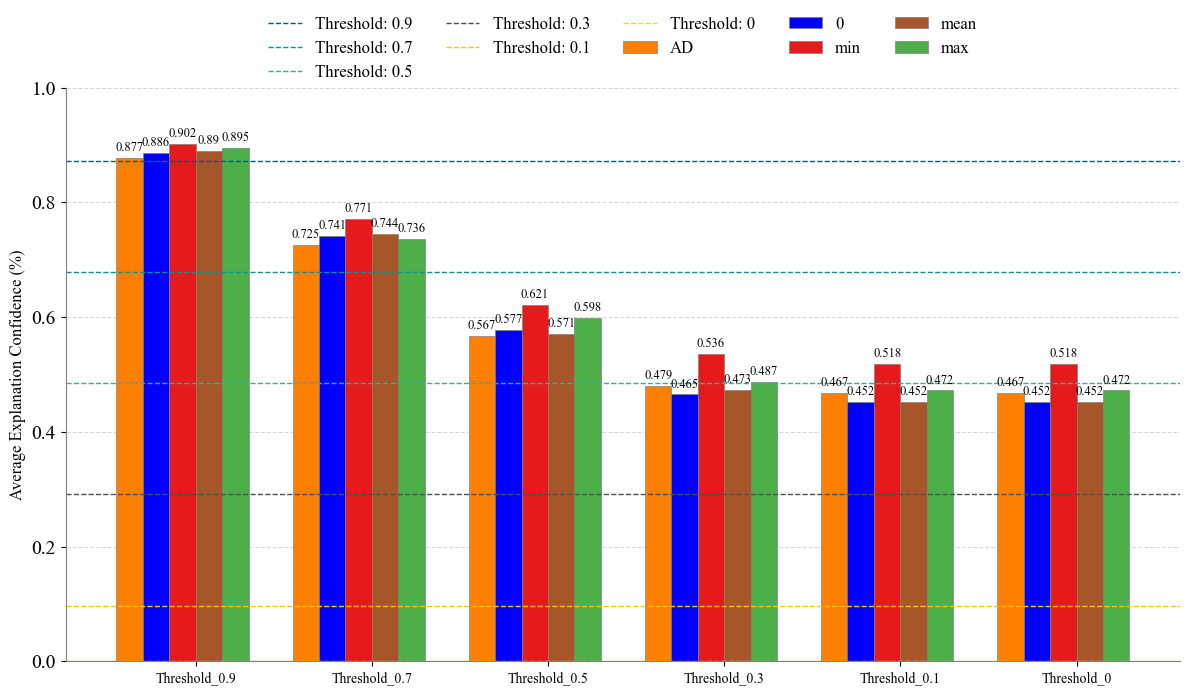}
    \caption{Average Explanation Confidence of different methods for EFN-v2 on PASCAL-VOC}
    \label{fig:efn_v2_avg_conf_voc_app}
\end{figure}

\section{Discoverability}
\label{appendix:Discoverability}
\subsection{Discoverability of Explanations}
\begin{figure}[!h]
    \centering
    \input{Plots/discoverability_rex}
    \caption{Discoverability}
    \label{fig:discoverability}
\end{figure}

One of the primary objectives of XAI tools is to analyze the workings of the model being explained in order to identify bugs and edge-cases in the model's decision-making process (artifacts such as spurious correlations or incorrect attribution) that can lead to unwanted outcomes, such as poor robustness and increased susceptibility to adversarial attacks. 

To this end, it is imperative that a given perturbation strategy have the ability to \textit{discover} explanations for the model and a particular input across various confidence thresholds. The high discoverability ensures that the XAI tool comprehensively identifies faults and edge cases along the model's complete confidence spectrum while minimizing the required number of test samples, which are hallmarks of high-quality and effective test suites \cite{kuhn2004,andrews2005,just2014}. Formally, we state it as:

\dfn[{Discoverability at $[\gamma, \gamma']$}]\label{def:discover}
The discoverability, $\zeta$, of a strategy at interval $\left[\gamma, \gamma'\right], \gamma \le \gamma'$, 
given a set of test samples $\eta$, is the proportion of $\eta$ for which the strategy was able to discover an explanation with confidence $c', \gamma \cdot c \le c' \le \gamma' c$, where $c$ is the model's confidence on the unperturbed test sample. 
\edfn

 A strategy with $\zeta = 1$ at $[\gamma, \gamma']$ means that the XAI tool using the strategy was able to discover an explanation with confidence $c' \in \left[\gamma\cdot c, \gamma' \cdot c \right]$ for all test examples in $\eta$. It is important to note that discoverability for competing strategies should only be compared for the same $\eta$ and $[\gamma, \gamma']$.

As stated in \cref{def:discover}, we evaluate the discoverability to determine the efficacy of the strategy in discovering explanations. For each perturbation strategy, model and dataset,  we fixed the test suite, $\eta$ to same 150 images, allowing for direct comparisions across the different strategies. For each strategy, discoverability is calculated at $[0, 0.1], [0.1, 0.3], [0.3,0.5], [0.5,0.7], [0.7, 0.9]$ and $[0.9, 1]$. We find that \convad consistently outperforms most, if not all of the competing strategies in terms of discoverability across all the models and datasets tested.  The discoverability gap is especially visible in the case of ResNet-50 and EfficientNet, \convad attains the best discoverability scores across thresholds for IN-1k, IN-v2 and PASCAL-VOC and in the case of ResNet-50, LayerMask and LayerMask (Ablated) have significantly lower discoverability. On average, \convad has $1.96\times$ greater discoverability than LayerMask and $13.26\times$ greater than LayerMask (Ablated). For EfficientNet-V2, however, the gap between \convad and LayerMask is reduced drastically, with \convad having $14.85\%$ greater discoverability than LayerMask. The gap between \convad and LayerMask (Ablated) is still sizable with \convad having $4.36\times$ greater discoverability.

However, in the case of RegNet we observe an anomaly. LayerMask marginally outperforms \convad in terms of discoverability, being $2.96\%$ greater; however, this difference can be offset by the large disparity in robustness of the explanations between \convad and LayerMask. However, \convad still outperforms LayerMask (Ablated), having $37.86\%$ greater discoverability on average.

\FloatBarrier

\clearpage

\section{Transferability}\label{appendix:transferability}
\input{Figures/all_transferability_matrix}

\clearpage

\section{Overlap}

\begin{table}[h]
\centering
\caption{Mean pairwise \textbf{DICE} scores between explanation strategy groups across methods (GradCAM, LIME, ReX), models, and datasets. $\mathcal{M}$×$\mathcal{M}$ = Masking vs.\ Masking (Max, Zero, Mean, Min), AD×$\mathcal{M}$ = AD vs.\ Masking, AD×LM = AD vs.\ LayerMask/LayerMask (Ablated), $\mathcal{M}$×LM = Masking vs.\ LayerMask/LayerMask (Ablated), LM×LMA = LayerMask vs.\ LayerMask (Ablated). All values are averaged over thresholds $\{0, 0.1, 0.3, 0.5, 0.7, 0.9\}$.}
\label{tab:dice}
\resizebox{\linewidth}{!}{%
\begin{tabular}{llrrrrrrrrrrrrrrr}
\toprule
 &  & \multicolumn{5}{c}{\textbf{ImageNet-1k}} & \multicolumn{5}{c}{\textbf{ImageNet-1k v2}} & \multicolumn{5}{c}{\textbf{PASCAL-VOC}} \\
\cmidrule(lr){3-7} \cmidrule(lr){8-12} \cmidrule(lr){13-17}
\textbf{Method} & \textbf{Model} & $\mathcal{M}$×$\mathcal{M}$ & AD×$\mathcal{M}$ & AD×LM & $\mathcal{M}$×LM & LM×LMA & $\mathcal{M}$×$\mathcal{M}$ & AD×$\mathcal{M}$ & AD×LM & $\mathcal{M}$×LM & LM×LMA & $\mathcal{M}$×$\mathcal{M}$ & AD×$\mathcal{M}$ & AD×LM & $\mathcal{M}$×LM & LM×LMA \\
\midrule
\multirow{3}{*}{\textbf{GradCAM}} & EfficientNet-v2 & 0.815 & 0.694 & 0.590 & 0.506 & 0.359 & 0.814 & 0.683 & 0.597 & 0.500 & 0.353 & 0.608 & 0.465 & 0.456 & 0.360 & 0.327 \\
 & RegNet & 0.802 & 0.433 & 0.789 & 0.370 & 0.713 & 0.790 & 0.452 & 0.791 & 0.386 & 0.716 & 0.742 & 0.067 & 0.614 & 0.039 & 0.353 \\
 & ResNet & 0.721 & 0.675 & 0.121 & 0.111 & 0.340 & 0.738 & 0.676 & 0.115 & 0.108 & 0.329 & 0.695 & 0.525 & 0.116 & 0.070 & 0.480 \\
\midrule
\multirow{3}{*}{\textbf{LIME}} & EfficientNet-v2 & 0.750 & 0.474 & 0.422 & 0.347 & 0.247 & 0.782 & 0.461 & 0.428 & 0.325 & 0.266 & 0.579 & 0.274 & 0.219 & 0.141 & 0.042 \\
 & RegNet & 0.627 & 0.416 & 0.482 & 0.334 & 0.336 & 0.640 & 0.431 & 0.472 & 0.334 & 0.330 & 0.676 & 0.174 & 0.306 & 0.085 & 0.167 \\
 & ResNet & 0.625 & 0.508 & 0.124 & 0.120 & 0.080 & 0.632 & 0.521 & 0.137 & 0.124 & 0.081 & 0.571 & 0.355 & 0.063 & 0.052 & 0.094 \\
\midrule
\multirow{3}{*}{\textbf{ReX}} & EfficientNet-v2 & 0.647 & 0.479 & 0.428 & 0.301 & 0.237 & 0.654 & 0.471 & 0.419 & 0.288 & 0.236 & 0.314 & 0.201 & 0.158 & 0.094 & 0.044 \\
 & RegNet & 0.561 & 0.233 & 0.529 & 0.200 & 0.392 & 0.560 & 0.235 & 0.524 & 0.201 & 0.351 & 0.374 & 0.054 & 0.375 & 0.031 & 0.217 \\
 & ResNet & 0.498 & 0.439 & 0.237 & 0.182 & 0.142 & 0.526 & 0.463 & 0.240 & 0.186 & 0.138 & 0.394 & 0.245 & 0.083 & 0.085 & 0.031 \\
\bottomrule
\end{tabular}
}%
\end{table}

\begin{table}[h]
\centering
\caption{Mean pairwise \textbf{IoU} scores between explanation strategy groups across methods (GradCAM, LIME, ReX), models, and datasets. $\mathcal{M}$×$\mathcal{M}$ = Masking vs.\ Masking (Max, Zero, Mean, Min), AD×$\mathcal{M}$ = AD vs.\ Masking, AD×LM = AD vs.\ LayerMask/LayerMask (Ablated), $\mathcal{M}$×LM = Masking vs.\ LayerMask/LayerMask (Ablated), LM×LMA = LayerMask vs.\ LayerMask (Ablated). All values are averaged over thresholds $\{0, 0.1, 0.3, 0.5, 0.7, 0.9\}$.}
\label{tab:iou}
\resizebox{\textwidth}{!}{%
\begin{tabular}{llrrrrrrrrrrrrrrr}
\toprule
 &  & \multicolumn{5}{c}{\textbf{ImageNet-1k}} & \multicolumn{5}{c}{\textbf{ImageNet-1k v2}} & \multicolumn{5}{c}{\textbf{PASCAL-VOC}} \\
\cmidrule(lr){3-7} \cmidrule(lr){8-12} \cmidrule(lr){13-17}
\textbf{Method} & \textbf{Model} & $\mathcal{M}$×$\mathcal{M}$ & AD×$\mathcal{M}$ & AD×LM & $\mathcal{M}$×LM & LM×LMA & $\mathcal{M}$×$\mathcal{M}$ & AD×$\mathcal{M}$ & AD×LM & $\mathcal{M}$×LM & LM×LMA & $\mathcal{M}$×$\mathcal{M}$ & AD×$\mathcal{M}$ & AD×LM & $\mathcal{M}$×LM & LM×LMA \\
\midrule
\multirow{3}{*}{\textbf{GradCAM}} & EfficientNet-v2 & 0.734 & 0.578 & 0.482 & 0.404 & 0.248 & 0.733 & 0.563 & 0.484 & 0.396 & 0.243 & 0.522 & 0.373 & 0.375 & 0.284 & 0.254 \\
 & RegNet & 0.714 & 0.311 & 0.691 & 0.261 & 0.589 & 0.704 & 0.329 & 0.693 & 0.270 & 0.595 & 0.642 & 0.036 & 0.564 & 0.020 & 0.345 \\
 & ResNet & 0.629 & 0.568 & 0.075 & 0.070 & 0.267 & 0.652 & 0.573 & 0.071 & 0.068 & 0.249 & 0.606 & 0.428 & 0.087 & 0.047 & 0.399 \\
\midrule
\multirow{3}{*}{\textbf{LIME}} & EfficientNet-v2 & 0.656 & 0.337 & 0.301 & 0.236 & 0.159 & 0.689 & 0.324 & 0.303 & 0.212 & 0.172 & 0.486 & 0.186 & 0.161 & 0.095 & 0.024 \\
 & RegNet & 0.513 & 0.283 & 0.342 & 0.218 & 0.209 & 0.525 & 0.294 & 0.330 & 0.216 & 0.204 & 0.574 & 0.107 & 0.214 & 0.049 & 0.103 \\
 & ResNet & 0.515 & 0.366 & 0.073 & 0.072 & 0.050 & 0.519 & 0.374 & 0.081 & 0.074 & 0.049 & 0.469 & 0.253 & 0.039 & 0.034 & 0.080 \\
\midrule
\multirow{3}{*}{\textbf{ReX}} & EfficientNet-v2 & 0.515 & 0.347 & 0.308 & 0.203 & 0.148 & 0.525 & 0.342 & 0.297 & 0.193 & 0.149 & 0.232 & 0.133 & 0.109 & 0.061 & 0.027 \\
 & RegNet & 0.424 & 0.152 & 0.391 & 0.128 & 0.266 & 0.421 & 0.153 & 0.391 & 0.128 & 0.233 & 0.267 & 0.031 & 0.295 & 0.017 & 0.158 \\
 & ResNet & 0.369 & 0.312 & 0.150 & 0.113 & 0.083 & 0.391 & 0.329 & 0.152 & 0.116 & 0.079 & 0.289 & 0.163 & 0.050 & 0.052 & 0.017 \\
\bottomrule
\end{tabular}
}%
\end{table}

 \label{appendix:overlap}


%% file: Plots/updated_robustness_gradcam.tex
\begin{figure}[!htbp]
    \centering
    \begin{center}

    \begin{subfigure}{0.24\textwidth}    
        \centering
        \begin{tikzpicture}
            \begin{axis}[grid, ylabel=ResNet50, width=\linewidth, xticklabel=\empty, height=3.0cm, ymin=0, ymax=1, xmin=0, xmax=1,xtick={0, 0.25, 0.5, 0.75, 1},title=ImageNet-1k]
                \addplot[color=min_color, error bars/.cd, y dir=both, y explicit]
                    coordinates {(0, 0.0703) +- (0, 0.0154) (0.2, 0.2308) +- (0, 0.0273) (0.4, 0.5679) +- (0, 0.0321) (0.6, 0.7319) +- (0, 0.0302) (0.8, 0.8079) +- (0, 0.0273) (1, 0.9026) +- (0, 0.0198)};
                \addplot[color=avg_color, error bars/.cd, y dir=both, y explicit]
                    coordinates {(0, 0.0271) +- (0, 0.0063) (0.2, 0.1325) +- (0, 0.0209) (0.4, 0.3089) +- (0, 0.0309) (0.6, 0.4661) +- (0, 0.0347) (0.8, 0.6145) +- (0, 0.0345) (1, 0.7431) +- (0, 0.0297)};
                \addplot[color=zero_color, error bars/.cd, y dir=both, y explicit]
                    coordinates {(0, 0.0246) +- (0, 0.0088) (0.2, 0.0690) +- (0, 0.0150) (0.4, 0.2227) +- (0, 0.0275) (0.6, 0.3328) +- (0, 0.0323) (0.8, 0.4667) +- (0, 0.0339) (1, 0.6146) +- (0, 0.0351)};
                \addplot[color=max_color, error bars/.cd, y dir=both, y explicit]
                    coordinates {(0, 0.1237) +- (0, 0.0159) (0.2, 0.3897) +- (0, 0.0257) (0.4, 0.7338) +- (0, 0.0283) (0.6, 0.8633) +- (0, 0.0212) (0.8, 0.8895) +- (0, 0.0213) (1, 0.9283) +- (0, 0.0185)};
                \addplot[color=ad_color, error bars/.cd, y dir=both, y explicit]
                    coordinates {(0, 0.1552) +- (0, 0.0294) (0.2, 0.4536) +- (0, 0.0350) (0.4, 0.7137) +- (0, 0.0321) (0.6, 0.8322) +- (0, 0.0250) (0.8, 0.8528) +- (0, 0.0269) (1, 0.9068) +- (0, 0.0212)};
                \addplot[color=lma_color, error bars/.cd, y dir=both, y explicit]
                    coordinates {(0, 0.0000) +- (0, 0.0000) (0.2, 0.0000) +- (0, 0.0000) (0.4, 0.0000) +- (0, 0.0000) (0.6, 0.0000) +- (0, 0.0000) (0.8, 0.0000) +- (0, 0.0000) (1, 0.0000) +- (0, 0.0000)};
                \addplot[color=lm_color, error bars/.cd, y dir=both, y explicit]
                    coordinates {(0, 0.0000) +- (0, 0.0000) (0.2, 0.0000) +- (0, 0.0000) (0.4, 0.0000) +- (0, 0.0000) (0.6, 0.0000) +- (0, 0.0000) (0.8, 0.0000) +- (0, 0.0000) (1, 0.0000) +- (0, 0.0000)};
            \end{axis}
        \end{tikzpicture}
    \end{subfigure}\hspace{6mm}
    \begin{subfigure}{0.24\textwidth}
        \centering
        \begin{tikzpicture}
            \begin{axis}[grid, yticklabel=\empty,  width=\linewidth, xticklabel=\empty, height=3.0cm, ymin=0, ymax=1, xmin=0, xmax=1,xtick={0, 0.25, 0.5, 0.75, 1},title=ImageNet-v2]
                \addplot[color=min_color, error bars/.cd, y dir=both, y explicit]
                    coordinates {(0, 0.0660) +- (0, 0.0115) (0.2, 0.2144) +- (0, 0.0228) (0.4, 0.5506) +- (0, 0.0307) (0.6, 0.7379) +- (0, 0.0274) (0.8, 0.8194) +- (0, 0.0254) (1, 0.8637) +- (0, 0.0236)};
                \addplot[color=avg_color, error bars/.cd, y dir=both, y explicit]
                    coordinates {(0, 0.0462) +- (0, 0.0029) (0.2, 0.1949) +- (0, 0.0192) (0.4, 0.3166) +- (0, 0.0295) (0.6, 0.4961) +- (0, 0.0345) (0.8, 0.6209) +- (0, 0.0360) (1, 0.7437) +- (0, 0.0308)};
                \addplot[color=zero_color, error bars/.cd, y dir=both, y explicit]
                    coordinates {(0, 0.0457) +- (0, 0.0029) (0.2, 0.1790) +- (0, 0.0122) (0.4, 0.1833) +- (0, 0.0242) (0.6, 0.3300) +- (0, 0.0316) (0.8, 0.4829) +- (0, 0.0353) (1, 0.5890) +- (0, 0.0358)};
                \addplot[color=max_color, error bars/.cd, y dir=both, y explicit]
                    coordinates {(0, 0.1892) +- (0, 0.0147) (0.2, 0.4121) +- (0, 0.0250) (0.4, 0.7051) +- (0, 0.0278) (0.6, 0.8404) +- (0, 0.0218) (0.8, 0.8729) +- (0, 0.0232) (1, 0.9172) +- (0, 0.0184)};
                \addplot[color=ad_color, error bars/.cd, y dir=both, y explicit]
                    coordinates {(0, 0.5459) +- (0, 0.0261) (0.2, 0.7296) +- (0, 0.0356) (0.4, 0.7048) +- (0, 0.0318) (0.6, 0.8387) +- (0, 0.0256) (0.8, 0.9185) +- (0, 0.0177) (1, 0.9250) +- (0, 0.0174)};
                \addplot[color=lma_color, error bars/.cd, y dir=both, y explicit]
                    coordinates {(0, 0.0000) +- (0, 0.0000) (0.2, 0.1215) +- (0, 0.0000) (0.4, 0.0449) +- (0, 0.0000) (0.6, 0.0000) +- (0, 0.0000) (0.8, 0.0000) +- (0, 0.0000) (1, 0.0000) +- (0, 0.0000)};
                \addplot[color=lm_color, error bars/.cd, y dir=both, y explicit]
                    coordinates {(0, 0.0000) +- (0, 0.0000) (0.2, 0.7730) +- (0, 0.0000) (0.4, 0.5388) +- (0, 0.0000) (0.6, 0.0000) +- (0, 0.0000) (0.8, 0.0000) +- (0, 0.0000) (1, 0.0000) +- (0, 0.0000)};
            \end{axis}
        \end{tikzpicture}
    \end{subfigure}\hspace{6mm}
    \begin{subfigure}{0.24\textwidth}
        \centering
        \begin{tikzpicture}
            \begin{axis}[grid, yticklabel=\empty,xticklabel=\empty, width=\linewidth,  height=3.0cm, ymin=0, ymax=1, xmin=0, xmax=1,
            xtick={0, 0.25, 0.5, 0.75, 1}, title=PASCAL-VOC]
                \addplot[color=min_color, error bars/.cd, y dir=both, y explicit]
                    coordinates {(0, 0.0000) +- (0, 0.0000) (0.2, 0.0416) +- (0, 0.0142) (0.4, 0.2386) +- (0, 0.0330) (0.6, 0.4550) +- (0, 0.0380) (0.8, 0.6018) +- (0, 0.0345) (1, 0.7361) +- (0, 0.0286)};
                \addplot[color=avg_color, error bars/.cd, y dir=both, y explicit]
                    coordinates {(0, 0.0000) +- (0, 0.0000) (0.2, 0.0551) +- (0, 0.0139) (0.4, 0.1358) +- (0, 0.0193) (0.6, 0.2601) +- (0, 0.0260) (0.8, 0.3566) +- (0, 0.0279) (1, 0.5478) +- (0, 0.0281)};
                \addplot[color=zero_color, error bars/.cd, y dir=both, y explicit]
                    coordinates {(0, 0.0000) +- (0, 0.0000) (0.2, 0.0526) +- (0, 0.0134) (0.4, 0.0859) +- (0, 0.0150) (0.6, 0.1808) +- (0, 0.0228) (0.8, 0.3166) +- (0, 0.0281) (1, 0.4694) +- (0, 0.0302)};
                \addplot[color=max_color, error bars/.cd, y dir=both, y explicit]
                    coordinates {(0, 0.0000) +- (0, 0.0000) (0.2, 0.0457) +- (0, 0.0121) (0.4, 0.1789) +- (0, 0.0217) (0.6, 0.2993) +- (0, 0.0261) (0.8, 0.4380) +- (0, 0.0287) (1, 0.6229) +- (0, 0.0271)};
                \addplot[color=ad_color, error bars/.cd, y dir=both, y explicit]
                    coordinates {(0, 0.0000) +- (0, 0.0000) (0.2, 0.1696) +- (0, 0.0368) (0.4, 0.4444) +- (0, 0.0378) (0.6, 0.5550) +- (0, 0.0357) (0.8, 0.6446) +- (0, 0.0326) (1, 0.7342) +- (0, 0.0294)};
                \addplot[color=lma_color, error bars/.cd, y dir=both, y explicit]
                    coordinates {(0, 0.0000) +- (0, 0.0000) (0.2, 0.0000) +- (0, 0.0178) (0.4, 0.0000) +- (0, 0.0085) (0.6, 0.0000) +- (0, 0.0002) (0.8, 0.0005) +- (0, 0.0001) (1, 0.0018) +- (0, 0.0048)};
                \addplot[color=lm_color, error bars/.cd, y dir=both, y explicit]
                    coordinates {(0, 0.0000) +- (0, 0.0000) (0.2, 0.0255) +- (0, 0.0000) (0.4, 0.0085) +- (0, 0.0000) (0.6, 0.0002) +- (0, 0.0000) (0.8, 0.0001) +- (0, 0.0003) (1, 0.0063) +- (0, 0.0006)};
            \end{axis}
        \end{tikzpicture}
    \end{subfigure}

    \begin{subfigure}{0.24\textwidth}
        \centering
        \begin{tikzpicture}
            \begin{axis}[grid, ylabel=RegNetY,xticklabel=\empty, width=\linewidth, height=3.0cm, ymin=0, ymax=1, xmin=0, xmax=1,xtick={0, 0.25, 0.5, 0.75, 1}]
                \addplot[color=min_color, error bars/.cd, y dir=both, y explicit]
                    coordinates {(0, 0.0374) +- (0, 0.0105) (0.2, 0.2149) +- (0, 0.0271) (0.4, 0.4187) +- (0, 0.0299) (0.6, 0.4959) +- (0, 0.0335) (0.8, 0.6502) +- (0, 0.0322) (1, 0.7968) +- (0, 0.0247)};
                \addplot[color=avg_color, error bars/.cd, y dir=both, y explicit]
                    coordinates {(0, 0.0493) +- (0, 0.0105) (0.2, 0.2213) +- (0, 0.0237) (0.4, 0.3860) +- (0, 0.0290) (0.6, 0.5565) +- (0, 0.0303) (0.8, 0.6898) +- (0, 0.0286) (1, 0.8089) +- (0, 0.0234)};
                \addplot[color=zero_color, error bars/.cd, y dir=both, y explicit]
                    coordinates {(0, 0.0379) +- (0, 0.0095) (0.2, 0.2033) +- (0, 0.0230) (0.4, 0.3297) +- (0, 0.0291) (0.6, 0.4778) +- (0, 0.0310) (0.8, 0.6752) +- (0, 0.0274) (1, 0.7810) +- (0, 0.0255)};
                \addplot[color=max_color, error bars/.cd, y dir=both, y explicit]
                    coordinates {(0, 0.1456) +- (0, 0.0218) (0.2, 0.4451) +- (0, 0.0292) (0.4, 0.6076) +- (0, 0.0302) (0.6, 0.7076) +- (0, 0.0297) (0.8, 0.8235) +- (0, 0.0234) (1, 0.9109) +- (0, 0.0161)};
                \addplot[color=ad_color, error bars/.cd, y dir=both, y explicit]
                    coordinates {(0, 0.5382) +- (0, 0.0436) (0.2, 0.7743) +- (0, 0.0355) (0.4, 0.8362) +- (0, 0.0315) (0.6, 0.8912) +- (0, 0.0254) (0.8, 0.9179) +- (0, 0.0225) (1, 0.9203) +- (0, 0.0211)};
                \addplot[color=lma_color, error bars/.cd, y dir=both, y explicit]
                    coordinates {(0, 0.0669) +- (0, 0.0426) (0.2, 0.1867) +- (0, 0.0335) (0.4, 0.3728) +- (0, 0.0260) (0.6, 0.4538) +- (0, 0.0187) (0.8, 0.5149) +- (0, 0.0213) (1, 0.6717) +- (0, 0.0157)};
                \addplot[color=lm_color, error bars/.cd, y dir=both, y explicit]
                    coordinates {(0, 0.6121) +- (0, 0.0228) (0.2, 0.8005) +- (0, 0.0343) (0.4, 0.8803) +- (0, 0.0419) (0.6, 0.9408) +- (0, 0.0430) (0.8, 0.9271) +- (0, 0.0431) (1, 0.9558) +- (0, 0.0394)};
            \end{axis}
        \end{tikzpicture}
    \end{subfigure}\hspace{6mm}
    \begin{subfigure}{0.24\textwidth}
        \centering
        \begin{tikzpicture}
            \begin{axis}[grid, yticklabel=\empty,xticklabel=\empty, width=\linewidth, height=3.0cm, ymin=0, ymax=1, xmin=0, xmax=1,xtick={0, 0.25, 0.5, 0.75, 1}]
                \addplot[color=min_color, error bars/.cd, y dir=both, y explicit]
                    coordinates {(0, 0.0660) +- (0, 0.0159) (0.2, 0.2144) +- (0, 0.0244) (0.4, 0.4149) +- (0, 0.0323) (0.6, 0.5637) +- (0, 0.0318) (0.8, 0.6732) +- (0, 0.0295) (1, 0.8037) +- (0, 0.0263)};
                \addplot[color=avg_color, error bars/.cd, y dir=both, y explicit]
                    coordinates {(0, 0.0462) +- (0, 0.0110) (0.2, 0.1949) +- (0, 0.0226) (0.4, 0.3650) +- (0, 0.0285) (0.6, 0.4861) +- (0, 0.0305) (0.8, 0.6399) +- (0, 0.0305) (1, 0.7995) +- (0, 0.0241)};
                \addplot[color=zero_color, error bars/.cd, y dir=both, y explicit]
                    coordinates {(0, 0.0457) +- (0, 0.0122) (0.2, 0.1790) +- (0, 0.0224) (0.4, 0.3641) +- (0, 0.0292) (0.6, 0.4531) +- (0, 0.0306) (0.8, 0.5957) +- (0, 0.0313) (1, 0.7675) +- (0, 0.0264)};
                \addplot[color=max_color, error bars/.cd, y dir=both, y explicit]
                    coordinates {(0, 0.1892) +- (0, 0.0238) (0.2, 0.4121) +- (0, 0.0287) (0.4, 0.5766) +- (0, 0.0305) (0.6, 0.7416) +- (0, 0.0273) (0.8, 0.8448) +- (0, 0.0213) (1, 0.9113) +- (0, 0.0167)};
                \addplot[color=ad_color, error bars/.cd, y dir=both, y explicit]
                    coordinates {(0, 0.5459) +- (0, 0.0422) (0.2, 0.7160) +- (0, 0.0374) (0.4, 0.8338) +- (0, 0.0305) (0.6, 0.8605) +- (0, 0.0286) (0.8, 0.8996) +- (0, 0.0246) (1, 0.9521) +- (0, 0.0169)};
                \addplot[color=lma_color, error bars/.cd, y dir=both, y explicit]
                    coordinates {(0, 0.0316) +- (0, 0.0397) (0.2, 0.1215) +- (0, 0.0335) (0.4, 0.2179) +- (0, 0.0285) (0.6, 0.3329) +- (0, 0.0266) (0.8, 0.4774) +- (0, 0.0233) (1, 0.6518) +- (0, 0.0238)};
                \addplot[color=lm_color, error bars/.cd, y dir=both, y explicit]
                    coordinates {(0, 0.6277) +- (0, 0.0155) (0.2, 0.7730) +- (0, 0.0277) (0.4, 0.8569) +- (0, 0.0357) (0.6, 0.8789) +- (0, 0.0408) (0.8, 0.9088) +- (0, 0.0431) (1, 0.9091) +- (0, 0.0405)};
            \end{axis}
        \end{tikzpicture}
    \end{subfigure}\hspace{6mm}
    \begin{subfigure}{0.24\textwidth}
        \centering
        \begin{tikzpicture}
            \begin{axis}[grid, yticklabel=\empty,xticklabel=\empty,height=3.0cm,  width=\linewidth, ymin=0, ymax=1, xmin=0, xmax=1,xtick={0, 0.25, 0.5, 0.75, 1}]
                \addplot[color=min_color, error bars/.cd, y dir=both, y explicit]
                    coordinates {(0, 0.0000) +- (0, 0.0000) (0.2, 0.0000) +- (0, 0.0000) (0.4, 0.1133) +- (0, 0.0925) (0.6, 0.4467) +- (0, 0.1374) (0.8, 0.3700) +- (0, 0.2220) (1, 0.6350) +- (0, 0.1874)};
                \addplot[color=avg_color, error bars/.cd, y dir=both, y explicit]
                    coordinates {(0, 0.0000) +- (0, 0.0000) (0.2, 0.0067) +- (0, 0.0054) (0.4, 0.2267) +- (0, 0.1851) (0.6, 0.3533) +- (0, 0.1559) (0.8, 0.4067) +- (0, 0.1785) (1, 0.7400) +- (0, 0.0141)};
                \addplot[color=zero_color, error bars/.cd, y dir=both, y explicit]
                    coordinates {(0, 0.0000) +- (0, 0.0000) (0.2, 0.0067) +- (0, 0.0054) (0.4, 0.0033) +- (0, 0.0027) (0.6, 0.2700) +- (0, 0.1687) (0.8, 0.3367) +- (0, 0.1872) (1, 0.3400) +- (0, 0.1626)};
                \addplot[color=max_color, error bars/.cd, y dir=both, y explicit]
                    coordinates {(0, 0.0000) +- (0, 0.0000) (0.2, 0.0467) +- (0, 0.0381) (0.4, 0.2200) +- (0, 0.1302) (0.6, 0.6133) +- (0, 0.0687) (0.8, 0.5600) +- (0, 0.1320) (1, 0.7350) +- (0, 0.0460)};
                \addplot[color=ad_color, error bars/.cd, y dir=both, y explicit]
                    coordinates {(0, 0.0000) +- (0, 0.0000) (0.2, 0.3300) +- (0, 0.2694) (0.4, 0.3300) +- (0, 0.2694) (0.6, 0.9867) +- (0, 0.0072) (0.8, 0.9867) +- (0, 0.0072) (1, 0.9950) +- (0, 0.0035)};
                \addplot[color=lma_color, error bars/.cd, y dir=both, y explicit]
                    coordinates {(0, 0.0000) +- (0, 0.0000) (0.2, 0.0000) +- (0, 0.2667) (0.4, 0.0000) +- (0, 0.2667) (0.6, 0.0000) +- (0, 0.2681) (0.8, 0.0000) +- (0, 0.0072) (1, 0.0050) +- (0, 0.0035)};
                \addplot[color=lm_color, error bars/.cd, y dir=both, y explicit]
                    coordinates {(0, 0.0000) +- (0, 0.0000) (0.2, 0.3267) +- (0, 0.0000) (0.4, 0.3267) +- (0, 0.0000) (0.6, 0.6567) +- (0, 0.0000) (0.8, 0.9767) +- (0, 0.0000) (1, 0.9850) +- (0, 0.0035)};
            \end{axis}
        \end{tikzpicture}
    \end{subfigure}

    \begin{subfigure}{0.24\textwidth}
        \centering
        \begin{tikzpicture}
            \begin{axis}[grid, xlabel=Conf. threshold,  width=\linewidth, ylabel=EfficientNet, height=3.0cm, ymin=0, ymax=1, xmin=0, xmax=1,xtick={0, 0.25, 0.5, 0.75, 1}]
                \addplot[color=min_color, error bars/.cd, y dir=both, y explicit]
                    coordinates {(0, 0.0515) +- (0, 0.0257) (0.2, 0.1568) +- (0, 0.0266) (0.4, 0.2647) +- (0, 0.0310) (0.6, 0.4125) +- (0, 0.0335) (0.8, 0.5150) +- (0, 0.0335) (1, 0.6717) +- (0, 0.0320)};
                \addplot[color=avg_color, error bars/.cd, y dir=both, y explicit]
                    coordinates {(0, 0.0485) +- (0, 0.0166) (0.2, 0.1442) +- (0, 0.0200) (0.4, 0.3296) +- (0, 0.0256) (0.6, 0.4531) +- (0, 0.0273) (0.8, 0.5903) +- (0, 0.0285) (1, 0.7256) +- (0, 0.0262)};
                \addplot[color=zero_color, error bars/.cd, y dir=both, y explicit]
                    coordinates {(0, 0.0226) +- (0, 0.0085) (0.2, 0.1491) +- (0, 0.0183) (0.4, 0.3027) +- (0, 0.0234) (0.6, 0.4403) +- (0, 0.0268) (0.8, 0.5869) +- (0, 0.0284) (1, 0.7384) +- (0, 0.0252)};
                \addplot[color=max_color, error bars/.cd, y dir=both, y explicit]
                    coordinates {(0, 0.0185) +- (0, 0.0082) (0.2, 0.1267) +- (0, 0.0215) (0.4, 0.2419) +- (0, 0.0259) (0.6, 0.3971) +- (0, 0.0299) (0.8, 0.5583) +- (0, 0.0310) (1, 0.6791) +- (0, 0.0294)};
                \addplot[color=ad_color, error bars/.cd, y dir=both, y explicit]
                    coordinates {(0, 0.0387) +- (0, 0.0235) (0.2, 0.3902) +- (0, 0.0384) (0.4, 0.5847) +- (0, 0.0368) (0.6, 0.7296) +- (0, 0.0311) (0.8, 0.8014) +- (0, 0.0287) (1, 0.8568) +- (0, 0.0241)};
                \addplot[color=lma_color, error bars/.cd, y dir=both, y explicit]
                    coordinates {(0, 0.0000) +- (0, 0.0218) (0.2, 0.0168) +- (0, 0.0323) (0.4, 0.0228) +- (0, 0.0372) (0.6, 0.0449) +- (0, 0.0368) (0.8, 0.0673) +- (0, 0.0363) (1, 0.1423) +- (0, 0.0338)};
                \addplot[color=lm_color, error bars/.cd, y dir=both, y explicit]
                    coordinates {(0, 0.0345) +- (0, 0.0000) (0.2, 0.2267) +- (0, 0.0117) (0.4, 0.4177) +- (0, 0.0130) (0.6, 0.5388) +- (0, 0.0179) (0.8, 0.6121) +- (0, 0.0217) (1, 0.7217) +- (0, 0.0302)};
            \end{axis}
        \end{tikzpicture}
    \end{subfigure}\hspace{6mm}
    \begin{subfigure}{0.24\textwidth}
        \centering
        \begin{tikzpicture}
            \begin{axis}[grid, xlabel=Conf. Threshold,  width=\linewidth, yticklabel=\empty,height=3.0cm, ymin=0, ymax=1, xmin=0, xmax=1,xtick={0, 0.25, 0.5, 0.75, 1}]
                \addplot[color=min_color, error bars/.cd, y dir=both, y explicit]
                    coordinates {(0, 0.0000) +- (0, 0.0015) (0.2, 0.0649) +- (0, 0.0290) (0.4, 0.2185) +- (0, 0.0342) (0.6, 0.3635) +- (0, 0.0350) (0.8, 0.4768) +- (0, 0.0344) (1, 0.5912) +- (0, 0.0324)};
                \addplot[color=avg_color, error bars/.cd, y dir=both, y explicit]
                    coordinates {(0, 0.0000) +- (0, 0.0149) (0.2, 0.1109) +- (0, 0.0207) (0.4, 0.2489) +- (0, 0.0246) (0.6, 0.3777) +- (0, 0.0276) (0.8, 0.4946) +- (0, 0.0286) (1, 0.5749) +- (0, 0.0264)};
                \addplot[color=zero_color, error bars/.cd, y dir=both, y explicit]
                    coordinates {(0, 0.0000) +- (0, 0.0119) (0.2, 0.0933) +- (0, 0.0185) (0.4, 0.2529) +- (0, 0.0243) (0.6, 0.3438) +- (0, 0.0271) (0.8, 0.4653) +- (0, 0.0262) (1, 0.5794) +- (0, 0.0231)};
                \addplot[color=max_color, error bars/.cd, y dir=both, y explicit]
                    coordinates {(0, 0.0000) +- (0, 0.0279) (0.2, 0.0637) +- (0, 0.0252) (0.4, 0.1471) +- (0, 0.0314) (0.6, 0.2535) +- (0, 0.0323) (0.8, 0.4380) +- (0, 0.0321) (1, 0.5055) +- (0, 0.0309)};
                \addplot[color=ad_color, error bars/.cd, y dir=both, y explicit]
                    coordinates {(0, 0.0000) +- (0, 0.0296) (0.2, 0.0675) +- (0, 0.0400) (0.4, 0.3573) +- (0, 0.0365) (0.6, 0.5254) +- (0, 0.0319) (0.8, 0.6446) +- (0, 0.0302) (1, 0.7745) +- (0, 0.0271)};
                \addplot[color=lma_color, error bars/.cd, y dir=both, y explicit]
                    coordinates {(0, 0.0000) +- (0, 0.0285) (0.2, 0.0000) +- (0, 0.0320) (0.4, 0.0000) +- (0, 0.0356) (0.6, 0.0154) +- (0, 0.0359) (0.8, 0.0005) +- (0, 0.0333) (1, 0.1281) +- (0, 0.0336)};
                \addplot[color=lm_color, error bars/.cd, y dir=both, y explicit]
                    coordinates {(0, 0.0000) +- (0, 0.0000) (0.2, 0.0888) +- (0, 0.0081) (0.4, 0.2140) +- (0, 0.0106) (0.6, 0.3065) +- (0, 0.0099) (0.8, 0.4419) +- (0, 0.0226) (1, 0.5965) +- (0, 0.0241)};
            \end{axis}
        \end{tikzpicture}
    \end{subfigure}\hspace{6mm}
    \begin{subfigure}{0.24\textwidth}
        \centering
        \begin{tikzpicture}
            \begin{axis}[grid, xlabel=Conf. Threshold, yticklabel=\empty,  width=\linewidth, height=3.0cm, ymin=0, ymax=1, xmin=0, xmax=1,xtick={0, 0.25, 0.5, 0.75, 1}]
                \addplot[color=min_color, error bars/.cd, y dir=both, y explicit]
                    coordinates {(0, 0.0000) +- (0, 0.0000) (0.2, 0.0416) +- (0, 0.0227) (0.4, 0.1133) +- (0, 0.0292) (0.6, 0.4467) +- (0, 0.0332) (0.8, 0.3700) +- (0, 0.0329) (1, 0.0017) +- (0, 0.0308)};
                \addplot[color=avg_color, error bars/.cd, y dir=both, y explicit]
                    coordinates {(0, 0.0000) +- (0, 0.0000) (0.2, 0.0551) +- (0, 0.0253) (0.4, 0.2267) +- (0, 0.0258) (0.6, 0.3533) +- (0, 0.0278) (0.8, 0.4067) +- (0, 0.0269) (1, 0.0374) +- (0, 0.0266)};
                \addplot[color=zero_color, error bars/.cd, y dir=both, y explicit]
                    coordinates {(0, 0.0000) +- (0, 0.0000) (0.2, 0.0526) +- (0, 0.0232) (0.4, 0.0033) +- (0, 0.0254) (0.6, 0.2700) +- (0, 0.0267) (0.8, 0.3367) +- (0, 0.0258) (1, 0.0333) +- (0, 0.0256)};
                \addplot[color=max_color, error bars/.cd, y dir=both, y explicit]
                    coordinates {(0, 0.0000) +- (0, 0.0000) (0.2, 0.0457) +- (0, 0.0208) (0.4, 0.2200) +- (0, 0.0228) (0.6, 0.6133) +- (0, 0.0269) (0.8, 0.5600) +- (0, 0.0266) (1, 0.0626) +- (0, 0.0283)};
                \addplot[color=ad_color, error bars/.cd, y dir=both, y explicit]
                    coordinates {(0, 0.0000) +- (0, 0.0000) (0.2, 0.1696) +- (0, 0.0260) (0.4, 0.3300) +- (0, 0.0360) (0.6, 0.9867) +- (0, 0.0354) (0.8, 0.9867) +- (0, 0.0316) (1, 0.0428) +- (0, 0.0274)};
                \addplot[color=lma_color, error bars/.cd, y dir=both, y explicit]
                    coordinates {(0, 0.0000) +- (0, 0.0000) (0.2, 0.0000) +- (0, 0.0248) (0.4, 0.0000) +- (0, 0.0277) (0.6, 0.0000) +- (0, 0.0301) (0.8, 0.0000) +- (0, 0.0323) (1, 0.0000) +- (0, 0.0306)};
                \addplot[color=lm_color, error bars/.cd, y dir=both, y explicit]
                    coordinates {(0, 0.0000) +- (0, 0.0000) (0.2, 0.0000) +- (0, 0.0000) (0.4, 0.0000) +- (0, 0.0000) (0.6, 0.3267) +- (0, 0.0108) (0.8, 0.9767) +- (0, 0.0078) (1, 0.0255) +- (0, 0.0240)};
            \end{axis}
        \end{tikzpicture}
    \end{subfigure}

    \vspace{1mm}
    \centering
    \begin{tikzpicture}
        \node[draw,color=black,rounded corners] (legend) {\quad \textcolor{ad_color}{\textbf{ConvAD}} \quad \textcolor{min_color}{\textbf{Min}} \quad \textcolor{max_color}{\textbf{Max}} \quad \textcolor{avg_color}{\textbf{Avg}} \quad \textcolor{zero_color}{\textbf{Zero}} \quad \textcolor{lma_color}{\textbf{Layer Mask (Ablated)}} \quad \textcolor{lm_color}{\textbf{Layer Mask}} \quad};
    \end{tikzpicture}
    \end{center}
    \caption{$\beta$-robustness (rows) of \gradcam calculated explanations against \ood backgrounds on ImageNet-1k, ImageNet-v2 and PASCAL-VOC for ResNet-50, RegNet-Y and EfficientNet-v2. \convad is consistently the best performing strategy, often across all thresholds and models. Error bars represents Standard Error of the Mean (SEM).}
    \label{fig:gradcam}
\end{figure}

%% file: Plots/updated_robustness_lime.tex
\begin{figure}[!htbp]
\centering
\begin{center}
\begin{subfigure}{0.24\textwidth}
    \centering
    \begin{tikzpicture}
        \begin{axis}[grid, ylabel=ResNet50, xticklabel=\empty, height=3.0cm,   width=\linewidth, ymin=0, ymax=1, xmin=0, xmax=1,xtick={0, 0.2, 0.4, 0.6, 0.8, 1},title=ImageNet-1k]
            \addplot[color=ad_color, error bars/.cd, y dir=both, y explicit] coordinates {(0.0, 0.1188) +- (0.0, 0.0227) (0.2, 0.2875) +- (0.0, 0.0267) (0.4, 0.4475) +- (0.0, 0.0295) (0.6, 0.5883) +- (0.0, 0.0296) (0.8, 0.6892) +- (0.0, 0.0274) (1.0, 0.7429) +- (0.0, 0.0261)};
            \addplot[color=zero_color, error bars/.cd, y dir=both, y explicit] coordinates {(0.0, 0.0323) +- (0.0, 0.0113) (0.2, 0.0448) +- (0.0, 0.0097) (0.4, 0.1079) +- (0.0, 0.0168) (0.6, 0.1534) +- (0.0, 0.0215) (0.8, 0.2105) +- (0.0, 0.0259) (1.0, 0.2925) +- (0.0, 0.0305)};
            \addplot[color=lm_color, error bars/.cd, y dir=both, y explicit] coordinates {(0.0, 0.0000) +- (0.0, 0.0000) (0.2, 0.0000) +- (0.0, 0.0000) (0.4, 0.0000) +- (0.0, 0.0000) (0.6, 0.0000) +- (0.0, 0.0000) (0.8, 0.0000) +- (0.0, 0.0000) (1.0, 0.0000) +- (0.0, 0.0000)};
            \addplot[color=lma_color, error bars/.cd, y dir=both, y explicit] coordinates {(0.0, 0.0123) +- (0.0, 0.0102) (0.2, 0.0000) +- (0.0, 0.0000) (0.4, 0.0000) +- (0.0, 0.0000) (0.6, 0.0000) +- (0.0, 0.0000) (0.8, 0.0000) +- (0.0, 0.0000) (1.0, 0.0000) +- (0.0, 0.0000)};
            \addplot[color=min_color, error bars/.cd, y dir=both, y explicit] coordinates {(0.0, 0.0697) +- (0.0, 0.0189) (0.2, 0.1776) +- (0.0, 0.0233) (0.4, 0.3279) +- (0.0, 0.0294) (0.6, 0.4349) +- (0.0, 0.0329) (0.8, 0.5086) +- (0.0, 0.0348) (1.0, 0.5847) +- (0.0, 0.0350)};
            \addplot[color=max_color, error bars/.cd, y dir=both, y explicit] coordinates {(0.0, 0.1033) +- (0.0, 0.0235) (0.2, 0.4003) +- (0.0, 0.0328) (0.4, 0.6030) +- (0.0, 0.0308) (0.6, 0.6593) +- (0.0, 0.0332) (0.8, 0.7422) +- (0.0, 0.0294) (1.0, 0.7845) +- (0.0, 0.0266)};
            \addplot[color=avg_color, error bars/.cd, y dir=both, y explicit] coordinates {(0.0, 0.0293) +- (0.0, 0.0108) (0.2, 0.0727) +- (0.0, 0.0136) (0.4, 0.1463) +- (0.0, 0.0197) (0.6, 0.2381) +- (0.0, 0.0252) (0.8, 0.3333) +- (0.0, 0.0302) (1.0, 0.3786) +- (0.0, 0.0305)};
        \end{axis}
    \end{tikzpicture}
\end{subfigure}\hspace{6mm}
\begin{subfigure}{0.24\textwidth}
    \centering
    \begin{tikzpicture}
        \begin{axis}[grid, yticklabel=\empty, xticklabel=\empty,  width=\linewidth, height=3.0cm, ymin=0, ymax=1, xmin=0, xmax=1,xtick={0, 0.2, 0.4, 0.6, 0.8, 1},title=ImageNet-v2]
            \addplot[color=ad_color, error bars/.cd, y dir=both, y explicit] coordinates {(0.0, 0.1502) +- (0.0, 0.0242) (0.2, 0.2676) +- (0.0, 0.0249) (0.4, 0.4101) +- (0.0, 0.0276) (0.6, 0.5275) +- (0.0, 0.0282) (0.8, 0.6044) +- (0.0, 0.0287) (1.0, 0.6797) +- (0.0, 0.0269)};
            \addplot[color=zero_color, error bars/.cd, y dir=both, y explicit] coordinates {(0.0, 0.0257) +- (0.0, 0.0097) (0.2, 0.0536) +- (0.0, 0.0119) (0.4, 0.0884) +- (0.0, 0.0150) (0.6, 0.1204) +- (0.0, 0.0187) (0.8, 0.1630) +- (0.0, 0.0220) (1.0, 0.2711) +- (0.0, 0.0279)};
            \addplot[color=lm_color, error bars/.cd, y dir=both, y explicit] coordinates {(0.0, 0.0000) +- (0.0, 0.0000) (0.2, 0.0000) +- (0.0, 0.0000) (0.4, 0.0000) +- (0.0, 0.0000) (0.6, 0.0000) +- (0.0, 0.0000) (0.8, 0.0000) +- (0.0, 0.0000) (1.0, 0.0000) +- (0.0, 0.0000)};
            \addplot[color=lma_color, error bars/.cd, y dir=both, y explicit] coordinates {(0.0, 0.0000) +- (0.0, 0.0000) (0.2, 0.0000) +- (0.0, 0.0000) (0.4, 0.0000) +- (0.0, 0.0000) (0.6, 0.0000) +- (0.0, 0.0000) (0.8, 0.0000) +- (0.0, 0.0000) (1.0, 0.0000) +- (0.0, 0.0000)};
            \addplot[color=min_color, error bars/.cd, y dir=both, y explicit] coordinates {(0.0, 0.0560) +- (0.0, 0.0172) (0.2, 0.1612) +- (0.0, 0.0235) (0.4, 0.2944) +- (0.0, 0.0282) (0.6, 0.4270) +- (0.0, 0.0317) (0.8, 0.4860) +- (0.0, 0.0318) (1.0, 0.6090) +- (0.0, 0.0303)};
            \addplot[color=max_color, error bars/.cd, y dir=both, y explicit] coordinates {(0.0, 0.1240) +- (0.0, 0.0277) (0.2, 0.3944) +- (0.0, 0.0320) (0.4, 0.5402) +- (0.0, 0.0313) (0.6, 0.6664) +- (0.0, 0.0297) (0.8, 0.7567) +- (0.0, 0.0265) (1.0, 0.8253) +- (0.0, 0.0218)};
            \addplot[color=avg_color, error bars/.cd, y dir=both, y explicit] coordinates {(0.0, 0.0135) +- (0.0, 0.0045) (0.2, 0.0831) +- (0.0, 0.0141) (0.4, 0.1498) +- (0.0, 0.0196) (0.6, 0.2196) +- (0.0, 0.0247) (0.8, 0.2753) +- (0.0, 0.0268) (1.0, 0.3668) +- (0.0, 0.0298)};
        \end{axis}
    \end{tikzpicture}
\end{subfigure}\hspace{6mm}
\begin{subfigure}{0.24\textwidth}
    \centering
    \begin{tikzpicture}
        \begin{axis}[grid, yticklabel=\empty,xticklabel=\empty,  width=\linewidth, height=3.0cm, ymin=0, ymax=1, xmin=0, xmax=1, xtick={0, 0.2, 0.4, 0.6, 0.8, 1}, title=PASCAL-VOC]
            \addplot[color=ad_color, error bars/.cd, y dir=both, y explicit] coordinates {(0.0, 0.0000) +- (0.0, 0.0000) (0.2, 0.1082) +- (0.0, 0.0233) (0.4, 0.2066) +- (0.0, 0.0255) (0.6, 0.2767) +- (0.0, 0.0279) (0.8, 0.3707) +- (0.0, 0.0293) (1.0, 0.4567) +- (0.0, 0.0285)};
            \addplot[color=zero_color, error bars/.cd, y dir=both, y explicit] coordinates {(0.0, 0.0000) +- (0.0, 0.0000) (0.2, 0.0474) +- (0.0, 0.0172) (0.4, 0.0765) +- (0.0, 0.0180) (0.6, 0.1063) +- (0.0, 0.0192) (0.8, 0.1814) +- (0.0, 0.0253) (1.0, 0.2072) +- (0.0, 0.0239)};
            \addplot[color=lm_color, error bars/.cd, y dir=both, y explicit] coordinates {(0.0, 0.6900) +- (0.0, 0.0047) (0.2, 0.0000) +- (0.0, 0.0000) (0.4, 0.0002) +- (0.0, 0.0002) (0.6, 0.0002) +- (0.0, 0.0002) (0.8, 0.0002) +- (0.0, 0.0002) (1.0, 0.0007) +- (0.0, 0.0003)};
            \addplot[color=lma_color, error bars/.cd, y dir=both, y explicit] coordinates {(0.0, 0.0000) +- (0.0, 0.0000) (0.2, 0.0249) +- (0.0, 0.0145) (0.4, 0.0000) +- (0.0, 0.0000) (0.6, 0.0000) +- (0.0, 0.0000) (0.8, 0.0000) +- (0.0, 0.0000) (1.0, 0.0086) +- (0.0, 0.0065)};
            \addplot[color=min_color, error bars/.cd, y dir=both, y explicit] coordinates {(0.0, 0.0000) +- (0.0, 0.0000) (0.2, 0.0359) +- (0.0, 0.0158) (0.4, 0.1912) +- (0.0, 0.0309) (0.6, 0.3574) +- (0.0, 0.0368) (0.8, 0.4690) +- (0.0, 0.0344) (1.0, 0.5521) +- (0.0, 0.0332)};
            \addplot[color=max_color, error bars/.cd, y dir=both, y explicit] coordinates {(0.0, 0.0000) +- (0.0, 0.0000) (0.2, 0.0489) +- (0.0, 0.0173) (0.4, 0.2296) +- (0.0, 0.0306) (0.6, 0.3078) +- (0.0, 0.0308) (0.8, 0.4132) +- (0.0, 0.0306) (1.0, 0.5209) +- (0.0, 0.0298)};
            \addplot[color=avg_color, error bars/.cd, y dir=both, y explicit] coordinates {(0.0, 0.0000) +- (0.0, 0.0000) (0.2, 0.0411) +- (0.0, 0.0157) (0.4, 0.0582) +- (0.0, 0.0133) (0.6, 0.1208) +- (0.0, 0.0202) (0.8, 0.1453) +- (0.0, 0.0213) (1.0, 0.2215) +- (0.0, 0.0239)};
        \end{axis}
    \end{tikzpicture}
\end{subfigure}

\begin{subfigure}{0.24\textwidth}
    \centering
    \begin{tikzpicture}
        \begin{axis}[grid, ylabel=RegNetY,xticklabel=\empty,height=3.0cm,  width=\linewidth, ymin=0, ymax=1, xmin=0, xmax=1,xtick={0, 0.2, 0.4, 0.6, 0.8, 1}]
            \addplot[color=ad_color, error bars/.cd, y dir=both, y explicit] coordinates {(0.0, 0.1556) +- (0.0, 0.0269) (0.2, 0.4431) +- (0.0, 0.0335) (0.4, 0.5092) +- (0.0, 0.0350) (0.6, 0.6091) +- (0.0, 0.0334) (0.8, 0.6865) +- (0.0, 0.0301) (1.0, 0.7680) +- (0.0, 0.0269)};
            \addplot[color=zero_color, error bars/.cd, y dir=both, y explicit] coordinates {(0.0, 0.0381) +- (0.0, 0.0118) (0.2, 0.0606) +- (0.0, 0.0123) (0.4, 0.1066) +- (0.0, 0.0177) (0.6, 0.1329) +- (0.0, 0.0201) (0.8, 0.1931) +- (0.0, 0.0232) (1.0, 0.2906) +- (0.0, 0.0276)};
            \addplot[color=lm_color, error bars/.cd, y dir=both, y explicit] coordinates {(0.0, 0.0259) +- (0.0, 0.0132) (0.2, 0.0304) +- (0.0, 0.0139) (0.4, 0.0492) +- (0.0, 0.0183) (0.6, 0.0429) +- (0.0, 0.0172) (0.8, 0.0963) +- (0.0, 0.0251) (1.0, 0.2002) +- (0.0, 0.0338)};
            \addplot[color=lma_color, error bars/.cd, y dir=both, y explicit] coordinates {(0.0, 0.2144) +- (0.0, 0.0314) (0.2, 0.4308) +- (0.0, 0.0346) (0.4, 0.5127) +- (0.0, 0.0351) (0.6, 0.5815) +- (0.0, 0.0345) (0.8, 0.6193) +- (0.0, 0.0348) (1.0, 0.7540) +- (0.0, 0.0281)};
            \addplot[color=min_color, error bars/.cd, y dir=both, y explicit] coordinates {(0.0, 0.0428) +- (0.0, 0.0134) (0.2, 0.2728) +- (0.0, 0.0299) (0.4, 0.4888) +- (0.0, 0.0336) (0.6, 0.6034) +- (0.0, 0.0320) (0.8, 0.7411) +- (0.0, 0.0279) (1.0, 0.8339) +- (0.0, 0.0221)};
            \addplot[color=max_color, error bars/.cd, y dir=both, y explicit] coordinates {(0.0, 0.0647) +- (0.0, 0.0171) (0.2, 0.3362) +- (0.0, 0.0317) (0.4, 0.5695) +- (0.0, 0.0329) (0.6, 0.6932) +- (0.0, 0.0294) (0.8, 0.7603) +- (0.0, 0.0275) (1.0, 0.8763) +- (0.0, 0.0196)};
            \addplot[color=avg_color, error bars/.cd, y dir=both, y explicit] coordinates {(0.0, 0.0299) +- (0.0, 0.0093) (0.2, 0.0890) +- (0.0, 0.0129) (0.4, 0.1441) +- (0.0, 0.0163) (0.6, 0.2221) +- (0.0, 0.0213) (0.8, 0.3174) +- (0.0, 0.0267) (1.0, 0.4945) +- (0.0, 0.0292)};
        \end{axis}
    \end{tikzpicture}
\end{subfigure}\hspace{6mm}
\begin{subfigure}{0.24\textwidth}
    \centering
    \begin{tikzpicture}
        \begin{axis}[grid,  width=\linewidth, yticklabel=\empty,xticklabel=\empty,height=3.0cm, ymin=0, ymax=1, xmin=0, xmax=1,xtick={0, 0.2, 0.4, 0.6, 0.8, 1}]
            \addplot[color=ad_color, error bars/.cd, y dir=both, y explicit] coordinates {(0.0, 0.1539) +- (0.0, 0.0258) (0.2, 0.3437) +- (0.0, 0.0320) (0.4, 0.4782) +- (0.0, 0.0341) (0.6, 0.5821) +- (0.0, 0.0328) (0.8, 0.6152) +- (0.0, 0.0331) (1.0, 0.7063) +- (0.0, 0.0303)};
            \addplot[color=zero_color, error bars/.cd, y dir=both, y explicit] coordinates {(0.0, 0.0219) +- (0.0, 0.0069) (0.2, 0.0563) +- (0.0, 0.0106) (0.4, 0.0914) +- (0.0, 0.0146) (0.6, 0.1334) +- (0.0, 0.0182) (0.8, 0.1882) +- (0.0, 0.0224) (1.0, 0.2871) +- (0.0, 0.0274)};
            \addplot[color=lm_color, error bars/.cd, y dir=both, y explicit] coordinates {(0.0, 0.0234) +- (0.0, 0.0134) (0.2, 0.0137) +- (0.0, 0.0097) (0.4, 0.0075) +- (0.0, 0.0075) (0.6, 0.0075) +- (0.0, 0.0075) (0.8, 0.0495) +- (0.0, 0.0184) (1.0, 0.1398) +- (0.0, 0.0297)};
            \addplot[color=lma_color, error bars/.cd, y dir=both, y explicit] coordinates {(0.0, 0.1878) +- (0.0, 0.0276) (0.2, 0.3570) +- (0.0, 0.0330) (0.4, 0.4767) +- (0.0, 0.0352) (0.6, 0.5460) +- (0.0, 0.0352) (0.8, 0.6010) +- (0.0, 0.0337) (1.0, 0.7091) +- (0.0, 0.0300)};
            \addplot[color=min_color, error bars/.cd, y dir=both, y explicit] coordinates {(0.0, 0.0483) +- (0.0, 0.0146) (0.2, 0.3121) +- (0.0, 0.0307) (0.4, 0.4497) +- (0.0, 0.0333) (0.6, 0.6027) +- (0.0, 0.0331) (0.8, 0.6792) +- (0.0, 0.0316) (1.0, 0.7787) +- (0.0, 0.0269)};
            \addplot[color=max_color, error bars/.cd, y dir=both, y explicit] coordinates {(0.0, 0.0965) +- (0.0, 0.0223) (0.2, 0.3222) +- (0.0, 0.0313) (0.4, 0.5683) +- (0.0, 0.0310) (0.6, 0.6864) +- (0.0, 0.0287) (0.8, 0.7587) +- (0.0, 0.0277) (1.0, 0.8489) +- (0.0, 0.0224)};
            \addplot[color=avg_color, error bars/.cd, y dir=both, y explicit] coordinates {(0.0, 0.0134) +- (0.0, 0.0045) (0.2, 0.1033) +- (0.0, 0.0150) (0.4, 0.1624) +- (0.0, 0.0190) (0.6, 0.2211) +- (0.0, 0.0227) (0.8, 0.3101) +- (0.0, 0.0267) (1.0, 0.4371) +- (0.0, 0.0300)};
        \end{axis}
    \end{tikzpicture}
\end{subfigure}\hspace{6mm}
\begin{subfigure}{0.24\textwidth}
    \centering
    \begin{tikzpicture}
        \begin{axis}[grid,  width=\linewidth, yticklabel=\empty,xticklabel=\empty,height=3.0cm, ymin=0, ymax=1, xmin=0, xmax=1,xtick={0, 0.2, 0.4, 0.6, 0.8, 1}]
            \addplot[color=ad_color, error bars/.cd, y dir=both, y explicit] coordinates {(0.0, 0.0000) +- (0.0, 0.0000) (0.2, 0.3505) +- (0.0, 0.0427) (0.4, 0.7170) +- (0.0, 0.0365) (0.6, 0.8019) +- (0.0, 0.0314) (0.8, 0.8518) +- (0.0, 0.0274) (1.0, 0.8824) +- (0.0, 0.0255)};
            \addplot[color=zero_color, error bars/.cd, y dir=both, y explicit] coordinates {(0.0, 0.0000) +- (0.0, 0.0000) (0.2, 0.0806) +- (0.0, 0.0160) (0.4, 0.1350) +- (0.0, 0.0181) (0.6, 0.2034) +- (0.0, 0.0206) (0.8, 0.2962) +- (0.0, 0.0238) (1.0, 0.3948) +- (0.0, 0.0245)};
            \addplot[color=lm_color, error bars/.cd, y dir=both, y explicit] coordinates {(0.0, 0.0000) +- (0.0, 0.0000) (0.2, 0.0000) +- (0.0, 0.0000) (0.4, 0.0376) +- (0.0, 0.0164) (0.6, 0.0513) +- (0.0, 0.0188) (0.8, 0.0760) +- (0.0, 0.0224) (1.0, 0.1688) +- (0.0, 0.0318)};
            \addplot[color=lma_color, error bars/.cd, y dir=both, y explicit] coordinates {(0.0, 0.0000) +- (0.0, 0.0000) (0.2, 0.1543) +- (0.0, 0.0330) (0.4, 0.6112) +- (0.0, 0.0414) (0.6, 0.7965) +- (0.0, 0.0326) (0.8, 0.8447) +- (0.0, 0.0289) (1.0, 0.8974) +- (0.0, 0.0233)};
            \addplot[color=min_color, error bars/.cd, y dir=both, y explicit] coordinates {(0.0, 0.0000) +- (0.0, 0.0000) (0.2, 0.0542) +- (0.0, 0.0142) (0.4, 0.1535) +- (0.0, 0.0222) (0.6, 0.2258) +- (0.0, 0.0249) (0.8, 0.3285) +- (0.0, 0.0279) (1.0, 0.5037) +- (0.0, 0.0285)};
            \addplot[color=max_color, error bars/.cd, y dir=both, y explicit] coordinates {(0.0, 0.5700) +- (0.0, 0.0141) (0.2, 0.0908) +- (0.0, 0.0166) (0.4, 0.1977) +- (0.0, 0.0209) (0.6, 0.3490) +- (0.0, 0.0260) (0.8, 0.4712) +- (0.0, 0.0268) (1.0, 0.6200) +- (0.0, 0.0251)};
            \addplot[color=avg_color, error bars/.cd, y dir=both, y explicit] coordinates {(0.0, 0.0000) +- (0.0, 0.0000) (0.2, 0.0334) +- (0.0, 0.0105) (0.4, 0.1007) +- (0.0, 0.0150) (0.6, 0.2141) +- (0.0, 0.0200) (0.8, 0.2985) +- (0.0, 0.0228) (1.0, 0.3925) +- (0.0, 0.0243)};
        \end{axis}
    \end{tikzpicture}
\end{subfigure}

\begin{subfigure}{0.24\textwidth}
    \centering
    \begin{tikzpicture}
        \begin{axis}[grid,  width=\linewidth, xlabel=Conf. threshold, ylabel=EfficientNet, height=3.0cm, ymin=0, ymax=1, xmin=0, xmax=1,xtick={0, 0.2, 0.4, 0.6, 0.8, 1}]
            \addplot[color=ad_color, error bars/.cd, y dir=both, y explicit] coordinates {(0.0, 0.0000) +- (0.0, 0.0000) (0.2, 0.2767) +- (0.0, 0.0544) (0.4, 0.3155) +- (0.0, 0.0545) (0.6, 0.5204) +- (0.0, 0.0449) (0.8, 0.7635) +- (0.0, 0.0431) (1.0, 0.8678) +- (0.0, 0.0305)};
            \addplot[color=zero_color, error bars/.cd, y dir=both, y explicit] coordinates {(0.0, 0.0000) +- (0.0, 0.0229) (0.2, 0.2002) +- (0.0, 0.0388) (0.4, 0.3310) +- (0.0, 0.0407) (0.6, 0.4318) +- (0.0, 0.0427) (0.8, 0.5053) +- (0.0, 0.0426) (1.0, 0.6224) +- (0.0, 0.0433)};
            \addplot[color=lm_color, error bars/.cd, y dir=both, y explicit] coordinates {(0.0, 0.6900) +- (0.0, 0.0000) (0.2, 0.0002) +- (0.0, 0.0000) (0.4, 0.0000) +- (0.0, 0.0198) (0.6, 0.0200) +- (0.0, 0.0198) (0.8, 0.0800) +- (0.0, 0.0383) (1.0, 0.1400) +- (0.0, 0.0490)};
            \addplot[color=lma_color, error bars/.cd, y dir=both, y explicit] coordinates {(0.0, 0.0000) +- (0.0, 0.0430) (0.2, 0.0000) +- (0.0, 0.0388) (0.4, 0.1468) +- (0.0, 0.0483) (0.6, 0.2643) +- (0.0, 0.0525) (0.8, 0.4486) +- (0.0, 0.0560) (1.0, 0.5367) +- (0.0, 0.0553)};
            \addplot[color=min_color, error bars/.cd, y dir=both, y explicit] coordinates {(0.0, 0.0000) +- (0.0, 0.0430) (0.2, 0.3334) +- (0.0, 0.0589) (0.4, 0.4265) +- (0.0, 0.0590) (0.6, 0.5543) +- (0.0, 0.0553) (0.8, 0.6680) +- (0.0, 0.0487) (1.0, 0.7982) +- (0.0, 0.0397)};
            \addplot[color=max_color, error bars/.cd, y dir=both, y explicit] coordinates {(0.0, 0.0000) +- (0.0, 0.0332) (0.2, 0.2983) +- (0.0, 0.0508) (0.4, 0.4953) +- (0.0, 0.0502) (0.6, 0.6786) +- (0.0, 0.0471) (0.8, 0.7367) +- (0.0, 0.0426) (1.0, 0.8194) +- (0.0, 0.0382)};
            \addplot[color=avg_color, error bars/.cd, y dir=both, y explicit] coordinates {(0.0, 0.0000) +- (0.0, 0.0238) (0.2, 0.1572) +- (0.0, 0.0366) (0.4, 0.2796) +- (0.0, 0.0420) (0.6, 0.3727) +- (0.0, 0.0415) (0.8, 0.4357) +- (0.0, 0.0443) (1.0, 0.5792) +- (0.0, 0.0417)};
        \end{axis}
    \end{tikzpicture}
\end{subfigure}\hspace{6mm}
\begin{subfigure}{0.24\textwidth}
    \centering
    \begin{tikzpicture}
        \begin{axis}[grid,  width=\linewidth, xlabel=Conf. Threshold, yticklabel=\empty,height=3.0cm, ymin=0, ymax=1, xmin=0, xmax=1,xtick={0, 0.2, 0.4, 0.6, 0.8, 1}]
            \addplot[color=ad_color, error bars/.cd, y dir=both, y explicit] coordinates {(0.0, 0.0260) +- (0.0, 0.0198) (0.2, 0.3686) +- (0.0, 0.0365) (0.4, 0.5655) +- (0.0, 0.0348) (0.6, 0.7120) +- (0.0, 0.0298) (0.8, 0.7718) +- (0.0, 0.0273) (1.0, 0.8439) +- (0.0, 0.0236)};
            \addplot[color=zero_color, error bars/.cd, y dir=both, y explicit] coordinates {(0.0, 0.0312) +- (0.0, 0.0189) (0.2, 0.1394) +- (0.0, 0.0180) (0.4, 0.2680) +- (0.0, 0.0217) (0.6, 0.4033) +- (0.0, 0.0244) (0.8, 0.5111) +- (0.0, 0.0252) (1.0, 0.6199) +- (0.0, 0.0263)};
            \addplot[color=lm_color, error bars/.cd, y dir=both, y explicit] coordinates {(0.0, 0.0000) +- (0.0, 0.0000) (0.2, 0.0000) +- (0.0, 0.0000) (0.4, 0.0000) +- (0.0, 0.0000) (0.6, 0.0000) +- (0.0, 0.0000) (0.8, 0.0074) +- (0.0, 0.0074) (1.0, 0.0732) +- (0.0, 0.0223)};
            \addplot[color=lma_color, error bars/.cd, y dir=both, y explicit] coordinates {(0.0, 0.0276) +- (0.0, 0.0198) (0.2, 0.2042) +- (0.0, 0.0284) (0.4, 0.3449) +- (0.0, 0.0319) (0.6, 0.4495) +- (0.0, 0.0328) (0.8, 0.4856) +- (0.0, 0.0339) (1.0, 0.6090) +- (0.0, 0.0331)};
            \addplot[color=min_color, error bars/.cd, y dir=both, y explicit] coordinates {(0.0, 0.0378) +- (0.0, 0.0206) (0.2, 0.2636) +- (0.0, 0.0327) (0.4, 0.4826) +- (0.0, 0.0359) (0.6, 0.6510) +- (0.0, 0.0329) (0.8, 0.7346) +- (0.0, 0.0296) (1.0, 0.7844) +- (0.0, 0.0280)};
            \addplot[color=max_color, error bars/.cd, y dir=both, y explicit] coordinates {(0.0, 0.0478) +- (0.0, 0.0263) (0.2, 0.2685) +- (0.0, 0.0309) (0.4, 0.4838) +- (0.0, 0.0314) (0.6, 0.5954) +- (0.0, 0.0299) (0.8, 0.6508) +- (0.0, 0.0298) (1.0, 0.7478) +- (0.0, 0.0272)};
            \addplot[color=avg_color, error bars/.cd, y dir=both, y explicit] coordinates {(0.0, 0.0182) +- (0.0, 0.0113) (0.2, 0.1118) +- (0.0, 0.0165) (0.4, 0.2419) +- (0.0, 0.0216) (0.6, 0.3611) +- (0.0, 0.0239) (0.8, 0.4662) +- (0.0, 0.0258) (1.0, 0.5820) +- (0.0, 0.0280)};
        \end{axis}
    \end{tikzpicture}
\end{subfigure}\hspace{6mm}
\begin{subfigure}{0.24\textwidth}
    \centering
    \begin{tikzpicture}
        \begin{axis}[grid,  width=\linewidth, xlabel=Conf. Threshold, yticklabel=\empty, height=3.0cm, ymin=0, ymax=1, xmin=0, xmax=1,xtick={0, 0.2, 0.4, 0.6, 0.8, 1}]
            \addplot[color=ad_color, error bars/.cd, y dir=both, y explicit] coordinates {(0.0, 0.0000) +- (0.0, 0.0000) (0.2, 0.3505) +- (0.0, 0.1252) (0.4, 0.7170) +- (0.0, 0.0991) (0.6, 0.3400) +- (0.0, 0.1258) (0.8, 0.6611) +- (0.0, 0.1230) (1.0, 0.8711) +- (0.0, 0.0470)};
            \addplot[color=zero_color, error bars/.cd, y dir=both, y explicit] coordinates {(0.0, 0.0000) +- (0.0, 0.0000) (0.2, 0.0806) +- (0.0, 0.0505) (0.4, 0.1350) +- (0.0, 0.0348) (0.6, 0.2044) +- (0.0, 0.0653) (0.8, 0.2967) +- (0.0, 0.0818) (1.0, 0.4011) +- (0.0, 0.0883)};
            \addplot[color=lm_color, error bars/.cd, y dir=both, y explicit] coordinates {(0.0, 0.0000) +- (0.0, 0.0000) (0.2, 0.0000) +- (0.0, 0.0000) (0.4, 0.0376) +- (0.0, 0.0000) (0.6, 0.0000) +- (0.0, 0.0000) (0.8, 0.0000) +- (0.0, 0.0000) (1.0, 0.0000) +- (0.0, 0.0000)};
            \addplot[color=lma_color, error bars/.cd, y dir=both, y explicit] coordinates {(0.0, 0.0000) +- (0.0, 0.0000) (0.2, 0.1543) +- (0.0, 0.0089) (0.4, 0.6112) +- (0.0, 0.1296) (0.6, 0.3022) +- (0.0, 0.1285) (0.8, 0.2767) +- (0.0, 0.1280) (1.0, 0.4433) +- (0.0, 0.1215)};
            \addplot[color=min_color, error bars/.cd, y dir=both, y explicit] coordinates {(0.0, 0.0495) +- (0.0, 0.0000) (0.2, 0.0542) +- (0.0, 0.0894) (0.4, 0.1535) +- (0.0, 0.0157) (0.6, 0.2711) +- (0.0, 0.1096) (0.8, 0.4556) +- (0.0, 0.1008) (1.0, 0.5967) +- (0.0, 0.1122)};
            \addplot[color=max_color, error bars/.cd, y dir=both, y explicit] coordinates {(0.0, 0.0486) +- (0.0, 0.0000) (0.2, 0.0908) +- (0.0, 0.0215) (0.4, 0.1977) +- (0.0, 0.0579) (0.6, 0.3978) +- (0.0, 0.1043) (0.8, 0.4122) +- (0.0, 0.1008) (1.0, 0.5767) +- (0.0, 0.0863)};
            \addplot[color=avg_color, error bars/.cd, y dir=both, y explicit] coordinates {(0.0, 0.0500) +- (0.0, 0.0000) (0.2, 0.0334) +- (0.0, 0.0730) (0.4, 0.0582) +- (0.0, 0.0538) (0.6, 0.1208) +- (0.0, 0.0635) (0.8, 0.1453) +- (0.0, 0.0748) (1.0, 0.2215) +- (0.0, 0.0916)};
        \end{axis}
    \end{tikzpicture}
\end{subfigure}

\vspace{1mm}
\centering
\begin{tikzpicture}
    \node[draw,color=black,rounded corners] (legend) {
        \quad \textcolor{ad_color}{\textbf{ConvAD}} 
        \quad \textcolor{min_color}{\textbf{Min}}
        \quad \textcolor{max_color}{\textbf{Max}} 
        \quad \textcolor{avg_color}{\textbf{Avg}} 
        \quad \textcolor{zero_color}{\textbf{Zero}} 
        \quad \textcolor{lma_color}{\textbf{Layer Mask (Ablated)}} 
        \quad \textcolor{lm_color}{\textbf{Layer Mask}} 
    };
\end{tikzpicture}
\end{center}

\caption{$\beta$-robustness (rows) of \lime calculated explanations against \ood backgrounds on ImageNet-1k, ImageNet-v2 and PASCAL-VOC for ResNet-50, RegNet-Y and EfficientNet-v2. Error bars represents Standard Error of the Mean (SEM).}
\label{fig:limerob}

\end{figure}

%% file: Plots/updated_robustness_rex_iid.tex
\begin{figure}[!htbp]
\centering
\begin{center}
    \begin{subfigure}{0.24\textwidth}
        \centering
        \begin{tikzpicture}
            \begin{axis}[grid, width=\linewidth, ylabel=ResNet50, xticklabel=\empty, height=3.0cm, ymin=0, ymax=1, xmin=0, xmax=1,xtick={0, 0.25, 0.5, 0.75, 1},title=ImageNet-1k]
                \addplot[color=min_color, error bars/.cd, y dir=both, y explicit] coordinates {(0, 0.0026) +- (0, 0.0006) (0.2, 0.0045) +- (0, 0.0009) (0.4, 0.0178) +- (0, 0.0034) (0.6, 0.0463) +- (0, 0.0079) (0.8, 0.1005) +- (0, 0.0153) (1, 0.1928) +- (0, 0.0227) };
                \addplot[color=avg_color, error bars/.cd, y dir=both, y explicit] coordinates {(0, 0.0014) +- (0, 0.0004) (0.2, 0.0024) +- (0, 0.0004) (0.4, 0.0039) +- (0, 0.0007) (0.6, 0.0098) +- (0, 0.0023) (0.8, 0.0287) +- (0, 0.0093) (1, 0.0711) +- (0, 0.0150) };
                \addplot[color=zero_color, error bars/.cd, y dir=both, y explicit] coordinates {(0, 0.0021) +- (0, 0.0004) (0.2, 0.0024) +- (0, 0.0004) (0.4, 0.0029) +- (0, 0.0005) (0.6, 0.0050) +- (0, 0.0012) (0.8, 0.0128) +- (0, 0.0028) (1, 0.0427) +- (0, 0.0103) };
                \addplot[color=max_color, error bars/.cd, y dir=both, y explicit] coordinates {(0, 0.0034) +- (0, 0.0007) (0.2, 0.0085) +- (0, 0.0012) (0.4, 0.0224) +- (0, 0.0032) (0.6, 0.0454) +- (0, 0.0062) (0.8, 0.1028) +- (0, 0.0128) (1, 0.2027) +- (0, 0.0237) };
                \addplot[color=ad_color, error bars/.cd, y dir=both, y explicit] coordinates {(0, 0.0029) +- (0, 0.0005) (0.2, 0.0118) +- (0, 0.0027) (0.4, 0.0374) +- (0, 0.0061) (0.6, 0.0884) +- (0, 0.0107) (0.8, 0.1605) +- (0, 0.0174) (1, 0.2487) +- (0, 0.0238) };
                \addplot[color=lma_color, error bars/.cd, y dir=both, y explicit] coordinates {(0, 0.0000) +- (0, 0.0000) (0.2, 0.0512) +- (0, 0.0196) (0.4, 0.2742) +- (0, 0.0393) (0.6, 0.4287) +- (0, 0.0425) (0.8, 0.4918) +- (0, 0.0436) (1, 0.7107) +- (0, 0.0367) };
                \addplot[color=lm_color, error bars/.cd, y dir=both, y explicit] coordinates {(0, 0.0004) +- (0, 0.0002) (0.2, 0.0015) +- (0, 0.0008) (0.4, 0.0016) +- (0, 0.0004) (0.6, 0.0039) +- (0, 0.0011) (0.8, 0.0056) +- (0, 0.0013) (1, 0.0152) +- (0, 0.0054) };
            \end{axis}
        \end{tikzpicture}
    \end{subfigure}\hspace{6mm}
    \begin{subfigure}{0.24\textwidth}
        \centering
        \begin{tikzpicture}
            \begin{axis}[grid, width=\linewidth, yticklabel=\empty, xticklabel=\empty, height=3.0cm, ymin=0, ymax=1, xmin=0, xmax=1,xtick={0, 0.25, 0.5, 0.75, 1},title=ImageNet-v2]
                \addplot[color=min_color, error bars/.cd, y dir=both, y explicit] coordinates {(0, 0.0000) +- (0, 0.0003) (0.2, 0.0103) +- (0, 0.0006) (0.4, 0.0456) +- (0, 0.0030) (0.6, 0.0672) +- (0, 0.0078) (0.8, 0.0803) +- (0, 0.0162) (1, 0.1912) +- (0, 0.0244) };
                \addplot[color=avg_color, error bars/.cd, y dir=both, y explicit] coordinates {(0, 0.0000) +- (0, 0.0003) (0.2, 0.0200) +- (0, 0.0004) (0.4, 0.0604) +- (0, 0.0008) (0.6, 0.0827) +- (0, 0.0019) (0.8, 0.1014) +- (0, 0.0106) (1, 0.0719) +- (0, 0.0138) };
                \addplot[color=zero_color, error bars/.cd, y dir=both, y explicit] coordinates {(0, 0.0000) +- (0, 0.0002) (0.2, 0.0354) +- (0, 0.0003) (0.4, 0.0684) +- (0, 0.0004) (0.6, 0.0900) +- (0, 0.0008) (0.8, 0.1047) +- (0, 0.0040) (1, 0.0571) +- (0, 0.0120) };
                \addplot[color=max_color, error bars/.cd, y dir=both, y explicit] coordinates {(0, 0.0000) +- (0, 0.0005) (0.2, 0.0311) +- (0, 0.0014) (0.4, 0.0575) +- (0, 0.0074) (0.6, 0.0756) +- (0, 0.0104) (0.8, 0.0975) +- (0, 0.0167) (1, 0.2373) +- (0, 0.0245) };
                \addplot[color=ad_color, error bars/.cd, y dir=both, y explicit] coordinates {(0, 0.0000) +- (0, 0.0005) (0.2, 0.0504) +- (0, 0.0021) (0.4, 0.0760) +- (0, 0.0058) (0.6, 0.1137) +- (0, 0.0087) (0.8, 0.1395) +- (0, 0.0159) (1, 0.3087) +- (0, 0.0259) };
                \addplot[color=lma_color, error bars/.cd, y dir=both, y explicit] coordinates {(0, 0.0540) +- (0, 0.0000) (0.2, 0.0000) +- (0, 0.0250) (0.4, 0.0103) +- (0, 0.0410) (0.6, 0.0435) +- (0, 0.0429) (0.8, 0.0657) +- (0, 0.0423) (1, 0.7390) +- (0, 0.0353) };
                \addplot[color=lm_color, error bars/.cd, y dir=both, y explicit] coordinates {(0, 0.0540) +- (0, 0.0000) (0.2, 0.0062) +- (0, 0.0004) (0.4, 0.0369) +- (0, 0.0035) (0.6, 0.0590) +- (0, 0.0044) (0.8, 0.0820) +- (0, 0.0046) (1, 0.0185) +- (0, 0.0053) };
            \end{axis}
        \end{tikzpicture}
    \end{subfigure}\hspace{6mm}
    \begin{subfigure}{0.24\textwidth}
        \centering
        \begin{tikzpicture}
            \begin{axis}[grid, width=\linewidth, yticklabel=\empty,xticklabel=\empty, height=3.0cm, ymin=0, ymax=1, xmin=0, xmax=1,
            xtick={0, 0.25, 0.5, 0.75, 1}, title=PASCAL-VOC]
                \addplot[color=min_color, error bars/.cd, y dir=both, y explicit] coordinates {(0, 0.0032) +- (0, 0.0000) (0.2, 0.0235) +- (0, 0.0034) (0.4, 0.0456) +- (0, 0.0091) (0.6, 0.1119) +- (0, 0.0116) (0.8, 0.1603) +- (0, 0.0135) (1, 0.1942) +- (0, 0.0164) };
                \addplot[color=avg_color, error bars/.cd, y dir=both, y explicit] coordinates {(0, 0.0007) +- (0, 0.0000) (0.2, 0.0310) +- (0, 0.0051) (0.4, 0.0604) +- (0, 0.0064) (0.6, 0.0655) +- (0, 0.0066) (0.8, 0.0738) +- (0, 0.0064) (1, 0.0988) +- (0, 0.0078) };
                \addplot[color=zero_color, error bars/.cd, y dir=both, y explicit] coordinates {(0, 0.0005) +- (0, 0.0000) (0.2, 0.0234) +- (0, 0.0057) (0.4, 0.0684) +- (0, 0.0056) (0.6, 0.0675) +- (0, 0.0061) (0.8, 0.0738) +- (0, 0.0060) (1, 0.0967) +- (0, 0.0080) };
                \addplot[color=max_color, error bars/.cd, y dir=both, y explicit] coordinates {(0, 0.0012) +- (0, 0.0000) (0.2, 0.0251) +- (0, 0.0058) (0.4, 0.0575) +- (0, 0.0063) (0.6, 0.0758) +- (0, 0.0060) (0.8, 0.0893) +- (0, 0.0063) (1, 0.1057) +- (0, 0.0074) };
                \addplot[color=ad_color, error bars/.cd, y dir=both, y explicit] coordinates {(0, 0.0035) +- (0, 0.0000) (0.2, 0.0209) +- (0, 0.0057) (0.4, 0.0760) +- (0, 0.0056) (0.6, 0.0851) +- (0, 0.0056) (0.8, 0.1012) +- (0, 0.0060) (1, 0.1217) +- (0, 0.0079) };
                \addplot[color=lma_color, error bars/.cd, y dir=both, y explicit] coordinates {(0, 0.2432) +- (0, 0.0057) (0.2, 0.0155) +- (0, 0.0000) (0.4, 0.0103) +- (0, 0.0000) (0.6, 0.0435) +- (0, 0.0107) (0.8, 0.0135) +- (0, 0.0097) (1, 0.0860) +- (0, 0.0178) };
                \addplot[color=lm_color, error bars/.cd, y dir=both, y explicit] coordinates {(0, 0.0032) +- (0, 0.0057) (0.2, 0.0146) +- (0, 0.0028) (0.4, 0.0369) +- (0, 0.0034) (0.6, 0.0590) +- (0, 0.0030) (0.8, 0.0132) +- (0, 0.0034) (1, 0.0359) +- (0, 0.0056) };
            \end{axis}
        \end{tikzpicture}
    \end{subfigure}

    \begin{subfigure}{0.24\textwidth}
        \centering
        \begin{tikzpicture}
            \begin{axis}[grid, width=\linewidth, ylabel=RegNetY,xticklabel=\empty,height=3.0cm, ymin=0, ymax=1, xmin=0, xmax=1,xtick={0, 0.25, 0.5, 0.75, 1}]
                \addplot[color=min_color, error bars/.cd, y dir=both, y explicit] coordinates {(0, 0.0013) +- (0, 0.0003) (0.2, 0.0026) +- (0, 0.0004) (0.4, 0.0039) +- (0, 0.0006) (0.6, 0.0065) +- (0, 0.0009) (0.8, 0.0104) +- (0, 0.0014) (1, 0.0389) +- (0, 0.0074) };
                \addplot[color=avg_color, error bars/.cd, y dir=both, y explicit] coordinates {(0, 0.0016) +- (0, 0.0004) (0.2, 0.0021) +- (0, 0.0004) (0.4, 0.0025) +- (0, 0.0005) (0.6, 0.0027) +- (0, 0.0005) (0.8, 0.0042) +- (0, 0.0007) (1, 0.0157) +- (0, 0.0054) };
                \addplot[color=zero_color, error bars/.cd, y dir=both, y explicit] coordinates {(0, 0.0016) +- (0, 0.0004) (0.2, 0.0018) +- (0, 0.0003) (0.4, 0.0023) +- (0, 0.0004) (0.6, 0.0031) +- (0, 0.0005) (0.8, 0.0032) +- (0, 0.0006) (1, 0.0144) +- (0, 0.0065) };
                \addplot[color=max_color, error bars/.cd, y dir=both, y explicit] coordinates {(0, 0.0013) +- (0, 0.0003) (0.2, 0.0027) +- (0, 0.0005) (0.4, 0.0047) +- (0, 0.0007) (0.6, 0.0091) +- (0, 0.0015) (0.8, 0.0111) +- (0, 0.0022) (1, 0.0320) +- (0, 0.0062) };
                \addplot[color=ad_color, error bars/.cd, y dir=both, y explicit] coordinates {(0, 0.0053) +- (0, 0.0011) (0.2, 0.0142) +- (0, 0.0025) (0.4, 0.0239) +- (0, 0.0034) (0.6, 0.0432) +- (0, 0.0063) (0.8, 0.0642) +- (0, 0.0083) (1, 0.1145) +- (0, 0.0140) };
                \addplot[color=lma_color, error bars/.cd, y dir=both, y explicit] coordinates {(0, 0.0788) +- (0, 0.0125) (0.2, 0.1559) +- (0, 0.0177) (0.4, 0.2070) +- (0, 0.0214) (0.6, 0.2754) +- (0, 0.0243) (0.8, 0.3139) +- (0, 0.0275) (1, 0.4736) +- (0, 0.0278) };
                \addplot[color=lm_color, error bars/.cd, y dir=both, y explicit] coordinates {(0, 0.0045) +- (0, 0.0009) (0.2, 0.0123) +- (0, 0.0018) (0.4, 0.0238) +- (0, 0.0033) (0.6, 0.0362) +- (0, 0.0042) (0.8, 0.0544) +- (0, 0.0062) (1, 0.1039) +- (0, 0.0127) };
            \end{axis}
        \end{tikzpicture}
    \end{subfigure}\hspace{6mm}
    \begin{subfigure}{0.24\textwidth}
        \centering
        \begin{tikzpicture}
            \begin{axis}[grid, width=\linewidth, yticklabel=\empty,xticklabel=\empty,height=3.0cm, ymin=0, ymax=1, xmin=0, xmax=1,xtick={0, 0.25, 0.5, 0.75, 1}]
                \addplot[color=min_color, error bars/.cd, y dir=both, y explicit] coordinates {(0, 0.0006) +- (0, 0.0002) (0.2, 0.0235) +- (0, 0.0015) (0.4, 0.0630) +- (0, 0.0037) (0.6, 0.0672) +- (0, 0.0041) (0.8, 0.1603) +- (0, 0.0050) (1, 0.0967) +- (0, 0.0077) };
                \addplot[color=avg_color, error bars/.cd, y dir=both, y explicit] coordinates {(0, 0.0003) +- (0, 0.0002) (0.2, 0.0310) +- (0, 0.0009) (0.4, 0.0533) +- (0, 0.0013) (0.6, 0.0720) +- (0, 0.0014) (0.8, 0.0738) +- (0, 0.0016) (1, 0.0948) +- (0, 0.0016) };
                \addplot[color=zero_color, error bars/.cd, y dir=both, y explicit] coordinates {(0, 0.0008) +- (0, 0.0003) (0.2, 0.0234) +- (0, 0.0009) (0.4, 0.0518) +- (0, 0.0010) (0.6, 0.0652) +- (0, 0.0008) (0.8, 0.0738) +- (0, 0.0011) (1, 0.0963) +- (0, 0.0025) };
                \addplot[color=max_color, error bars/.cd, y dir=both, y explicit] coordinates {(0, 0.0012) +- (0, 0.0005) (0.2, 0.0251) +- (0, 0.0012) (0.4, 0.0597) +- (0, 0.0028) (0.6, 0.0758) +- (0, 0.0015) (0.8, 0.0893) +- (0, 0.0030) (1, 0.1018) +- (0, 0.0037) };
                \addplot[color=ad_color, error bars/.cd, y dir=both, y explicit] coordinates {(0, 0.0035) +- (0, 0.0029) (0.2, 0.0209) +- (0, 0.0039) (0.4, 0.0758) +- (0, 0.0070) (0.6, 0.1217) +- (0, 0.0068) (0.8, 0.1012) +- (0, 0.0138) (1, 0.2887) +- (0, 0.0241) };
                \addplot[color=lma_color, error bars/.cd, y dir=both, y explicit] coordinates {(0, 0.2432) +- (0, 0.0196) (0.2, 0.0155) +- (0, 0.0114) (0.4, 0.0481) +- (0, 0.0516) (0.6, 0.0860) +- (0, 0.0529) (0.8, 0.0135) +- (0, 0.0554) (1, 0.3715) +- (0, 0.0546) };
                \addplot[color=lm_color, error bars/.cd, y dir=both, y explicit] coordinates {(0, 0.0032) +- (0, 0.0011) (0.2, 0.0146) +- (0, 0.0044) (0.4, 0.0639) +- (0, 0.0064) (0.6, 0.0359) +- (0, 0.0078) (0.8, 0.0132) +- (0, 0.0089) (1, 0.3390) +- (0, 0.0135) };
            \end{axis}
        \end{tikzpicture}
    \end{subfigure}\hspace{6mm}
    \begin{subfigure}{0.24\textwidth}
        \centering
        \begin{tikzpicture}
            \begin{axis}[grid, width=\linewidth, yticklabel=\empty,xticklabel=\empty,height=3.0cm, ymin=0, ymax=1, xmin=0, xmax=1,xtick={0, 0.25, 0.5, 0.75, 1}]
                \addplot[color=min_color, error bars/.cd, y dir=both, y explicit] coordinates {(0, 0.0032) +- (0, 0.0000) (0.2, 0.0235) +- (0, 0.0064) (0.4, 0.0456) +- (0, 0.0063) (0.6, 0.0672) +- (0, 0.0064) (0.8, 0.0919) +- (0, 0.0061) (1, 0.0967) +- (0, 0.0064) };
                \addplot[color=avg_color, error bars/.cd, y dir=both, y explicit] coordinates {(0, 0.0007) +- (0, 0.0000) (0.2, 0.0310) +- (0, 0.0055) (0.4, 0.0604) +- (0, 0.0061) (0.6, 0.0827) +- (0, 0.0061) (0.8, 0.1014) +- (0, 0.0061) (1, 0.0948) +- (0, 0.0062) };
                \addplot[color=zero_color, error bars/.cd, y dir=both, y explicit] coordinates {(0, 0.0005) +- (0, 0.0000) (0.2, 0.0234) +- (0, 0.0061) (0.4, 0.0684) +- (0, 0.0058) (0.6, 0.0900) +- (0, 0.0059) (0.8, 0.1047) +- (0, 0.0061) (1, 0.0963) +- (0, 0.0064) };
                \addplot[color=max_color, error bars/.cd, y dir=both, y explicit] coordinates {(0, 0.0012) +- (0, 0.0000) (0.2, 0.0251) +- (0, 0.0060) (0.4, 0.0575) +- (0, 0.0062) (0.6, 0.0756) +- (0, 0.0065) (0.8, 0.0975) +- (0, 0.0068) (1, 0.1018) +- (0, 0.0070) };
                \addplot[color=ad_color, error bars/.cd, y dir=both, y explicit] coordinates {(0, 0.0035) +- (0, 0.0000) (0.2, 0.0209) +- (0, 0.0109) (0.4, 0.0760) +- (0, 0.0153) (0.6, 0.1137) +- (0, 0.0153) (0.8, 0.1395) +- (0, 0.0162) (1, 0.2887) +- (0, 0.0166) };
                \addplot[color=lma_color, error bars/.cd, y dir=both, y explicit] coordinates {(0, 0.2432) +- (0, 0.0000) (0.2, 0.0155) +- (0, 0.0154) (0.4, 0.0103) +- (0, 0.0184) (0.6, 0.0435) +- (0, 0.0208) (0.8, 0.0657) +- (0, 0.0250) (1, 0.3715) +- (0, 0.0280) };
                \addplot[color=lm_color, error bars/.cd, y dir=both, y explicit] coordinates {(0, 0.0032) +- (0, 0.0000) (0.2, 0.0146) +- (0, 0.0122) (0.4, 0.0369) +- (0, 0.0163) (0.6, 0.0590) +- (0, 0.0162) (0.8, 0.0820) +- (0, 0.0166) (1, 0.3390) +- (0, 0.0185) };
            \end{axis}
        \end{tikzpicture}
    \end{subfigure}

    \begin{subfigure}{0.24\textwidth}
        \centering
        \begin{tikzpicture}
            \begin{axis}[grid, width=\linewidth, xlabel=Conf. threshold, ylabel=EfficientNet, height=3.0cm, ymin=0, ymax=1, xmin=0, xmax=1,xtick={0, 0.25, 0.5, 0.75, 1}]
                \addplot[color=min_color, error bars/.cd, y dir=both, y explicit] coordinates {(0, 0.0011) +- (0, 0.0004) (0.2, 0.0047) +- (0, 0.0011) (0.4, 0.0167) +- (0, 0.0020) (0.6, 0.0497) +- (0, 0.0033) (0.8, 0.1038) +- (0, 0.0051) (1, 0.1044) +- (0, 0.0198) };
                \addplot[color=avg_color, error bars/.cd, y dir=both, y explicit] coordinates {(0, 0.0015) +- (0, 0.0005) (0.2, 0.0055) +- (0, 0.0017) (0.4, 0.0037) +- (0, 0.0031) (0.6, 0.0095) +- (0, 0.0035) (0.8, 0.0375) +- (0, 0.0063) (1, 0.0880) +- (0, 0.0189) };
                \addplot[color=zero_color, error bars/.cd, y dir=both, y explicit] coordinates {(0, 0.0011) +- (0, 0.0004) (0.2, 0.0052) +- (0, 0.0018) (0.4, 0.0017) +- (0, 0.0026) (0.6, 0.0043) +- (0, 0.0037) (0.8, 0.0166) +- (0, 0.0077) (1, 0.0834) +- (0, 0.0181) };
                \addplot[color=max_color, error bars/.cd, y dir=both, y explicit] coordinates {(0, 0.0012) +- (0, 0.0000) (0.2, 0.0053) +- (0, 0.0011) (0.4, 0.0329) +- (0, 0.0026) (0.6, 0.0751) +- (0, 0.0055) (0.8, 0.2373) +- (0, 0.0131) (1, 0.1090) +- (0, 0.0221) };
                \addplot[color=ad_color, error bars/.cd, y dir=both, y explicit] coordinates {(0, 0.0035) +- (0, 0.0006) (0.2, 0.0236) +- (0, 0.0038) (0.4, 0.0420) +- (0, 0.0055) (0.6, 0.0872) +- (0, 0.0075) (0.8, 0.3087) +- (0, 0.0132) (1, 0.2244) +- (0, 0.0259) };
                \addplot[color=lma_color, error bars/.cd, y dir=both, y explicit] coordinates {(0, 0.3298) +- (0, 0.0625) (0.2, 0.2396) +- (0, 0.0377) (0.4, 0.3483) +- (0, 0.0362) (0.6, 0.4511) +- (0, 0.0400) (0.8, 0.7390) +- (0, 0.0420) (1, 0.6972) +- (0, 0.0396) };
                \addplot[color=lm_color, error bars/.cd, y dir=both, y explicit] coordinates {(0, 0.0013) +- (0, 0.0008) (0.2, 0.0070) +- (0, 0.0015) (0.4, 0.0071) +- (0, 0.0028) (0.6, 0.0105) +- (0, 0.0034) (0.8, 0.0185) +- (0, 0.0071) (1, 0.1104) +- (0, 0.0198) };
            \end{axis}
        \end{tikzpicture}
    \end{subfigure}\hspace{6mm}
    \begin{subfigure}{0.24\textwidth}
        \centering
        \begin{tikzpicture}
            \begin{axis}[grid, width=\linewidth, xlabel=Conf. Threshold, yticklabel=\empty,height=3.0cm, ymin=0, ymax=1, xmin=0, xmax=1,xtick={0, 0.25, 0.5, 0.75, 1}]
                \addplot[color=min_color, error bars/.cd, y dir=both, y explicit] coordinates {(0, 0.0032) +- (0, 0.0025) (0.2, 0.0026) +- (0, 0.0021) (0.4, 0.0037) +- (0, 0.0042) (0.6, 0.0075) +- (0, 0.0073) (0.8, 0.0276) +- (0, 0.0107) (1, 0.0866) +- (0, 0.0171) };
                \addplot[color=avg_color, error bars/.cd, y dir=both, y explicit] coordinates {(0, 0.0007) +- (0, 0.0004) (0.2, 0.0011) +- (0, 0.0016) (0.4, 0.0025) +- (0, 0.0022) (0.6, 0.0019) +- (0, 0.0037) (0.8, 0.0269) +- (0, 0.0085) (1, 0.0720) +- (0, 0.0155) };
                \addplot[color=zero_color, error bars/.cd, y dir=both, y explicit] coordinates {(0, 0.0005) +- (0, 0.0003) (0.2, 0.0012) +- (0, 0.0023) (0.4, 0.0023) +- (0, 0.0027) (0.6, 0.0018) +- (0, 0.0039) (0.8, 0.0361) +- (0, 0.0091) (1, 0.0738) +- (0, 0.0157) };
                \addplot[color=max_color, error bars/.cd, y dir=both, y explicit] coordinates {(0, 0.0012) +- (0, 0.0006) (0.2, 0.0043) +- (0, 0.0010) (0.4, 0.0047) +- (0, 0.0025) (0.6, 0.0094) +- (0, 0.0042) (0.8, 0.0478) +- (0, 0.0063) (1, 0.1018) +- (0, 0.0140) };
                \addplot[color=ad_color, error bars/.cd, y dir=both, y explicit] coordinates {(0, 0.0035) +- (0, 0.0020) (0.2, 0.0344) +- (0, 0.0044) (0.4, 0.0239) +- (0, 0.0054) (0.6, 0.0493) +- (0, 0.0086) (0.8, 0.1058) +- (0, 0.0134) (1, 0.2887) +- (0, 0.0257) };
                \addplot[color=lma_color, error bars/.cd, y dir=both, y explicit] coordinates {(0, 0.3298) +- (0, 0.0565) (0.2, 0.2420) +- (0, 0.0311) (0.4, 0.2070) +- (0, 0.0334) (0.6, 0.2882) +- (0, 0.0365) (0.8, 0.3716) +- (0, 0.0394) (1, 0.3715) +- (0, 0.0397) };
                \addplot[color=lm_color, error bars/.cd, y dir=both, y explicit] coordinates {(0, 0.0013) +- (0, 0.0028) (0.2, 0.0280) +- (0, 0.0030) (0.4, 0.0238) +- (0, 0.0050) (0.6, 0.0418) +- (0, 0.0076) (0.8, 0.0410) +- (0, 0.0101) (1, 0.3390) +- (0, 0.0183) };
            \end{axis}
        \end{tikzpicture}
    \end{subfigure}\hspace{6mm}
    \begin{subfigure}{0.24\textwidth}
        \centering
        \begin{tikzpicture}
            \begin{axis}[grid, width=\linewidth, xlabel=Conf. Threshold, yticklabel=\empty, height=3.0cm, ymin=0, ymax=1, xmin=0, xmax=1,xtick={0, 0.25, 0.5, 0.75, 1}]
                \addplot[color=min_color, error bars/.cd, y dir=both, y explicit] coordinates {(0, 0.0032) +- (0, 0.0000) (0.2, 0.0235) +- (0, 0.0050) (0.4, 0.0456) +- (0, 0.0048) (0.6, 0.0672) +- (0, 0.0057) (0.8, 0.0803) +- (0, 0.0062) (1, 0.1102) +- (0, 0.0064) };
                \addplot[color=avg_color, error bars/.cd, y dir=both, y explicit] coordinates {(0, 0.0007) +- (0, 0.0000) (0.2, 0.0310) +- (0, 0.0062) (0.4, 0.0604) +- (0, 0.0055) (0.6, 0.0827) +- (0, 0.0061) (0.8, 0.1014) +- (0, 0.0082) (1, 0.1118) +- (0, 0.0083) };
                \addplot[color=zero_color, error bars/.cd, y dir=both, y explicit] coordinates {(0, 0.0005) +- (0, 0.0000) (0.2, 0.0234) +- (0, 0.0050) (0.4, 0.0684) +- (0, 0.0061) (0.6, 0.0900) +- (0, 0.0078) (0.8, 0.1047) +- (0, 0.0083) (1, 0.1231) +- (0, 0.0106) };
                \addplot[color=max_color, error bars/.cd, y dir=both, y explicit] coordinates {(0, 0.0012) +- (0, 0.0000) (0.2, 0.0251) +- (0, 0.0054) (0.4, 0.0575) +- (0, 0.0077) (0.6, 0.0756) +- (0, 0.0081) (0.8, 0.0975) +- (0, 0.0080) (1, 0.1124) +- (0, 0.0088) };
                \addplot[color=ad_color, error bars/.cd, y dir=both, y explicit] coordinates {(0, 0.0035) +- (0, 0.0000) (0.2, 0.0209) +- (0, 0.0046) (0.4, 0.0760) +- (0, 0.0077) (0.6, 0.1137) +- (0, 0.0090) (0.8, 0.1395) +- (0, 0.0097) (1, 0.1673) +- (0, 0.0106) };
                \addplot[color=lma_color, error bars/.cd, y dir=both, y explicit] coordinates {(0, 0.2432) +- (0, 0.0000) (0.2, 0.0155) +- (0, 0.0125) (0.4, 0.0103) +- (0, 0.0083) (0.6, 0.0435) +- (0, 0.0169) (0.8, 0.0657) +- (0, 0.0208) (1, 0.1666) +- (0, 0.0278) };
                \addplot[color=lm_color, error bars/.cd, y dir=both, y explicit] coordinates {(0, 0.0032) +- (0, 0.0000) (0.2, 0.0146) +- (0, 0.0038) (0.4, 0.0369) +- (0, 0.0049) (0.6, 0.0590) +- (0, 0.0062) (0.8, 0.0820) +- (0, 0.0076) (1, 0.1193) +- (0, 0.0075) };
            \end{axis}
        \end{tikzpicture}
    \end{subfigure}

\centering
\begin{tikzpicture}
    \node[draw,color=black,rounded corners] (legend) {\quad \textcolor{ad_color}{\textbf{ConvAD}} \quad \textcolor{min_color}{\textbf{Min}} \quad \textcolor{max_color}{\textbf{Max}} \quad \textcolor{avg_color}{\textbf{Avg}} \quad \textcolor{zero_color}{\textbf{Zero}} \quad \textcolor{lma_color}{\textbf{LayerMask (Ablated)}} \quad \textcolor{lm_color}{\textbf{LayerMask}} \quad};
\end{tikzpicture}
\end{center}

\caption{
$\beta$-robustness (rows) of explanation against \iid background on ImageNet-1k, ImageNet-v2 for ResNet-50, RegNet-Y and PASCAL-VOC on EfficientNet-v2 models. Class robustness of the \convad explanation against the different masking values is more consistent and more robust. Error bars represents Standard Error of the Mean (SEM).}
\label{fig:robustness_bg_image}

\end{figure}

%% file: Plots/updated_robustness_lime_iid.tex
\begin{figure}[!htbp]
\begin{center}
    \begin{subfigure}{0.24\textwidth}
    \centering
    \begin{tikzpicture}
        \begin{axis}[grid, width=\linewidth, ylabel=ResNet50, xticklabel=\empty, height=3.0cm, ymin=0, ymax=1, xmin=0, xmax=1,xtick={0, 0.25, 0.5, 0.75, 1},title=ImageNet-1k]
            \addplot[color=min_color, error bars/.cd, y dir=both, y explicit] coordinates {(0.0, 0.0017) +- (0, 0.0008) (0.1, 0.0108) +- (0, 0.0035) (0.3, 0.0295) +- (0, 0.0054) (0.5, 0.0673) +- (0, 0.0116) (0.7, 0.1196) +- (0, 0.0183) (0.9, 0.1879) +- (0, 0.0248)};
            \addplot[color=max_color, error bars/.cd, y dir=both, y explicit] coordinates {(0.0, 0.0059) +- (0, 0.0018) (0.1, 0.0632) +- (0, 0.0122) (0.3, 0.1276) +- (0, 0.0177) (0.5, 0.2194) +- (0, 0.0262) (0.7, 0.3061) +- (0, 0.0296) (0.9, 0.3880) +- (0, 0.0331)};
            \addplot[color=avg_color, error bars/.cd, y dir=both, y explicit] coordinates {(0.0, 0.0012) +- (0, 0.0004) (0.1, 0.0029) +- (0, 0.0006) (0.3, 0.0071) +- (0, 0.0012) (0.5, 0.0215) +- (0, 0.0045) (0.7, 0.0448) +- (0, 0.0084) (0.9, 0.0801) +- (0, 0.0171)};
            \addplot[color=zero_color, error bars/.cd, y dir=both, y explicit] coordinates {(0.0, 0.0018) +- (0, 0.0004) (0.1, 0.0028) +- (0, 0.0006) (0.3, 0.0049) +- (0, 0.0010) (0.5, 0.0083) +- (0, 0.0019) (0.7, 0.0293) +- (0, 0.0100) (0.9, 0.0555) +- (0, 0.0143)};
            \addplot[color=ad_color, error bars/.cd, y dir=both, y explicit] coordinates {(0.0, 0.0061) +- (0, 0.0015) (0.1, 0.0200) +- (0, 0.0045) (0.3, 0.0447) +- (0, 0.0070) (0.5, 0.0925) +- (0, 0.0119) (0.7, 0.1448) +- (0, 0.0163) (0.9, 0.2215) +- (0, 0.0232)};
            \addplot[color=lma_color, error bars/.cd, y dir=both, y explicit] coordinates {(0.0, 0.0000) +- (0, 0.0000) (0.1, 0.0000) +- (0, 0.0000) (0.3, 0.0000) +- (0, 0.0000) (0.5, 0.0000) +- (0, 0.0000) (0.7, 0.0000) +- (0, 0.0000) (0.9, 0.0000) +- (0, 0.0000)};
            \addplot[color=lm_color, error bars/.cd, y dir=both, y explicit] coordinates {(0.0, 0.0020) +- (0, 0.0016) (0.1, 0.0000) +- (0, 0.0000) (0.3, 0.0000) +- (0, 0.0000) (0.5, 0.0000) +- (0, 0.0000) (0.7, 0.0000) +- (0, 0.0000) (0.9, 0.0000) +- (0, 0.0000)};
        \end{axis}
    \end{tikzpicture}
\end{subfigure}\hspace{6mm}
\begin{subfigure}{0.24\textwidth}
    \centering
    \begin{tikzpicture}
        \begin{axis}[grid, width=\linewidth, yticklabel=\empty, xticklabel=\empty, height=3.0cm, ymin=0, ymax=1, xmin=0, xmax=1,xtick={0, 0.25, 0.5, 0.75, 1},title=ImageNet-v2]
            \addplot[color=min_color, error bars/.cd, y dir=both, y explicit] coordinates {(0.0, 0.0009) +- (0, 0.0009) (0.1, 0.0071) +- (0, 0.0045) (0.3, 0.0102) +- (0, 0.0078) (0.5, 0.0744) +- (0, 0.0138) (0.7, 0.1128) +- (0, 0.0178) (0.9, 0.1854) +- (0, 0.0243)};
            \addplot[color=max_color, error bars/.cd, y dir=both, y explicit] coordinates {(0.0, 0.0027) +- (0, 0.0022) (0.1, 0.0128) +- (0, 0.0048) (0.3, 0.0280) +- (0, 0.0082) (0.5, 0.2361) +- (0, 0.0262) (0.7, 0.3146) +- (0, 0.0292) (0.9, 0.4129) +- (0, 0.0307)};
            \addplot[color=avg_color, error bars/.cd, y dir=both, y explicit] coordinates {(0.0, 0.0004) +- (0, 0.0003) (0.1, 0.0010) +- (0, 0.0014) (0.3, 0.0024) +- (0, 0.0027) (0.5, 0.0267) +- (0, 0.0104) (0.7, 0.0400) +- (0, 0.0110) (0.9, 0.0740) +- (0, 0.0161)};
            \addplot[color=zero_color, error bars/.cd, y dir=both, y explicit] coordinates {(0.0, 0.0006) +- (0, 0.0003) (0.1, 0.0028) +- (0, 0.0014) (0.3, 0.0034) +- (0, 0.0062) (0.5, 0.0133) +- (0, 0.0076) (0.7, 0.0212) +- (0, 0.0086) (0.9, 0.0553) +- (0, 0.0161)};
            \addplot[color=ad_color, error bars/.cd, y dir=both, y explicit] coordinates {(0.0, 0.0078) +- (0, 0.0022) (0.1, 0.0018) +- (0, 0.0041) (0.3, 0.0177) +- (0, 0.0087) (0.5, 0.0898) +- (0, 0.0125) (0.7, 0.1480) +- (0, 0.0176) (0.9, 0.2227) +- (0, 0.0238)};
            \addplot[color=lma_color, error bars/.cd, y dir=both, y explicit] coordinates {(0.0, 0.0000) +- (0, 0.0000) (0.1, 0.0000) +- (0, 0.0000) (0.3, 0.0000) +- (0, 0.0000) (0.5, 0.0000) +- (0, 0.0000) (0.7, 0.0000) +- (0, 0.0000) (0.9, 0.0000) +- (0, 0.0000)};
            \addplot[color=lm_color, error bars/.cd, y dir=both, y explicit] coordinates {(0.0, 0.0000) +- (0, 0.0000) (0.1, 0.0000) +- (0, 0.0000) (0.3, 0.0000) +- (0, 0.0000) (0.5, 0.0000) +- (0, 0.0000) (0.7, 0.0000) +- (0, 0.0000) (0.9, 0.0000) +- (0, 0.0000)};
        \end{axis}
    \end{tikzpicture}
\end{subfigure}\hspace{6mm}
\begin{subfigure}{0.24\textwidth}
    \centering
    \begin{tikzpicture}
        \begin{axis}[grid, width=\linewidth, yticklabel=\empty,xticklabel=\empty, height=3.0cm, ymin=0, ymax=1, xmin=0, xmax=1,
        xtick={0, 0.25, 0.5, 0.75, 1}, title=PASCAL-VOC]
            \addplot[color=min_color, error bars/.cd, y dir=both, y explicit] coordinates {(0.0, 0.0000) +- (0, 0.0000) (0.1, 0.0206) +- (0, 0.0042) (0.3, 0.0522) +- (0, 0.0147) (0.5, 0.0570) +- (0, 0.0173) (0.7, 0.0515) +- (0, 0.0183) (0.9, 0.2447) +- (0, 0.0200)};
            \addplot[color=max_color, error bars/.cd, y dir=both, y explicit] coordinates {(0.0, 0.0000) +- (0, 0.0000) (0.1, 0.0372) +- (0, 0.0082) (0.3, 0.0728) +- (0, 0.0106) (0.5, 0.1138) +- (0, 0.0125) (0.7, 0.1506) +- (0, 0.0131) (0.9, 0.1927) +- (0, 0.0159)};
            \addplot[color=avg_color, error bars/.cd, y dir=both, y explicit] coordinates {(0.0, 0.0000) +- (0, 0.0000) (0.1, 0.0237) +- (0, 0.0067) (0.3, 0.0603) +- (0, 0.0060) (0.5, 0.0127) +- (0, 0.0072) (0.7, 0.0890) +- (0, 0.0074) (0.9, 0.1041) +- (0, 0.0076)};
            \addplot[color=zero_color, error bars/.cd, y dir=both, y explicit] coordinates {(0.0, 0.0000) +- (0, 0.0000) (0.1, 0.0453) +- (0, 0.0075) (0.3, 0.0691) +- (0, 0.0063) (0.5, 0.0734) +- (0, 0.0065) (0.7, 0.0805) +- (0, 0.0062) (0.9, 0.1004) +- (0, 0.0072)};
            \addplot[color=ad_color, error bars/.cd, y dir=both, y explicit] coordinates {(0.0, 0.0000) +- (0, 0.0000) (0.1, 0.0738) +- (0, 0.0070) (0.3, 0.1292) +- (0, 0.0070) (0.5, 0.0453) +- (0, 0.0073) (0.7, 0.2255) +- (0, 0.0069) (0.9, 0.1283) +- (0, 0.0081)};
            \addplot[color=lma_color, error bars/.cd, y dir=both, y explicit] coordinates {(0.0, 0.0000) +- (0, 0.0000) (0.1, 0.0000) +- (0, 0.0000) (0.3, 0.0392) +- (0, 0.0007) (0.5, 0.0305) +- (0, 0.0006) (0.7, 0.1059) +- (0, 0.0006) (0.9, 0.1621) +- (0, 0.0017)};
            \addplot[color=lm_color, error bars/.cd, y dir=both, y explicit] coordinates {(0.0, 0.0000) +- (0, 0.0000) (0.1, 0.0468) +- (0, 0.0016) (0.3, 0.2773) +- (0, 0.0000) (0.5, 0.0358) +- (0, 0.0000) (0.7, 0.0616) +- (0, 0.0000) (0.9, 0.3414) +- (0, 0.0049)};
        \end{axis}
    \end{tikzpicture}
\end{subfigure}


\begin{subfigure}{0.24\textwidth}
    \centering
    \begin{tikzpicture}
        \begin{axis}[grid, width=\linewidth, ylabel=RegNetY,xticklabel=\empty,height=3.0cm, ymin=0, ymax=1, xmin=0, xmax=1,xtick={0, 0.25, 0.5, 0.75, 1}]
            \addplot[color=min_color, error bars/.cd, y dir=both, y explicit] coordinates {(0.0, 0.0014) +- (0, 0.0005) (0.1, 0.0112) +- (0, 0.0044) (0.3, 0.0302) +- (0, 0.0078) (0.5, 0.0495) +- (0, 0.0100) (0.7, 0.0869) +- (0, 0.0147) (0.9, 0.1668) +- (0, 0.0220)};
            \addplot[color=max_color, error bars/.cd, y dir=both, y explicit] coordinates {(0.0, 0.2093) +- (0, 0.0259) (0.1, 0.0632) +- (0, 0.0122) (0.3, 0.1124) +- (0, 0.0176) (0.5, 0.0715) +- (0, 0.0119) (0.7, 0.1276) +- (0, 0.0177) (0.9, 0.2194) +- (0, 0.0262)};
            \addplot[color=avg_color, error bars/.cd, y dir=both, y explicit] coordinates {(0.0, 0.0165) +- (0, 0.0025) (0.1, 0.0029) +- (0, 0.0006) (0.3, 0.0090) +- (0, 0.0019) (0.5, 0.0049) +- (0, 0.0009) (0.7, 0.0071) +- (0, 0.0012) (0.9, 0.0215) +- (0, 0.0045)};
            \addplot[color=zero_color, error bars/.cd, y dir=both, y explicit] coordinates {(0.0, 0.0089) +- (0, 0.0020) (0.1, 0.0028) +- (0, 0.0006) (0.3, 0.0048) +- (0, 0.0008) (0.5, 0.0039) +- (0, 0.0007) (0.7, 0.0049) +- (0, 0.0010) (0.9, 0.0083) +- (0, 0.0019)};
            \addplot[color=ad_color, error bars/.cd, y dir=both, y explicit] coordinates {(0.0, 0.0845) +- (0, 0.0107) (0.1, 0.0200) +- (0, 0.0045) (0.3, 0.0520) +- (0, 0.0069) (0.5, 0.0346) +- (0, 0.0055) (0.7, 0.0447) +- (0, 0.0070) (0.9, 0.0925) +- (0, 0.0119)};
            \addplot[color=lma_color, error bars/.cd, y dir=both, y explicit] coordinates {(0.0, 0.1415) +- (0, 0.0273) (0.1, 0.0000) +- (0, 0.0000) (0.3, 0.0522) +- (0, 0.0172) (0.5, 0.0347) +- (0, 0.0153) (0.7, 0.0000) +- (0, 0.0000) (0.9, 0.0000) +- (0, 0.0000)};
            \addplot[color=lm_color, error bars/.cd, y dir=both, y explicit] coordinates {(0.0, 0.0847) +- (0, 0.0119) (0.1, 0.0000) +- (0, 0.0000) (0.3, 0.0490) +- (0, 0.0067) (0.5, 0.0331) +- (0, 0.0045) (0.7, 0.0000) +- (0, 0.0000) (0.9, 0.0000) +- (0, 0.0000)};
        \end{axis}
    \end{tikzpicture}
\end{subfigure}\hspace{6mm}
\begin{subfigure}{0.24\textwidth}
    \centering
    \begin{tikzpicture}
        \begin{axis}[grid, width=\linewidth, yticklabel=\empty,xticklabel=\empty,height=3.0cm, ymin=0, ymax=1, xmin=0, xmax=1,xtick={0, 0.25, 0.5, 0.75, 1}]
            \addplot[color=min_color, error bars/.cd, y dir=both, y explicit] coordinates {(0.0, 0.0020) +- (0, 0.0007) (0.1, 0.0125) +- (0, 0.0019) (0.3, 0.0383) +- (0, 0.0021) (0.5, 0.0744) +- (0, 0.0061) (0.7, 0.1128) +- (0, 0.0098) (0.9, 0.1380) +- (0, 0.0211)};
            \addplot[color=max_color, error bars/.cd, y dir=both, y explicit] coordinates {(0.0, 0.0128) +- (0, 0.0004) (0.1, 0.0306) +- (0, 0.0038) (0.3, 0.0961) +- (0, 0.0054) (0.5, 0.1131) +- (0, 0.0100) (0.7, 0.1124) +- (0, 0.0157) (0.9, 0.2093) +- (0, 0.0237)};
            \addplot[color=avg_color, error bars/.cd, y dir=both, y explicit] coordinates {(0.0, 0.0010) +- (0, 0.0001) (0.1, 0.0040) +- (0, 0.0005) (0.3, 0.0076) +- (0, 0.0006) (0.5, 0.0267) +- (0, 0.0006) (0.7, 0.0400) +- (0, 0.0012) (0.9, 0.0144) +- (0, 0.0030)};
            \addplot[color=zero_color, error bars/.cd, y dir=both, y explicit] coordinates {(0.0, 0.0006) +- (0, 0.0002) (0.1, 0.0028) +- (0, 0.0003) (0.3, 0.0081) +- (0, 0.0003) (0.5, 0.0133) +- (0, 0.0005) (0.7, 0.0212) +- (0, 0.0006) (0.9, 0.0089) +- (0, 0.0017)};
            \addplot[color=ad_color, error bars/.cd, y dir=both, y explicit] coordinates {(0.0, 0.0018) +- (0, 0.0005) (0.1, 0.0106) +- (0, 0.0025) (0.3, 0.0934) +- (0, 0.0038) (0.5, 0.0898) +- (0, 0.0053) (0.7, 0.1480) +- (0, 0.0075) (0.9, 0.0845) +- (0, 0.0106)};
            \addplot[color=lma_color, error bars/.cd, y dir=both, y explicit] coordinates {(0.0, 0.0000) +- (0, 0.0052) (0.1, 0.0000) +- (0, 0.0033) (0.3, 0.0000) +- (0, 0.0030) (0.5, 0.0000) +- (0, 0.0033) (0.7, 0.0000) +- (0, 0.0127) (0.9, 0.1415) +- (0, 0.0239)};
            \addplot[color=lm_color, error bars/.cd, y dir=both, y explicit] coordinates {(0.0, 0.0000) +- (0, 0.0009) (0.1, 0.0000) +- (0, 0.0024) (0.3, 0.0368) +- (0, 0.0033) (0.5, 0.0000) +- (0, 0.0042) (0.7, 0.0000) +- (0, 0.0054) (0.9, 0.0847) +- (0, 0.0102)};
        \end{axis}
    \end{tikzpicture}
\end{subfigure}\hspace{6mm}
\begin{subfigure}{0.24\textwidth}
    \centering
    \begin{tikzpicture}
        \begin{axis}[grid, width=\linewidth, yticklabel=\empty,xticklabel=\empty,height=3.0cm, ymin=0, ymax=1, xmin=0, xmax=1,xtick={0, 0.25, 0.5, 0.75, 1}]
            \addplot[color=min_color, error bars/.cd, y dir=both, y explicit] coordinates {(0.0, 0.0012) +- (0, 0.0000) (0.1, 0.0162) +- (0, 0.0042) (0.3, 0.0522) +- (0, 0.0062) (0.5, 0.0570) +- (0, 0.0066) (0.7, 0.0515) +- (0, 0.0066) (0.9, 0.1006) +- (0, 0.0065)};
            \addplot[color=max_color, error bars/.cd, y dir=both, y explicit] coordinates {(0.0, 0.0012) +- (0, 0.0000) (0.1, 0.0160) +- (0, 0.0053) (0.3, 0.0439) +- (0, 0.0067) (0.5, 0.0511) +- (0, 0.0074) (0.7, 0.0728) +- (0, 0.0073) (0.9, 0.0984) +- (0, 0.0076)};
            \addplot[color=avg_color, error bars/.cd, y dir=both, y explicit] coordinates {(0.0, 0.0004) +- (0, 0.0000) (0.1, 0.0035) +- (0, 0.0046) (0.3, 0.0076) +- (0, 0.0063) (0.5, 0.0127) +- (0, 0.0065) (0.7, 0.0065) +- (0, 0.0065) (0.9, 0.0970) +- (0, 0.0068)};
            \addplot[color=zero_color, error bars/.cd, y dir=both, y explicit] coordinates {(0.0, 0.0006) +- (0, 0.0000) (0.1, 0.0032) +- (0, 0.0062) (0.3, 0.0295) +- (0, 0.0062) (0.5, 0.0453) +- (0, 0.0067) (0.7, 0.0778) +- (0, 0.0066) (0.9, 0.1117) +- (0, 0.0067)};
            \addplot[color=ad_color, error bars/.cd, y dir=both, y explicit] coordinates {(0.0, 0.0040) +- (0, 0.0000) (0.1, 0.0083) +- (0, 0.0121) (0.3, 0.1732) +- (0, 0.0129) (0.5, 0.0602) +- (0, 0.0133) (0.7, 0.0738) +- (0, 0.0135) (0.9, 0.2255) +- (0, 0.0138)};
            \addplot[color=lma_color, error bars/.cd, y dir=both, y explicit] coordinates {(0.0, 0.0000) +- (0, 0.0000) (0.1, 0.0000) +- (0, 0.0000) (0.3, 0.0076) +- (0, 0.0136) (0.5, 0.0000) +- (0, 0.0157) (0.7, 0.0000) +- (0, 0.0192) (0.9, 0.1621) +- (0, 0.0295)};
            \addplot[color=lm_color, error bars/.cd, y dir=both, y explicit] coordinates {(0.0, 0.0004) +- (0, 0.0000) (0.1, 0.0230) +- (0, 0.0124) (0.3, 0.0761) +- (0, 0.0188) (0.5, 0.0468) +- (0, 0.0191) (0.7, 0.0000) +- (0, 0.0197) (0.9, 0.3414) +- (0, 0.0195)};
        \end{axis}
    \end{tikzpicture}
\end{subfigure}


\begin{subfigure}{0.24\textwidth}
    \centering
    \begin{tikzpicture}
        \begin{axis}[grid, width=\linewidth, xlabel=Conf. threshold, ylabel=EfficientNet, height=3.0cm, ymin=0, ymax=1, xmin=0, xmax=1,xtick={0, 0.25, 0.5, 0.75, 1}]
            \addplot[color=min_color, error bars/.cd, y dir=both, y explicit] coordinates {(0.0, 0.0086) +- (0, 0.0071) (0.1, 0.0683) +- (0, 0.0208) (0.3, 0.0935) +- (0, 0.0261) (0.5, 0.1122) +- (0, 0.0268) (0.7, 0.1792) +- (0, 0.0393) (0.9, 0.2969) +- (0, 0.0497)};
            \addplot[color=max_color, error bars/.cd, y dir=both, y explicit] coordinates {(0.0, 0.0059) +- (0, 0.0022) (0.1, 0.0632) +- (0, 0.0096) (0.3, 0.0715) +- (0, 0.0197) (0.5, 0.0961) +- (0, 0.0239) (0.7, 0.1124) +- (0, 0.0282) (0.9, 0.1798) +- (0, 0.0392)};
            \addplot[color=avg_color, error bars/.cd, y dir=both, y explicit] coordinates {(0.0, 0.0012) +- (0, 0.0014) (0.1, 0.0070) +- (0, 0.0020) (0.3, 0.0143) +- (0, 0.0033) (0.5, 0.0239) +- (0, 0.0072) (0.7, 0.0310) +- (0, 0.0099) (0.9, 0.0908) +- (0, 0.0298)};
            \addplot[color=zero_color, error bars/.cd, y dir=both, y explicit] coordinates {(0.0, 0.0050) +- (0, 0.0044) (0.1, 0.0096) +- (0, 0.0027) (0.3, 0.0180) +- (0, 0.0037) (0.5, 0.0288) +- (0, 0.0075) (0.7, 0.0476) +- (0, 0.0127) (0.9, 0.1073) +- (0, 0.0312)};
            \addplot[color=ad_color, error bars/.cd, y dir=both, y explicit] coordinates {(0.0, 0.0000) +- (0, 0.0000) (0.1, 0.0061) +- (0, 0.0083) (0.3, 0.0925) +- (0, 0.0129) (0.5, 0.1643) +- (0, 0.0330) (0.7, 0.2061) +- (0, 0.0367) (0.9, 0.2957) +- (0, 0.0426)};
            \addplot[color=lma_color, error bars/.cd, y dir=both, y explicit] coordinates {(0.0, 0.0000) +- (0, 0.0000) (0.1, 0.0000) +- (0, 0.0000) (0.3, 0.0102) +- (0, 0.0129) (0.5, 0.0204) +- (0, 0.0202) (0.7, 0.0753) +- (0, 0.0365) (0.9, 0.1373) +- (0, 0.0483)};
            \addplot[color=lm_color, error bars/.cd, y dir=both, y explicit] coordinates {(0.0, 0.0000) +- (0, 0.0013) (0.1, 0.0064) +- (0, 0.0016) (0.3, 0.0153) +- (0, 0.0035) (0.5, 0.0294) +- (0, 0.0066) (0.7, 0.0380) +- (0, 0.0070) (0.9, 0.0890) +- (0, 0.0196)};
        \end{axis}
    \end{tikzpicture}
\end{subfigure}\hspace{6mm}
\begin{subfigure}{0.24\textwidth}
    \centering
    \begin{tikzpicture}
        \begin{axis}[grid, width=\linewidth, xlabel=Conf. Threshold, yticklabel=\empty,height=3.0cm, ymin=0, ymax=1, xmin=0, xmax=1,xtick={0, 0.25, 0.5, 0.75, 1}]
            \addplot[color=min_color, error bars/.cd, y dir=both, y explicit] coordinates {(0.0, 0.0000) +- (0, 0.0007) (0.1, 0.0206) +- (0, 0.0049) (0.3, 0.0933) +- (0, 0.0056) (0.5, 0.0570) +- (0, 0.0090) (0.7, 0.0949) +- (0, 0.0139) (0.9, 0.1671) +- (0, 0.0222)};
            \addplot[color=max_color, error bars/.cd, y dir=both, y explicit] coordinates {(0.0, 0.0000) +- (0, 0.0007) (0.1, 0.0250) +- (0, 0.0016) (0.3, 0.0439) +- (0, 0.0020) (0.5, 0.0967) +- (0, 0.0045) (0.7, 0.1117) +- (0, 0.0063) (0.9, 0.1205) +- (0, 0.0196)};
            \addplot[color=avg_color, error bars/.cd, y dir=both, y explicit] coordinates {(0.0, 0.0000) +- (0, 0.0003) (0.1, 0.0237) +- (0, 0.0008) (0.3, 0.0043) +- (0, 0.0013) (0.5, 0.0127) +- (0, 0.0022) (0.7, 0.0259) +- (0, 0.0059) (0.9, 0.0618) +- (0, 0.0137)};
            \addplot[color=zero_color, error bars/.cd, y dir=both, y explicit] coordinates {(0.0, 0.0000) +- (0, 0.0003) (0.1, 0.0006) +- (0, 0.0007) (0.3, 0.0453) +- (0, 0.0016) (0.5, 0.0596) +- (0, 0.0028) (0.7, 0.0295) +- (0, 0.0072) (0.9, 0.0575) +- (0, 0.0127)};
            \addplot[color=ad_color, error bars/.cd, y dir=both, y explicit] coordinates {(0.0, 0.0000) +- (0, 0.0034) (0.1, 0.0722) +- (0, 0.0095) (0.3, 0.0207) +- (0, 0.0147) (0.5, 0.0453) +- (0, 0.0172) (0.7, 0.1732) +- (0, 0.0214) (0.9, 0.2904) +- (0, 0.0291)};
            \addplot[color=lma_color, error bars/.cd, y dir=both, y explicit] coordinates {(0.0, 0.0000) +- (0, 0.0000) (0.1, 0.0000) +- (0, 0.0000) (0.3, 0.0392) +- (0, 0.0000) (0.5, 0.0305) +- (0, 0.0000) (0.7, 0.0000) +- (0, 0.0075) (0.9, 0.0746) +- (0, 0.0227)};
            \addplot[color=lm_color, error bars/.cd, y dir=both, y explicit] coordinates {(0.0, 0.0000) +- (0, 0.0003) (0.1, 0.0468) +- (0, 0.0073) (0.3, 0.0262) +- (0, 0.0082) (0.5, 0.0358) +- (0, 0.0104) (0.7, 0.0761) +- (0, 0.0138) (0.9, 0.1422) +- (0, 0.0219)};
        \end{axis}
    \end{tikzpicture}
\end{subfigure}\hspace{6mm}
\begin{subfigure}{0.24\textwidth}
    \centering
    \begin{tikzpicture}
        \begin{axis}[grid, width=\linewidth, xlabel=Conf. Threshold, yticklabel=\empty, height=3.0cm, ymin=0, ymax=1, xmin=0, xmax=1,xtick={0, 0.25, 0.5, 0.75, 1}]
            \addplot[color=min_color, error bars/.cd, y dir=both, y explicit] coordinates {(0.0, 0.0000) +- (0, 0.0000) (0.1, 0.0080) +- (0, 0.0072) (0.3, 0.0305) +- (0, 0.0056) (0.5, 0.0578) +- (0, 0.0207) (0.7, 0.1489) +- (0, 0.0320) (0.9, 0.2447) +- (0, 0.0200)};
            \addplot[color=max_color, error bars/.cd, y dir=both, y explicit] coordinates {(0.0, 0.0000) +- (0, 0.0000) (0.1, 0.0000) +- (0, 0.0143) (0.3, 0.0078) +- (0, 0.0073) (0.5, 0.0600) +- (0, 0.0248) (0.7, 0.1200) +- (0, 0.0307) (0.9, 0.1927) +- (0, 0.0159)};
            \addplot[color=avg_color, error bars/.cd, y dir=both, y explicit] coordinates {(0.0, 0.0000) +- (0, 0.0000) (0.1, 0.0000) +- (0, 0.0354) (0.3, 0.0072) +- (0, 0.0013) (0.5, 0.0844) +- (0, 0.0231) (0.7, 0.1044) +- (0, 0.0213) (0.9, 0.1041) +- (0, 0.0076)};
            \addplot[color=zero_color, error bars/.cd, y dir=both, y explicit] coordinates {(0.0, 0.0000) +- (0, 0.0000) (0.1, 0.0000) +- (0, 0.0333) (0.3, 0.0081) +- (0, 0.0016) (0.5, 0.0489) +- (0, 0.0205) (0.7, 0.1011) +- (0, 0.0207) (0.9, 0.1004) +- (0, 0.0072)};
            \addplot[color=ad_color, error bars/.cd, y dir=both, y explicit] coordinates {(0.0, 0.0000) +- (0, 0.0000) (0.1, 0.0000) +- (0, 0.0089) (0.3, 0.0144) +- (0, 0.0093) (0.5, 0.0578) +- (0, 0.0188) (0.7, 0.2333) +- (0, 0.0644) (0.9, 0.1283) +- (0, 0.0081)};
            \addplot[color=lma_color, error bars/.cd, y dir=both, y explicit] coordinates {(0.0, 0.0000) +- (0, 0.0000) (0.1, 0.0000) +- (0, 0.0000) (0.3, 0.0000) +- (0, 0.0000) (0.5, 0.0000) +- (0, 0.0000) (0.7, 0.0000) +- (0, 0.0000) (0.9, 0.1621) +- (0, 0.0295)};
            \addplot[color=lm_color, error bars/.cd, y dir=both, y explicit] coordinates {(0.0, 0.0000) +- (0, 0.0000) (0.1, 0.0000) +- (0, 0.0089) (0.3, 0.0208) +- (0, 0.0049) (0.5, 0.0622) +- (0, 0.0262) (0.7, 0.0989) +- (0, 0.0259) (0.9, 0.3414) +- (0, 0.0195)};
        \end{axis}
    \end{tikzpicture}
\end{subfigure}

\centering
\begin{tikzpicture}
    \node[draw,color=black,rounded corners] (legend) {\quad \textcolor{ad_color}{\textbf{ConvAD}} \quad \textcolor{min_color}{\textbf{Min}} \quad \textcolor{max_color}{\textbf{Max}} \quad \textcolor{avg_color}{\textbf{Avg}} \quad \textcolor{zero_color}{\textbf{Zero}} \quad \textcolor{lma_color}{\textbf{LayerMask (Ablated)}} \quad \textcolor{lm_color}{\textbf{LayerMask}} \quad};
\end{tikzpicture}
\end{center}

\caption{$\beta$-robustness (rows) of explanation against \iid background on ImageNet-1k, ImageNet-v2 and PASCAL-VOC for ResNet-50, RegNet-Y and EfficientNet-v2 models calculated via LIME. Error bars represents Standard Error of the Mean (SEM).}
\label{fig:robustness_bg_image_lime}
\end{figure}

%% file: Plots/updated_robustness_gradcam_iid.tex
\begin{figure}[htbp]
\begin{center}
    \begin{subfigure}{0.24\textwidth}
        \centering
        \begin{tikzpicture}
            \begin{axis}[grid, width=\linewidth, ylabel=ResNet50, xticklabel=\empty, height=3.0cm, ymin=0, ymax=1, xmin=0, xmax=1,xtick={0, 0.25, 0.5, 0.75, 1},title=ImageNet-1k]
                \addplot[color=min_color, error bars/.cd, y dir=both, y explicit] 
                    coordinates {(0, 0.0020) +- (0, 0.0004) (0.2, 0.0079) +- (0, 0.0017) (0.4, 0.0524) +- (0, 0.0067) (0.6, 0.1547) +- (0, 0.0187) (0.8, 0.2696) +- (0, 0.0286) (1, 0.4778) +- (0, 0.0344)};
                \addplot[color=avg_color, error bars/.cd, y dir=both, y explicit]
                    coordinates {(0, 0.0020) +- (0, 0.0004) (0.2, 0.0059) +- (0, 0.0013) (0.4, 0.0202) +- (0, 0.0052) (0.6, 0.0621) +- (0, 0.0133) (0.8, 0.1528) +- (0, 0.0232) (1, 0.3029) +- (0, 0.0332)};
                \addplot[color=zero_color, error bars/.cd, y dir=both, y explicit]
                    coordinates {(0, 0.0020) +- (0, 0.0004) (0.2, 0.0027) +- (0, 0.0006) (0.4, 0.0104) +- (0, 0.0040) (0.6, 0.0351) +- (0, 0.0097) (0.8, 0.0878) +- (0, 0.0182) (1, 0.2239) +- (0, 0.0298)};
                \addplot[color=max_color, error bars/.cd, y dir=both, y explicit]
                    coordinates {(0, 0.0035) +- (0, 0.0006) (0.2, 0.0140) +- (0, 0.0026) (0.4, 0.0762) +- (0, 0.0086) (0.6, 0.1856) +- (0, 0.0186) (0.8, 0.3471) +- (0, 0.0302) (1, 0.5281) +- (0, 0.0343)};
                \addplot[color=ad_color, error bars/.cd, y dir=both, y explicit]
                    coordinates {(0, 0.0021) +- (0, 0.0006) (0.2, 0.0254) +- (0, 0.0054) (0.4, 0.1021) +- (0, 0.0122) (0.6, 0.2283) +- (0, 0.0200) (0.8, 0.3665) +- (0, 0.0272) (1, 0.5430) +- (0, 0.0330)};
                \addplot[color=lma_color, error bars/.cd, y dir=both, y explicit]
                    coordinates {(0, 0.0000) +- (0, 0.0000) (0.2, 0.0000) +- (0, 0.0000) (0.4, 0.0000) +- (0, 0.0000) (0.6, 0.0000) +- (0, 0.0000) (0.8, 0.0000) +- (0, 0.0000) (1, 0.0000) +- (0, 0.0000)};
                \addplot[color=lm_color, error bars/.cd, y dir=both, y explicit]
                    coordinates {(0, 0.0000) +- (0, 0.0000) (0.2, 0.0000) +- (0, 0.0000) (0.4, 0.0000) +- (0, 0.0000) (0.6, 0.0000) +- (0, 0.0000) (0.8, 0.0000) +- (0, 0.0000) (1, 0.0000) +- (0, 0.0000)};
            \end{axis}
        \end{tikzpicture}
    \end{subfigure}\hspace{6mm}
    \begin{subfigure}{0.24\textwidth}
        \centering
        \begin{tikzpicture}
            \begin{axis}[grid, width=\linewidth, yticklabel=\empty, xticklabel=\empty, height=3.0cm, ymin=0, ymax=1, xmin=0, xmax=1,xtick={0, 0.25, 0.5, 0.75, 1},title=ImageNet-v2]
                \addplot[color=min_color, error bars/.cd, y dir=both, y explicit] 
                    coordinates {(0, 0.0000) +- (0, 0.0000) (0.2, 0.0084) +- (0, 0.0008) (0.4, 0.0672) +- (0, 0.0047) (0.6, 0.1375) +- (0, 0.0202) (0.8, 0.3002) +- (0, 0.0408) (1, 0.4398) +- (0, 0.0428)};
                \addplot[color=avg_color, error bars/.cd, y dir=both, y explicit]
                    coordinates {(0, 0.0000) +- (0, 0.0000) (0.2, 0.0071) +- (0, 0.0002) (0.4, 0.0547) +- (0, 0.0063) (0.6, 0.0606) +- (0, 0.0151) (0.8, 0.1494) +- (0, 0.0319) (1, 0.2783) +- (0, 0.0386)};
                \addplot[color=zero_color, error bars/.cd, y dir=both, y explicit]
                    coordinates {(0, 0.0000) +- (0, 0.0000) (0.2, 0.0044) +- (0, 0.0021) (0.4, 0.0553) +- (0, 0.0030) (0.6, 0.0240) +- (0, 0.0100) (0.8, 0.0782) +- (0, 0.0240) (1, 0.1787) +- (0, 0.0317)};
                \addplot[color=max_color, error bars/.cd, y dir=both, y explicit]
                    coordinates {(0, 0.0000) +- (0, 0.0001) (0.2, 0.0053) +- (0, 0.0019) (0.4, 0.0668) +- (0, 0.0060) (0.6, 0.1832) +- (0, 0.0320) (0.8, 0.3353) +- (0, 0.0470) (1, 0.5057) +- (0, 0.0467)};
                \addplot[color=ad_color, error bars/.cd, y dir=both, y explicit]
                    coordinates {(0, 0.0000) +- (0, 0.0000) (0.2, 0.0260) +- (0, 0.0025) (0.4, 0.0897) +- (0, 0.0159) (0.6, 0.2366) +- (0, 0.0334) (0.8, 0.3823) +- (0, 0.0465) (1, 0.5598) +- (0, 0.0440)};
                \addplot[color=lma_color, error bars/.cd, y dir=both, y explicit]
                    coordinates {(0, 0.0000) +- (0, 0.0000) (0.2, 0.0000) +- (0, 0.0000) (0.4, 0.0000) +- (0, 0.0000) (0.6, 0.0000) +- (0, 0.0000) (0.8, 0.0000) +- (0, 0.0000) (1, 0.0000) +- (0, 0.0000)};
                \addplot[color=lm_color, error bars/.cd, y dir=both, y explicit]
                    coordinates {(0, 0.0000) +- (0, 0.0000) (0.2, 0.0000) +- (0, 0.0000) (0.4, 0.0000) +- (0, 0.0000) (0.6, 0.0000) +- (0, 0.0000) (0.8, 0.0000) +- (0, 0.0000) (1, 0.0000) +- (0, 0.0000)};
            \end{axis}
        \end{tikzpicture}
    \end{subfigure}\hspace{6mm}
    \begin{subfigure}{0.24\textwidth}
        \centering
        \begin{tikzpicture}
            \begin{axis}[grid, width=\linewidth, yticklabel=\empty,xticklabel=\empty, height=3.0cm, ymin=0, ymax=1, xmin=0, xmax=1,
            xtick={0, 0.25, 0.5, 0.75, 1}, title=PASCAL-VOC]
                \addplot[color=min_color, error bars/.cd, y dir=both, y explicit] 
                    coordinates {(0, 0.0000) +- (0, 0.0000) (0.2, 0.0084) +- (0, 0.0040) (0.4, 0.0462) +- (0, 0.0095) (0.6, 0.1267) +- (0, 0.0140) (0.8, 0.1100) +- (0, 0.0162) (1, 0.2142) +- (0, 0.0180)};
                \addplot[color=avg_color, error bars/.cd, y dir=both, y explicit]
                    coordinates {(0, 0.0000) +- (0, 0.0000) (0.2, 0.0071) +- (0, 0.0066) (0.4, 0.0547) +- (0, 0.0063) (0.6, 0.0000) +- (0, 0.0063) (0.8, 0.1267) +- (0, 0.0067) (1, 0.1098) +- (0, 0.0082)};
                \addplot[color=zero_color, error bars/.cd, y dir=both, y explicit]
                    coordinates {(0, 0.0000) +- (0, 0.0000) (0.2, 0.0044) +- (0, 0.0072) (0.4, 0.0553) +- (0, 0.0058) (0.6, 0.1233) +- (0, 0.0056) (0.8, 0.1233) +- (0, 0.0061) (1, 0.1052) +- (0, 0.0082)};
                \addplot[color=max_color, error bars/.cd, y dir=both, y explicit]
                    coordinates {(0, 0.0000) +- (0, 0.0000) (0.2, 0.0053) +- (0, 0.0069) (0.4, 0.0668) +- (0, 0.0062) (0.6, 0.1333) +- (0, 0.0060) (0.8, 0.1067) +- (0, 0.0068) (1, 0.1216) +- (0, 0.0103)};
                \addplot[color=ad_color, error bars/.cd, y dir=both, y explicit]
                    coordinates {(0, 0.0000) +- (0, 0.0000) (0.2, 0.0260) +- (0, 0.0077) (0.4, 0.0897) +- (0, 0.0061) (0.6, 0.7000) +- (0, 0.0058) (0.8, 0.2100) +- (0, 0.0078) (1, 0.1520) +- (0, 0.0103)};
                \addplot[color=lma_color, error bars/.cd, y dir=both, y explicit]
                    coordinates {(0, 0.0000) +- (0, 0.0000) (0.2, 0.0000) +- (0, 0.0000) (0.4, 0.0000) +- (0, 0.0000) (0.6, 0.0000) +- (0, 0.0000) (0.8, 0.0000) +- (0, 0.0000) (1, 0.0000) +- (0, 0.0000)};
                \addplot[color=lm_color, error bars/.cd, y dir=both, y explicit]
                    coordinates {(0, 0.0000) +- (0, 0.0000) (0.2, 0.0000) +- (0, 0.0000) (0.4, 0.0000) +- (0, 0.0000) (0.6, 0.0000) +- (0, 0.0000) (0.8, 0.0000) +- (0, 0.0000) (1, 0.0000) +- (0, 0.0000)};
            \end{axis}
        \end{tikzpicture}
    \end{subfigure}

    \begin{subfigure}{0.24\textwidth}
        \centering
        \begin{tikzpicture}
            \begin{axis}[grid,  width=\linewidth, ylabel=RegNetY,xticklabel=\empty,height=3.0cm, ymin=0, ymax=1, xmin=0, xmax=1,xtick={0, 0.25, 0.5, 0.75, 1}]
                \addplot[color=min_color, error bars/.cd, y dir=both, y explicit] 
                    coordinates {(0, 0.0006) +- (0, 0.0003) (0.2, 0.0027) +- (0, 0.0008) (0.4, 0.0041) +- (0, 0.0009) (0.6, 0.0069) +- (0, 0.0015) (0.8, 0.0190) +- (0, 0.0034) (1, 0.0655) +- (0, 0.0122)};
                \addplot[color=avg_color, error bars/.cd, y dir=both, y explicit]
                    coordinates {(0, 0.0021) +- (0, 0.0004) (0.2, 0.0029) +- (0, 0.0005) (0.4, 0.0036) +- (0, 0.0007) (0.6, 0.0069) +- (0, 0.0014) (0.8, 0.0122) +- (0, 0.0022) (1, 0.0480) +- (0, 0.0103)};
                \addplot[color=zero_color, error bars/.cd, y dir=both, y explicit]
                    coordinates {(0, 0.0017) +- (0, 0.0003) (0.2, 0.0029) +- (0, 0.0006) (0.4, 0.0036) +- (0, 0.0006) (0.6, 0.0067) +- (0, 0.0014) (0.8, 0.0107) +- (0, 0.0022) (1, 0.0587) +- (0, 0.0143)};
                \addplot[color=max_color, error bars/.cd, y dir=both, y explicit]
                    coordinates {(0, 0.0021) +- (0, 0.0003) (0.2, 0.0041) +- (0, 0.0008) (0.4, 0.0071) +- (0, 0.0020) (0.6, 0.0106) +- (0, 0.0024) (0.8, 0.0222) +- (0, 0.0043) (1, 0.0864) +- (0, 0.0138)};
                \addplot[color=ad_color, error bars/.cd, y dir=both, y explicit]
                    coordinates {(0, 0.0017) +- (0, 0.0030) (0.2, 0.0604) +- (0, 0.0079) (0.4, 0.1057) +- (0, 0.0115) (0.6, 0.1366) +- (0, 0.0127) (0.8, 0.1871) +- (0, 0.0160) (1, 0.2767) +- (0, 0.0211)};
                \addplot[color=lma_color, error bars/.cd, y dir=both, y explicit]
                    coordinates {(0, 0.0000) +- (0, 0.0000) (0.2, 0.0000) +- (0, 0.0000) (0.4, 0.0000) +- (0, 0.0000) (0.6, 0.0000) +- (0, 0.0000) (0.8, 0.0000) +- (0, 0.0000) (1, 0.4620) +- (0, 0.0340)};
                \addplot[color=lm_color, error bars/.cd, y dir=both, y explicit]
                    coordinates {(0, 0.0000) +- (0, 0.0000) (0.2, 0.0000) +- (0, 0.0000) (0.4, 0.0000) +- (0, 0.0000) (0.6, 0.0000) +- (0, 0.0000) (0.8, 0.0000) +- (0, 0.0000) (1, 0.2997) +- (0, 0.0199)};
            \end{axis}
        \end{tikzpicture}
    \end{subfigure}\hspace{6mm}
    \begin{subfigure}{0.24\textwidth}
        \centering
        \begin{tikzpicture}
            \begin{axis}[grid, width=\linewidth, yticklabel=\empty,xticklabel=\empty,height=3.0cm, ymin=0, ymax=1, xmin=0, xmax=1,xtick={0, 0.25, 0.5, 0.75, 1}]
                \addplot[color=min_color, error bars/.cd, y dir=both, y explicit] 
                    coordinates {(0, 0.0000) +- (0, 0.0005) (0.2, 0.0113) +- (0, 0.0008) (0.4, 0.0206) +- (0, 0.0014) (0.6, 0.0230) +- (0, 0.0146) (0.8, 0.0548) +- (0, 0.0186) (1, 0.1450) +- (0, 0.0457)};
                \addplot[color=avg_color, error bars/.cd, y dir=both, y explicit]
                    coordinates {(0, 0.0000) +- (0, 0.0000) (0.2, 0.0339) +- (0, 0.0011) (0.4, 0.0111) +- (0, 0.0017) (0.6, 0.0260) +- (0, 0.0052) (0.8, 0.0565) +- (0, 0.0074) (1, 0.1550) +- (0, 0.0088)};
                \addplot[color=zero_color, error bars/.cd, y dir=both, y explicit]
                    coordinates {(0, 0.0000) +- (0, 0.0005) (0.2, 0.0453) +- (0, 0.0012) (0.4, 0.0118) +- (0, 0.0011) (0.6, 0.0298) +- (0, 0.0056) (0.8, 0.0439) +- (0, 0.0073) (1, 0.1500) +- (0, 0.0073)};
                \addplot[color=max_color, error bars/.cd, y dir=both, y explicit]
                    coordinates {(0, 0.0000) +- (0, 0.0012) (0.2, 0.0374) +- (0, 0.0007) (0.4, 0.0111) +- (0, 0.0006) (0.6, 0.0373) +- (0, 0.0041) (0.8, 0.0600) +- (0, 0.0083) (1, 0.1600) +- (0, 0.0289)};
                \addplot[color=ad_color, error bars/.cd, y dir=both, y explicit]
                    coordinates {(0, 0.0000) +- (0, 0.0079) (0.2, 0.0564) +- (0, 0.0084) (0.4, 0.0533) +- (0, 0.0158) (0.6, 0.0876) +- (0, 0.0158) (0.8, 0.1247) +- (0, 0.0403) (1, 0.6850) +- (0, 0.0497)};
                \addplot[color=lma_color, error bars/.cd, y dir=both, y explicit]
                    coordinates {(0, 0.0000) +- (0, 0.0000) (0.2, 0.0000) +- (0, 0.0000) (0.4, 0.0000) +- (0, 0.0000) (0.6, 0.0000) +- (0, 0.0000) (0.8, 0.0000) +- (0, 0.0000) (1, 0.3019) +- (0, 0.0312)};
                \addplot[color=lm_color, error bars/.cd, y dir=both, y explicit]
                    coordinates {(0, 0.0000) +- (0, 0.0000) (0.2, 0.0000) +- (0, 0.0000) (0.4, 0.0000) +- (0, 0.0000) (0.6, 0.0000) +- (0, 0.0000) (0.8, 0.0000) +- (0, 0.0000) (1, 0.2005) +- (0, 0.0162)};
            \end{axis}
        \end{tikzpicture}
    \end{subfigure}\hspace{6mm}
    \begin{subfigure}{0.24\textwidth}
        \centering
        \begin{tikzpicture}
            \begin{axis}[grid, width=\linewidth, yticklabel=\empty,xticklabel=\empty,height=3.0cm, ymin=0, ymax=1, xmin=0, xmax=1,xtick={0, 0.25, 0.5, 0.75, 1}]
                \addplot[color=min_color, error bars/.cd, y dir=both, y explicit] 
                    coordinates {(0, 0.0003) +- (0, 0.0000) (0.2, 0.0008) +- (0, 0.0000) (0.4, 0.0022) +- (0, 0.0019) (0.6, 0.0086) +- (0, 0.0046) (0.8, 0.0159) +- (0, 0.0057) (1, 0.0754) +- (0, 0.0053)};
                \addplot[color=avg_color, error bars/.cd, y dir=both, y explicit]
                    coordinates {(0, 0.0004) +- (0, 0.0000) (0.2, 0.0012) +- (0, 0.0016) (0.4, 0.0021) +- (0, 0.0019) (0.6, 0.0049) +- (0, 0.0045) (0.8, 0.0146) +- (0, 0.0046) (1, 0.0578) +- (0, 0.0060)};
                \addplot[color=zero_color, error bars/.cd, y dir=both, y explicit]
                    coordinates {(0, 0.0005) +- (0, 0.0000) (0.2, 0.0010) +- (0, 0.0016) (0.4, 0.0028) +- (0, 0.0016) (0.6, 0.0044) +- (0, 0.0042) (0.8, 0.0178) +- (0, 0.0040) (1, 0.0621) +- (0, 0.0057)};
                \addplot[color=max_color, error bars/.cd, y dir=both, y explicit]
                    coordinates {(0, 0.0010) +- (0, 0.0000) (0.2, 0.0021) +- (0, 0.0019) (0.4, 0.0049) +- (0, 0.0058) (0.6, 0.0172) +- (0, 0.0065) (0.8, 0.0277) +- (0, 0.0044) (1, 0.1186) +- (0, 0.0064)};
                \addplot[color=ad_color, error bars/.cd, y dir=both, y explicit]
                    coordinates {(0, 0.0015) +- (0, 0.0000) (0.2, 0.0017) +- (0, 0.0171) (0.4, 0.0021) +- (0, 0.0171) (0.6, 0.0552) +- (0, 0.0065) (0.8, 0.1413) +- (0, 0.0045) (1, 0.2953) +- (0, 0.0032)};
                \addplot[color=lma_color, error bars/.cd, y dir=both, y explicit]
                    coordinates {(0, 0.0000) +- (0, 0.0000) (0.2, 0.0000) +- (0, 0.0000) (0.4, 0.0000) +- (0, 0.0000) (0.6, 0.0000) +- (0, 0.0000) (0.8, 0.0000) +- (0, 0.0000) (1, 0.1050) +- (0, 0.0742)};
                \addplot[color=lm_color, error bars/.cd, y dir=both, y explicit]
                    coordinates {(0, 0.0000) +- (0, 0.0000) (0.2, 0.0000) +- (0, 0.0000) (0.4, 0.0000) +- (0, 0.0000) (0.6, 0.0000) +- (0, 0.0000) (0.8, 0.0000) +- (0, 0.0000) (1, 0.7833) +- (0, 0.0136)};
            \end{axis}
        \end{tikzpicture}
    \end{subfigure}

    \begin{subfigure}{0.24\textwidth}
        \centering
        \begin{tikzpicture}
            \begin{axis}[grid, width=\linewidth, xlabel=Conf. threshold, ylabel=EfficientNet, height=3.0cm, ymin=0, ymax=1, xmin=0, xmax=1,xtick={0, 0.25, 0.5, 0.75, 1}]
                \addplot[color=min_color, error bars/.cd, y dir=both, y explicit] 
                    coordinates {(0, 0.0003) +- (0, 0.0004) (0.2, 0.0044) +- (0, 0.0011) (0.4, 0.0097) +- (0, 0.0015) (0.6, 0.0230) +- (0, 0.0045) (0.8, 0.0548) +- (0, 0.0113) (1, 0.1184) +- (0, 0.0190)};
                \addplot[color=avg_color, error bars/.cd, y dir=both, y explicit]
                    coordinates {(0, 0.0004) +- (0, 0.0008) (0.2, 0.0045) +- (0, 0.0008) (0.4, 0.0160) +- (0, 0.0035) (0.6, 0.0260) +- (0, 0.0045) (0.8, 0.0565) +- (0, 0.0133) (1, 0.0885) +- (0, 0.0169)};
                \addplot[color=zero_color, error bars/.cd, y dir=both, y explicit]
                    coordinates {(0, 0.0005) +- (0, 0.0008) (0.2, 0.0047) +- (0, 0.0008) (0.4, 0.0161) +- (0, 0.0037) (0.6, 0.0298) +- (0, 0.0072) (0.8, 0.0439) +- (0, 0.0090) (1, 0.1057) +- (0, 0.0185)};
                \addplot[color=max_color, error bars/.cd, y dir=both, y explicit]
                    coordinates {(0, 0.0010) +- (0, 0.0007) (0.2, 0.0056) +- (0, 0.0016) (0.4, 0.0136) +- (0, 0.0031) (0.6, 0.0373) +- (0, 0.0090) (0.8, 0.0600) +- (0, 0.0128) (1, 0.0887) +- (0, 0.0166)};
                \addplot[color=ad_color, error bars/.cd, y dir=both, y explicit]
                    coordinates {(0, 0.0015) +- (0, 0.0015) (0.2, 0.0256) +- (0, 0.0044) (0.4, 0.0528) +- (0, 0.0097) (0.6, 0.0876) +- (0, 0.0129) (0.8, 0.1247) +- (0, 0.0164) (1, 0.2385) +- (0, 0.0260)};
                \addplot[color=lma_color, error bars/.cd, y dir=both, y explicit]
                    coordinates {(0, 0.0000) +- (0, 0.0000) (0.2, 0.0000) +- (0, 0.0000) (0.4, 0.0000) +- (0, 0.0000) (0.6, 0.0000) +- (0, 0.0000) (0.8, 0.0000) +- (0, 0.0000) (1, 0.1431) +- (0, 0.0304)};
                \addplot[color=lm_color, error bars/.cd, y dir=both, y explicit]
                    coordinates {(0, 0.0000) +- (0, 0.0000) (0.2, 0.0000) +- (0, 0.0000) (0.4, 0.0000) +- (0, 0.0000) (0.6, 0.0000) +- (0, 0.0000) (0.8, 0.0000) +- (0, 0.0000) (1, 0.1502) +- (0, 0.0208)};
            \end{axis}
        \end{tikzpicture}
    \end{subfigure}\hspace{6mm}
    \begin{subfigure}{0.24\textwidth}
        \centering
        \begin{tikzpicture}
            \begin{axis}[grid, width=\linewidth, xlabel=Conf. Threshold, yticklabel=\empty,height=3.0cm, ymin=0, ymax=1, xmin=0, xmax=1,xtick={0, 0.25, 0.5, 0.75, 1}]
                \addplot[color=min_color, error bars/.cd, y dir=both, y explicit] 
                    coordinates {(0, 0.0000) +- (0, 0.0000) (0.2, 0.0097) +- (0, 0.0638) (0.4, 0.0233) +- (0, 0.0394) (0.6, 0.1285) +- (0, 0.0365) (0.8, 0.1703) +- (0, 0.0387) (1, 0.1384) +- (0, 0.0423)};
                \addplot[color=avg_color, error bars/.cd, y dir=both, y explicit]
                    coordinates {(0, 0.0000) +- (0, 0.0000) (0.2, 0.0160) +- (0, 0.0007) (0.4, 0.0233) +- (0, 0.0136) (0.6, 0.0766) +- (0, 0.0342) (0.8, 0.0839) +- (0, 0.0371) (1, 0.1241) +- (0, 0.0408)};
                \addplot[color=zero_color, error bars/.cd, y dir=both, y explicit]
                    coordinates {(0, 0.0000) +- (0, 0.0000) (0.2, 0.0161) +- (0, 0.0009) (0.4, 0.0200) +- (0, 0.0156) (0.6, 0.0794) +- (0, 0.0263) (0.8, 0.0856) +- (0, 0.0355) (1, 0.1199) +- (0, 0.0389)};
                \addplot[color=max_color, error bars/.cd, y dir=both, y explicit]
                    coordinates {(0, 0.0000) +- (0, 0.0000) (0.2, 0.0136) +- (0, 0.0015) (0.4, 0.1333) +- (0, 0.0135) (0.6, 0.0510) +- (0, 0.0159) (0.8, 0.0915) +- (0, 0.0299) (1, 0.1395) +- (0, 0.0528)};
                \addplot[color=ad_color, error bars/.cd, y dir=both, y explicit]
                    coordinates {(0, 0.0000) +- (0, 0.0000) (0.2, 0.0256) +- (0, 0.0150) (0.4, 0.0528) +- (0, 0.0097) (0.6, 0.0710) +- (0, 0.0136) (0.8, 0.1214) +- (0, 0.0287) (1, 0.2072) +- (0, 0.0528)};
                \addplot[color=lma_color, error bars/.cd, y dir=both, y explicit]
                    coordinates {(0, 0.0000) +- (0, 0.0000) (0.2, 0.0000) +- (0, 0.0000) (0.4, 0.0000) +- (0, 0.0000) (0.6, 0.0000) +- (0, 0.0000) (0.8, 0.0000) +- (0, 0.0000) (1, 0.0000) +- (0, 0.0000)};
                \addplot[color=lm_color, error bars/.cd, y dir=both, y explicit]
                    coordinates {(0, 0.0000) +- (0, 0.0000) (0.2, 0.0000) +- (0, 0.0000) (0.4, 0.0000) +- (0, 0.0000) (0.6, 0.0000) +- (0, 0.0000) (0.8, 0.0000) +- (0, 0.0000) (1, 0.0000) +- (0, 0.0000)};
            \end{axis}
        \end{tikzpicture}
    \end{subfigure}\hspace{6mm}
    \begin{subfigure}{0.24\textwidth}
        \centering
        \begin{tikzpicture}
            \begin{axis}[grid, width=\linewidth, xlabel=Conf. Threshold, yticklabel=\empty, height=3.0cm, ymin=0, ymax=1, xmin=0, xmax=1,xtick={0, 0.25, 0.5, 0.75, 1}]
                \addplot[color=min_color, error bars/.cd, y dir=both, y explicit] 
                    coordinates {(0, 0.0006) +- (0, 0.0000) (0.2, 0.0027) +- (0, 0.0042) (0.4, 0.0041) +- (0, 0.0065) (0.6, 0.0290) +- (0, 0.0074) (0.8, 0.1088) +- (0, 0.0068) (1, 0.1292) +- (0, 0.0079)};
                \addplot[color=avg_color, error bars/.cd, y dir=both, y explicit]
                    coordinates {(0, 0.0021) +- (0, 0.0000) (0.2, 0.0029) +- (0, 0.0077) (0.4, 0.0036) +- (0, 0.0068) (0.6, 0.0177) +- (0, 0.0068) (0.8, 0.1152) +- (0, 0.0073) (1, 0.0968) +- (0, 0.0073)};
                \addplot[color=zero_color, error bars/.cd, y dir=both, y explicit]
                    coordinates {(0, 0.0017) +- (0, 0.0000) (0.2, 0.0029) +- (0, 0.0073) (0.4, 0.0036) +- (0, 0.0067) (0.6, 0.0427) +- (0, 0.0073) (0.8, 0.0972) +- (0, 0.0073) (1, 0.1047) +- (0, 0.0073)};
                \addplot[color=max_color, error bars/.cd, y dir=both, y explicit]
                    coordinates {(0, 0.0021) +- (0, 0.0000) (0.2, 0.0041) +- (0, 0.0057) (0.4, 0.0071) +- (0, 0.0062) (0.6, 0.0439) +- (0, 0.0077) (0.8, 0.0784) +- (0, 0.0070) (1, 0.1025) +- (0, 0.0101)};
                \addplot[color=ad_color, error bars/.cd, y dir=both, y explicit]
                    coordinates {(0, 0.0017) +- (0, 0.0000) (0.2, 0.0126) +- (0, 0.0042) (0.4, 0.1057) +- (0, 0.0068) (0.6, 0.1314) +- (0, 0.0120) (0.8, 0.1256) +- (0, 0.0122) (1, 0.2261) +- (0, 0.0134)};
                \addplot[color=lma_color, error bars/.cd, y dir=both, y explicit]
                    coordinates {(0, 0.0000) +- (0, 0.0000) (0.2, 0.0000) +- (0, 0.0000) (0.4, 0.0000) +- (0, 0.0000) (0.6, 0.0000) +- (0, 0.0000) (0.8, 0.0000) +- (0, 0.0000) (1, 0.0000) +- (0, 0.0000)};
                \addplot[color=lm_color, error bars/.cd, y dir=both, y explicit]
                    coordinates {(0, 0.0000) +- (0, 0.0000) (0.2, 0.0000) +- (0, 0.0000) (0.4, 0.0000) +- (0, 0.0000) (0.6, 0.0000) +- (0, 0.0000) (0.8, 0.0000) +- (0, 0.0000) (1, 0.0000) +- (0, 0.0000)};
            \end{axis}
        \end{tikzpicture}
    \end{subfigure}

\centering
\begin{tikzpicture}
    \node[draw,color=black,rounded corners] (legend) {\quad \textcolor{ad_color}{\textbf{ConvAD}} \quad \textcolor{min_color}{\textbf{Min}} \quad \textcolor{max_color}{\textbf{Max}} \quad \textcolor{avg_color}{\textbf{Avg}} \quad \textcolor{zero_color}{\textbf{Zero}} \quad \textcolor{lma_color}{\textbf{LayerMask (Ablated)}} \quad \textcolor{lm_color}{\textbf{LayerMask}} \quad};
\end{tikzpicture}
\end{center}

\caption{$\beta$-robustness (rows) of explanation against \iid background on ImageNet-1k, ImageNet-v2 and PASCAL-VOC for ResNet-50, RegNet-Y and EfficientNet-v2 models calculated via \gradcam. Error bars represents Standard Error of the Mean (SEM).}
\label{fig:robustness_bg_image_gradcam}

\end{figure}

%% file: Plots/discoverability_rex.tex
\pgfplotsset{
  cfbase/.style={
    ybar=0pt, 
    bar width=2pt,    
    x=0.8cm,          
    height=4.2cm,     
    ymin=0, ymax=108,
    ymajorgrids,
    grid style={dashed, black!25},
    symbolic x coords={t0,t1,t2,t3,t4,t5},
    xtick=data,
    tick align=outside,
    xtick pos=bottom, 
    ytick pos=left,   
    xticklabels={}, 
    ytick={0,25,50,75,100},
    yticklabel style={font=\scriptsize},
    tick style={black!60},
    axis line style={black!60},
    enlarge x limits={abs=0.55cm}, 
    every axis plot/.append style={draw=black!55, line width=0.1pt}, 
  },
  cflabels/.style={
    xticklabels={{$T=[0, 0.1]$}, {$T=[0.1, 0.3]$}, {$T=[0.3, 0.5]$}, {$T=[0.5, 0.7]$}, {$T=[0.7, 0.9]$}, {$T=[0.9, 1]$}},
    xticklabel style={rotate=45, anchor=north east, font=\scriptsize, inner sep=2pt},
  },
}

\makebox[\linewidth][c]{
\begin{subfigure}{5.8cm}
\centering
\begin{tikzpicture}
\begin{axis}[cfbase,
  title={ImageNet-1k}, title style={font=\small, yshift=-1mm}, 
  ylabel={ResNet50}, ylabel style={font=\small}]
  \addplot[fill=ad_color]   coordinates{(t0,70.0)(t1,94.17)(t2,95.0)(t3,96.67)(t4,95.83)(t5,100.0)};
  \addplot[fill=lm_color]   coordinates{(t0,8.33)(t1,22.5)(t2,44.17)(t3,50.83)(t4,63.33)(t5,94.17)};
  \addplot[fill=lma_color]  coordinates{(t0,0.83)(t1,5.83)(t2,30.83)(t3,51.67)(t4,57.5)(t5,87.5)};
  \addplot[fill=zero_color] coordinates{(t0,74.17)(t1,91.67)(t2,92.5)(t3,88.33)(t4,95.0)(t5,100.0)};
  \addplot[fill=avg_color]  coordinates{(t0,60.0)(t1,92.5)(t2,95.0)(t3,90.83)(t4,95.0)(t5,100.0)};
  \addplot[fill=max_color]  coordinates{(t0,65.83)(t1,90.83)(t2,91.67)(t3,93.33)(t4,94.17)(t5,100.0)};
  \addplot[fill=min_color]  coordinates{(t0,67.5)(t1,94.17)(t2,95.83)(t3,95.0)(t4,95.83)(t5,100.0)};
\end{axis}
\end{tikzpicture}
\end{subfigure}\hspace{2mm}%
\begin{subfigure}{5.8cm}
\centering
\begin{tikzpicture}
\begin{axis}[cfbase, yticklabel=\empty,
  title={ImageNet-v2}, title style={font=\small, yshift=-1mm}]
  \addplot[fill=ad_color]   coordinates{(t0,65.87)(t1,97.62)(t2,96.83)(t3,96.03)(t4,94.44)(t5,100.0)};
  \addplot[fill=lm_color]   coordinates{(t0,3.97)(t1,25.4)(t2,40.48)(t3,56.35)(t4,63.49)(t5,96.03)};
  \addplot[fill=lma_color]  coordinates{(t0,0.79)(t1,8.73)(t2,37.3)(t3,48.41)(t4,57.94)(t5,85.71)};
  \addplot[fill=zero_color] coordinates{(t0,72.22)(t1,93.65)(t2,93.65)(t3,93.65)(t4,94.44)(t5,100.0)};
  \addplot[fill=avg_color]  coordinates{(t0,73.81)(t1,94.44)(t2,94.44)(t3,92.86)(t4,93.65)(t5,100.0)};
  \addplot[fill=max_color]  coordinates{(t0,69.84)(t1,95.24)(t2,96.83)(t3,95.24)(t4,94.44)(t5,100.0)};
  \addplot[fill=min_color]  coordinates{(t0,65.87)(t1,93.65)(t2,96.03)(t3,95.24)(t4,95.24)(t5,100.0)};
\end{axis}
\end{tikzpicture}
\end{subfigure}\hspace{1mm}%
\begin{subfigure}{5.8cm}
\centering
\begin{tikzpicture}
\begin{axis}[cfbase, yticklabel=\empty,
  title={PASCAL-VOC}, title style={font=\small, yshift=-1mm}]
  \addplot[fill=ad_color]   coordinates{(t0,0.0)(t1,51.54)(t2,79.23)(t3,83.85)(t4,92.31)(t5,96.15)};
  \addplot[fill=lm_color]   coordinates{(t0,6.92)(t1,32.31)(t2,52.31)(t3,52.31)(t4,66.15)(t5,100.0)};
  \addplot[fill=lma_color]  coordinates{(t0,6.92)(t1,0.0)(t2,0.0)(t3,4.62)(t4,7.69)(t5,73.08)};
  \addplot[fill=zero_color] coordinates{(t0,0.0)(t1,30.77)(t2,60.77)(t3,70.77)(t4,81.54)(t5,96.15)};
  \addplot[fill=avg_color]  coordinates{(t0,0.0)(t1,17.69)(t2,53.08)(t3,65.38)(t4,77.69)(t5,96.15)};
  \addplot[fill=max_color]  coordinates{(t0,0.0)(t1,26.15)(t2,61.54)(t3,68.46)(t4,86.15)(t5,96.15)};
  \addplot[fill=min_color]  coordinates{(t0,0.0)(t1,10.0)(t2,53.08)(t3,78.46)(t4,90.77)(t5,96.15)};
\end{axis}
\end{tikzpicture}
\end{subfigure}
} 

\vspace{1.5mm} 

\makebox[\linewidth][c]{
\begin{subfigure}{5.8cm}
\centering
\begin{tikzpicture}
\begin{axis}[cfbase,
  ylabel={RegNetY}, ylabel style={font=\small}]
  \addplot[fill=ad_color]   coordinates{(t0,62.31)(t1,79.23)(t2,85.38)(t3,92.31)(t4,96.15)(t5,100.0)};
  \addplot[fill=lm_color]   coordinates{(t0,63.85)(t1,85.38)(t2,90.77)(t3,93.08)(t4,96.15)(t5,100.0)};
  \addplot[fill=lma_color]  coordinates{(t0,59.23)(t1,71.54)(t2,73.08)(t3,76.15)(t4,72.31)(t5,99.23)};
  \addplot[fill=zero_color] coordinates{(t0,55.38)(t1,80.0)(t2,86.92)(t3,85.38)(t4,90.77)(t5,100.0)};
  \addplot[fill=avg_color]  coordinates{(t0,54.62)(t1,80.0)(t2,86.15)(t3,87.69)(t4,93.08)(t5,100.0)};
  \addplot[fill=max_color]  coordinates{(t0,53.85)(t1,79.23)(t2,87.69)(t3,91.54)(t4,92.31)(t5,100.0)};
  \addplot[fill=min_color]  coordinates{(t0,50.77)(t1,86.15)(t2,85.38)(t3,86.92)(t4,92.31)(t5,100.0)};
\end{axis}
\end{tikzpicture}
\end{subfigure}\hspace{2mm}%
\begin{subfigure}{5.8cm}
\centering
\begin{tikzpicture}
\begin{axis}[cfbase, yticklabel=\empty]
  \addplot[fill=ad_color]   coordinates{(t0,65.12)(t1,88.37)(t2,95.35)(t3,90.7)(t4,96.12)(t5,100.0)};
  \addplot[fill=lm_color]   coordinates{(t0,70.54)(t1,89.15)(t2,94.57)(t3,96.12)(t4,93.02)(t5,100.0)};
  \addplot[fill=lma_color]  coordinates{(t0,55.04)(t1,76.74)(t2,77.52)(t3,80.62)(t4,82.95)(t5,100.0)};
  \addplot[fill=zero_color] coordinates{(t0,51.16)(t1,82.95)(t2,86.05)(t3,87.6)(t4,93.8)(t5,100.0)};
  \addplot[fill=avg_color]  coordinates{(t0,51.94)(t1,87.6)(t2,86.82)(t3,92.25)(t4,92.25)(t5,100.0)};
  \addplot[fill=max_color]  coordinates{(t0,58.14)(t1,86.05)(t2,86.82)(t3,93.02)(t4,96.12)(t5,100.0)};
  \addplot[fill=min_color]  coordinates{(t0,54.26)(t1,78.29)(t2,80.62)(t3,86.05)(t4,91.47)(t5,100.0)};
\end{axis}
\end{tikzpicture}
\end{subfigure}\hspace{1mm}%
\begin{subfigure}{5.8cm}
\centering
\begin{tikzpicture}
\begin{axis}[cfbase, yticklabel=\empty]
  \addplot[fill=ad_color]   coordinates{(t0,0.0)(t1,20.31)(t2,66.41)(t3,86.72)(t4,92.19)(t5,100.0)};
  \addplot[fill=lm_color]   coordinates{(t0,0.0)(t1,24.22)(t2,67.19)(t3,89.84)(t4,93.75)(t5,100.0)};
  \addplot[fill=lma_color]  coordinates{(t0,0.0)(t1,13.28)(t2,26.56)(t3,39.06)(t4,46.88)(t5,97.66)};
  \addplot[fill=zero_color] coordinates{(t0,0.0)(t1,60.16)(t2,78.12)(t3,92.19)(t4,96.88)(t5,100.0)};
  \addplot[fill=avg_color]  coordinates{(t0,0.0)(t1,39.06)(t2,81.25)(t3,86.72)(t4,94.53)(t5,100.0)};
  \addplot[fill=max_color]  coordinates{(t0,2.34)(t1,64.06)(t2,86.72)(t3,92.97)(t4,96.88)(t5,100.0)};
  \addplot[fill=min_color]  coordinates{(t0,0.0)(t1,64.06)(t2,87.5)(t3,90.62)(t4,98.44)(t5,100.0)};
\end{axis}
\end{tikzpicture}
\end{subfigure}
} 

\vspace{1.5mm} 

\makebox[\linewidth][c]{
\begin{subfigure}{5.8cm}
\centering
\begin{tikzpicture}
\begin{axis}[cfbase, cflabels,
  ylabel={EfficientNet}, ylabel style={font=\small}]
  \addplot[fill=ad_color]   coordinates{(t0,4.65)(t1,65.89)(t2,83.72)(t3,90.7)(t4,97.67)(t5,100.0)};
  \addplot[fill=lm_color]   coordinates{(t0,6.2)(t1,49.61)(t2,79.07)(t3,86.82)(t4,86.82)(t5,94.57)};
  \addplot[fill=lma_color]  coordinates{(t0,14.73)(t1,24.03)(t2,23.26)(t3,32.56)(t4,37.98)(t5,91.47)};
  \addplot[fill=zero_color] coordinates{(t0,13.18)(t1,61.24)(t2,82.17)(t3,93.8)(t4,95.35)(t5,100.0)};
  \addplot[fill=avg_color]  coordinates{(t0,11.63)(t1,67.44)(t2,89.15)(t3,93.02)(t4,98.45)(t5,100.0)};
  \addplot[fill=max_color]  coordinates{(t0,4.65)(t1,64.34)(t2,84.5)(t3,95.35)(t4,96.12)(t5,99.22)};
  \addplot[fill=min_color]  coordinates{(t0,14.73)(t1,59.69)(t2,84.5)(t3,93.02)(t4,93.8)(t5,100.0)};
\end{axis}
\end{tikzpicture}
\end{subfigure}\hspace{2mm}%
\begin{subfigure}{5.8cm}
\centering
\begin{tikzpicture}
\begin{axis}[cfbase, cflabels, yticklabel=\empty]
  \addplot[fill=ad_color]   coordinates{(t0,6.98)(t1,66.67)(t2,89.92)(t3,96.12)(t4,95.35)(t5,100.0)};
  \addplot[fill=lm_color]   coordinates{(t0,5.43)(t1,52.71)(t2,79.07)(t3,95.35)(t4,96.12)(t5,99.22)};
  \addplot[fill=lma_color]  coordinates{(t0,10.85)(t1,14.73)(t2,17.83)(t3,24.03)(t4,34.11)(t5,93.8)};
  \addplot[fill=zero_color] coordinates{(t0,14.73)(t1,72.09)(t2,85.27)(t3,96.9)(t4,93.8)(t5,100.0)};
  \addplot[fill=avg_color]  coordinates{(t0,16.28)(t1,68.22)(t2,85.27)(t3,96.12)(t4,96.12)(t5,100.0)};
  \addplot[fill=max_color]  coordinates{(t0,10.08)(t1,67.44)(t2,87.6)(t3,93.8)(t4,91.47)(t5,100.0)};
  \addplot[fill=min_color]  coordinates{(t0,11.63)(t1,66.67)(t2,86.82)(t3,92.25)(t4,96.12)(t5,100.0)};
\end{axis}
\end{tikzpicture}
\end{subfigure}\hspace{1mm}%
\begin{subfigure}{5.8cm}
\centering
\begin{tikzpicture}
\begin{axis}[cfbase, cflabels, yticklabel=\empty]
  \addplot[fill=ad_color]   coordinates{(t0,0.0)(t1,16.54)(t2,63.78)(t3,82.68)(t4,92.13)(t5,100.0)};
  \addplot[fill=lm_color]   coordinates{(t0,0.0)(t1,12.6)(t2,43.31)(t3,63.78)(t4,74.8)(t5,96.85)};
  \addplot[fill=lma_color]  coordinates{(t0,3.15)(t1,1.57)(t2,2.36)(t3,5.51)(t4,7.87)(t5,89.76)};
  \addplot[fill=zero_color] coordinates{(t0,0.0)(t1,20.47)(t2,70.87)(t3,88.19)(t4,92.91)(t5,100.0)};
  \addplot[fill=avg_color]  coordinates{(t0,0.0)(t1,22.83)(t2,70.08)(t3,84.25)(t4,93.7)(t5,100.0)};
  \addplot[fill=max_color]  coordinates{(t0,0.0)(t1,20.47)(t2,58.27)(t3,75.59)(t4,92.91)(t5,100.0)};
  \addplot[fill=min_color]  coordinates{(t0,0.0)(t1,20.47)(t2,59.06)(t3,74.02)(t4,79.53)(t5,100.0)};
\end{axis}
\end{tikzpicture}
\end{subfigure}
} 

\vspace{5mm} 

\makebox[\linewidth][c]{
\begin{tikzpicture}
  \node[draw=black, rounded corners, inner sep=6pt, font=\small\bfseries]{%
    \textcolor{ad_color}{AD}\hspace{1.5em}
    \textcolor{min_color}{Min}\hspace{1.5em}
    \textcolor{max_color}{Max}\hspace{1.5em}
    \textcolor{avg_color}{Avg}\hspace{1.5em}
    \textcolor{zero_color}{Zero}\hspace{1.5em}
    \textcolor{lma_color}{Layer Mask (Ablated)}\hspace{1.5em}
    \textcolor{lm_color}{Layer Mask}%
  };
\end{tikzpicture}
}

\vspace{2mm}